\documentclass[runningheads]{llncs}
\providecommand{\authcount}[1]{}
 
\usepackage{eccv}

\usepackage{eccvabbrv}

\usepackage{graphicx}
\usepackage{booktabs}
\usepackage{subcaption}
\usepackage{url}
\usepackage{arydshln}
\usepackage{multirow}
\usepackage{makecell}
\usepackage{arydshln}
\usepackage{algorithm}
\usepackage{algorithmic}
\usepackage{amsmath}
\usepackage{enumitem}
\usepackage[table]{xcolor}
\usepackage{array}
\newcolumntype{G}{>{\columncolor{gray!12}}c}
\usepackage{pifont}
\usepackage{float}
\usepackage{wrapfig}
\usepackage{tocloft}
\usepackage{url}            
\usepackage{booktabs}       
\usepackage{amsfonts}       
\usepackage{nicefrac}       
\usepackage{microtype}      
\usepackage{xcolor}         
\usepackage{multirow}
\usepackage{makecell}
\usepackage{diagbox} 
\usepackage{xspace}
\usepackage{multirow}   
\usepackage{makecell}   
\usepackage{arydshln}  
\usepackage{graphicx} 
\usepackage[accsupp]{axessibility}  

\usepackage{titletoc}

\usepackage[pagebackref,breaklinks,colorlinks,citecolor=eccvblue]{hyperref}
\usepackage{hyperref}
\definecolor{myblue}{RGB}{30,90,160}
\usepackage{orcidlink}

\begin{document}

\title{Universal Image Immunization against Diffusion-based Image Editing via Semantic Injection} 

\titlerunning{Universal Image Immunization}

\author{Chanhui Lee\inst{1} 
Donggyu Choi\inst{2} 
Seunghyun Shin\inst{2} 
Hae-Gon Jeon\inst{3} 
Jeany Son\inst{1}\thanks{Corresponding author.}}

\authorrunning{C.~Lee et al.}

\institute{\textsuperscript{1} POSTECH \quad
\textsuperscript{2} GIST \quad
\textsuperscript{3} Yonsei University\\
\url{https://ChanhuiLee1111.github.io/Universal-Immunization}}
\maketitle
\begin{abstract}
Diffusion model advances have enabled powerful text-guided image editing, but also raise ethical and legal risks such as deepfakes and unauthorized use.
To prevent these risks, adversarial attack-based image immunization has emerged as a promising defense against AI-driven semantic manipulation.
Yet, most existing approaches require image-specific optimization or additional neural networks at inference time, hindering scalability and practicality.
In this paper, we propose the first universal adversarial perturbation-based image immunization framework that generates a single, image-agnostic adversarial perturbation specifically designed for diffusion-based editing pipelines.
Inspired by UAP used in targeted attacks, our method aims to generate a UAP that induces diffusion models to misinterpret the input image as a specific semantic target. 
Simultaneously, it suppresses original content to misdirect the model's attention during editing, thereby effectively blocking unauthorized edits by overwriting the image’s original semantics via the UAP.
Extensive experiments show that our method, as the first universal immunization approach, significantly outperforms several baselines in the UAP setting. 
Notably, despite the inherent difficulty of universal perturbations, our method achieves competitive or superior performance compared to image-specific methods under a more restricted perturbation budget, while also exhibiting strong black-box transferability across diverse diffusion models.
\end{abstract}

\section{Introduction}
Diffusion models have emerged as a dominant paradigm in image synthesis, generating  high-fidelity, semantically rich images by iteratively denoising Gaussian noise through a learned reverse process~\cite{dhariwal2021diffusion, ho2020denoising, song2020denoising, songscore}.
Conditional extensions guided by text prompts, spatial masks, or reference images via cross-attention further enable fine-grained, user-controllable generation, and editing~\cite{rombach2022high,brooks2023instructpix2pix,zhang2023adding, sd3, flux}.
While diffusion models enable creative and interactive applications, their powerful user-guided generation capabilities also raise serious concerns about misuse, such as imperceptible malicious edits, identity spoofing, or unauthorized replication of copyrighted content~\cite{risk1, risk2, risk3, risk4}.
This highlights the need for robust defense mechanisms to mitigate harmful edits generated by diffusion models.

\begin{figure}[!t]
\centering    \includegraphics[width=\linewidth]{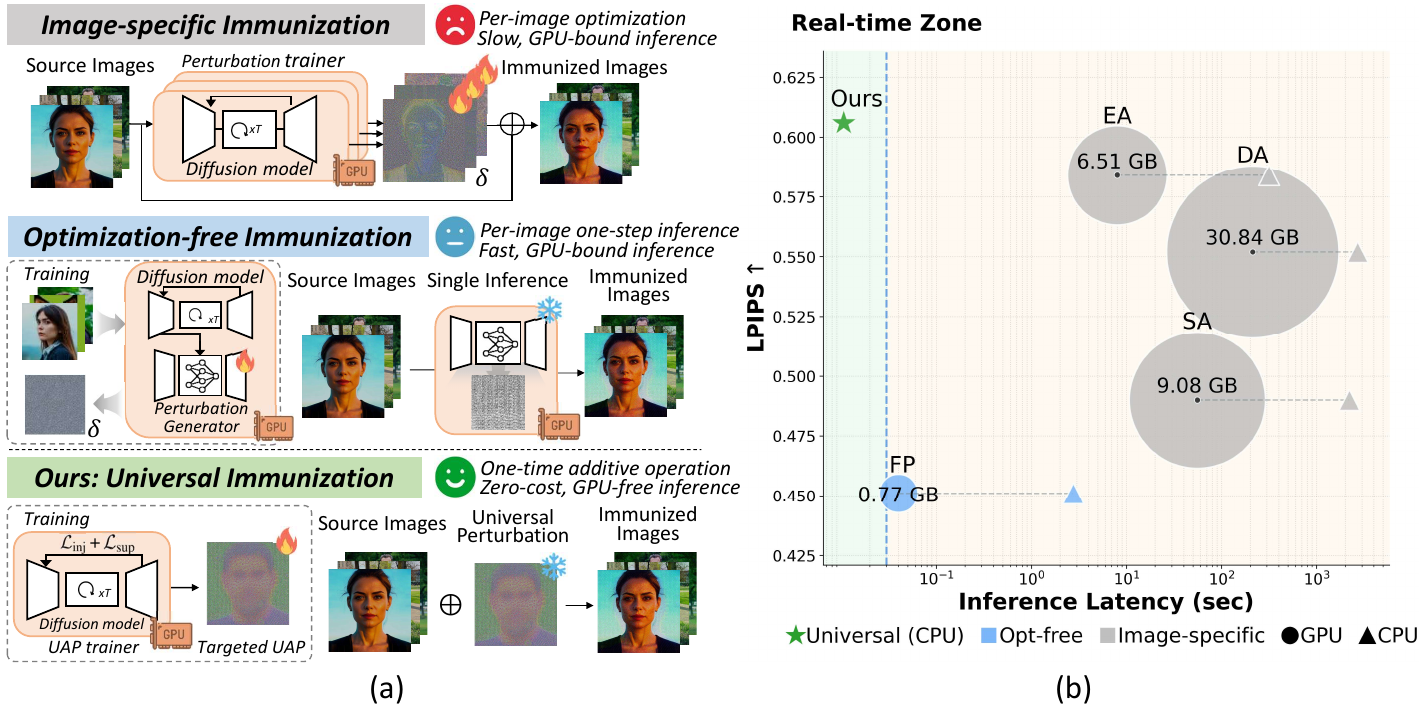}
    \caption{Illustrations of our universal immunization approach and comparison with prior image editing defenses. 
    (a) Unlike image-specific and optimization-free approaches (top and mid) requiring GPUs to per-image processing, our universal immunization method (bottom) employs a single, pre-computed UAP to safeguard images without any inference-time overhead.
    (b) Trade-off between latency, performance, and memory. Our method performs immunization with a simple addition operation with pretrained UAP, enabling nearly zero-cost inference while maintaining strong immunization performance. 
    EA, DA, SA, and FP denote Encoder Attack~\cite{raising}, Diffusion Attack~\cite{raising}, Semantic Attack~\cite{semantic}, and FastProtect~\cite{nearly}, respectively.
    }
    \label{fig:motivation}
\end{figure}

To prevent unauthorized image editing, adversarial perturbation-based immunization methods~\cite{LaS, ltp, raising, semantic, diffguard, trippodo2025immunizing} have been proposed to embed imperceptible perturbations into source images and disrupt potential manipulations.
With the growing adoption of diffusion models for image editing, recent approaches have mostly focused on diffusion-based editing pipelines, including text-guided image-to-image editing~\cite{raising, semantic, trippodo2025immunizing}, image inpainting~\cite{diffguard, advpaint, diffvax}, instruction-guided image editing~\cite{editshield, editaway}, and diffusion customization~\cite{nearly, gridrobust, styleguard}.
However, most existing approaches are image-specific (top row of Figure~\ref{fig:motivation}(a)), relying on computationally intensive per-image optimization at inference time, which limits their practicality under real-world latency and resource constraints. 

This deployment bottleneck motivates more efficient protection mechanisms. 
One potential direction is universal adversarial perturbations (UAP)~\cite{uap}, originally studied in image classification, where a single perturbation is shared across multiple inputs to amortize computational cost. 
Although classification studies demonstrate clear efficiency benefits, universality is often presumed to weaken protection strength and remains largely unexplored in the context of diffusion-based editing defenses. 
Another line of work seeks to improve efficiency through optimization-free approaches~\cite{nearly, diffvax} (middle row of Figure~\ref{fig:motivation}(a)).
These methods remove costly per-image optimization by employing pre-trained perturbation generators or fusion modules to produce image-adaptive perturbations in a single forward pass.
Although this strategy reduces computation time, it still relies on additional neural networks and GPU acceleration, and incurs non-trivial memory overhead to achieve real-time performance, limiting scalability in resource-constrained settings.

In this paper, we propose a novel universal adversarial perturbation (UAP) framework for immunizing images against diffusion-based editing.
Our method generates an image-agnostic perturbation that eliminates per-image optimization and image-adaptive inference, enabling efficient deployment without specialized hardware, achieving a better performance--latency--memory trade-off, as shown in Figure~\ref{fig:motivation}(b).
Inspired by targeted UAPs in classification~\cite{targeted-uap}, our method embeds target semantics into source images to induce diffusion models to misinterpret the original content.
To effectively inject target semantics into a UAP, we introduce two loss functions: a \textit{target semantic injection} loss, which encourages the embedding of target content into the perturbation, and a \textit{source semantic suppression} loss, which reduces the influence of the source image’s content.
Together, these objectives encourage diffusion models to interpret the perturbed image as containing target semantics at the input level, causing the models to act on the injected signal rather than the source content and thereby preventing unauthorized or malicious edits.

\begin{figure}[!t]
\centering    \includegraphics[width=\linewidth]{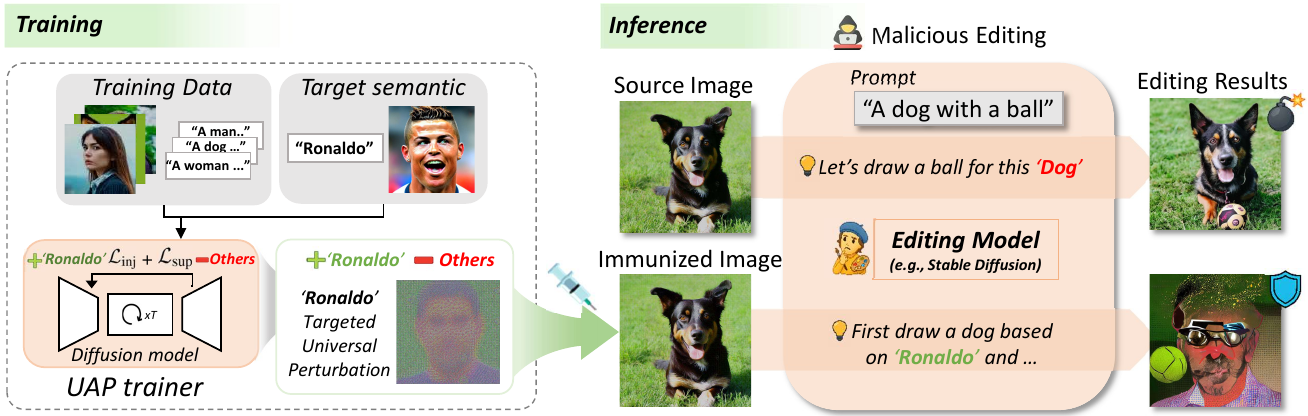}
    \caption{Framework and motivation of the proposed target semantic injection approach. A pretrained UAP injects target semantics into source images, misleading the editing model to interpret the source as the target concept (\eg, treating a dog as ``Ronaldo'' during editing), which causes it to lose the original content and produce failed edits. Note that the perturbations are scaled for improved visualization.
    }
    \label{fig:teasure}
\end{figure}

\begin{figure}[!t]
\centering    \includegraphics[width=0.95\linewidth]{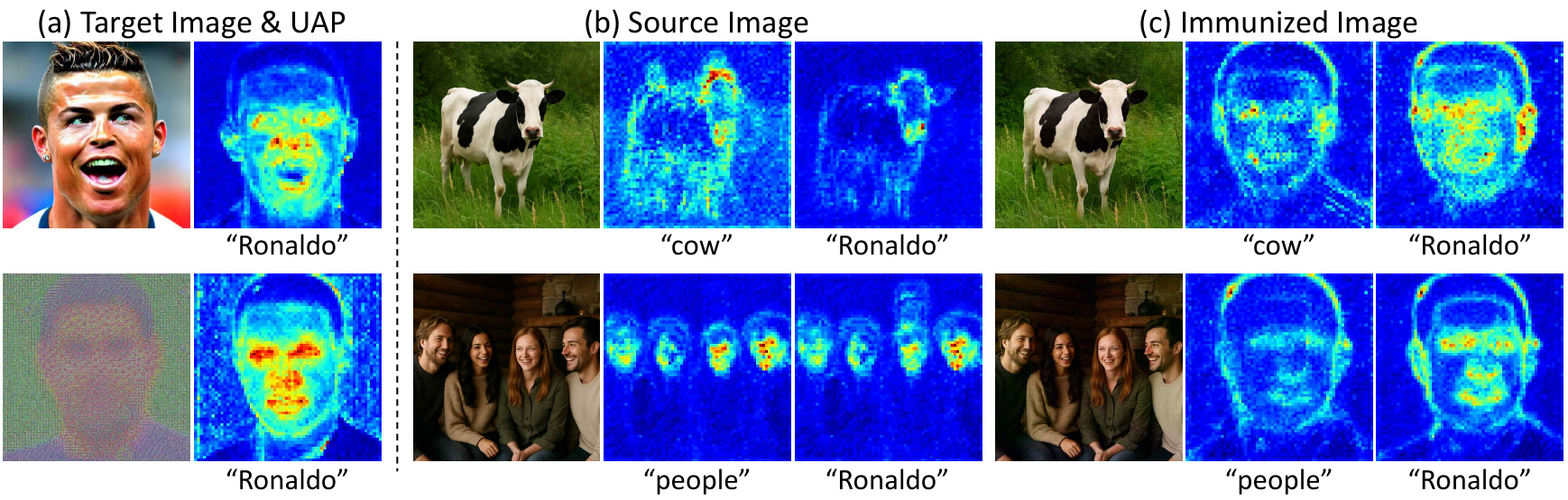}
    \caption{
    Visualization of cross-attention maps from the diffusion model under different conditions.
    (a) Target images generated by Stable Diffusion~\cite{rombach2022high} conditioned on the prompt `Ronaldo', along with the generated corresponding UAP.
    (b) {Source images show strong attention to their own content, but since they do not intrinsically contain the target semantics (`Ronaldo'), they are not able to produce any target-aligned attention.}
    (c) Immunized images, perturbed by the UAP, fail to focus on the original prompt and instead exhibit strong attention to the target concept `Ronaldo'. 
    Each attention map is annotated with the prompt used for conditioning.
    }
    \label{fig:attention}
\end{figure}

Our motivation for semantic injection is further illustrated in Figure~\ref{fig:teasure}.
Our UAP embeds target semantics into the source image, encouraging the diffusion model to edit the immunized input under the target semantics (e.g., \textit{`Ronaldo'}) rather than the true source content (e.g.,  \textit{`dog'}).
As a result, the generated output is unrelated to the true source content.
Additionally, the UAP jointly suppresses the original semantics of the source image, further reinforcing the dominance of the target and guiding the editing process toward target-{conditioned} generation.
The effectiveness {and empirical evidence} of our method are further demonstrated in Figure~\ref{fig:attention}.
As shown in Figure~\ref{fig:attention}(b), without immunization, the source image activates attention for the actual contents presented in the images, such as \textit{`cow'} and \textit{`people'}.
In contrast, the attention map of the immunized image (Figure~\ref{fig:attention}(c)) is sharply focused on the intended target, \textit{`Ronaldo'}, closely resembling the attention pattern in the target image (Figure~\ref{fig:attention}(a)), while failing to attend to the original semantic content.
This indicates that our UAP aligns the model's focus with the desired target semantic content{, effectively overriding and disrupting} the original semantic representation.  

Extensive experiments demonstrate that our method consistently and substantially outperforms all baselines across both quantitative metrics and qualitative analyses.
Moreover, it exhibits strong black-box transferability, maintaining high performance across diverse diffusion-based editing models without model-specific tuning.
Importantly, despite the substantially more challenging setting, our approach matches and even surpasses several state-of-the-art image-specific methods while reducing inference cost to nearly zero.
It also remains effective in data-free scenarios, where no real-world training data or prior domain knowledge is available, underscoring its practicality for real-world deployment.

The main contributions can be summarized as follows:
\begin{itemize}[label={$\bullet$}]
    \item We propose a novel universal immunization strategy that uses targeted universal adversarial perturbations (UAPs) to defend against malicious edits in diffusion-based models, {eliminating test-time adaptation and removing a key deployment bottleneck.}
    To the best of our knowledge, this is the first work to achieve image immunization via a single, input-agnostic perturbation.
    \item {We introduce two loss functions for training our UAP: the \textit{target semantic injection} loss encourages the injection of intended target semantics, leading the model to misinterpret the original contents,} and the \textit{source semantic suppression} loss suppresses {original} concepts present in the training set, enabling effective immunization on unseen inputs.
    \item {Our universal approach not only consistently surpasses universal baselines across diverse diffusion models, including our data-free variant, but also remains competitive with, or even outperforms, image-specific immunization methods despite the inherent difficulty of the universal setting, demonstrating strong black-box generalization and high test-time efficiency.}
\end{itemize}

\section{Related Work}
\paragraph{\textbf{\textup{Text-Conditioned Diffusion Models.}}}
Text-to-image diffusion models generate images conditioned on text prompts by learning to reverse a gradual noising process.
Recent models such as DALL·E 2~\cite{ramesh2022hierarchical}, Imagen~\cite{saharia2022photorealistic}, and Stable Diffusion~\cite{rombach2022high} have demonstrated remarkable performance by leveraging language models and cross-attention mechanisms.
Building on this success,
DiffEdit~\cite{couairon2022diffedit} introduces mask-based editing by identifying editable regions, while Imagic~\cite{kawar2023imagic} fine-tunes diffusion models on individual images to preserve identity.
InstructPix2Pix~\cite{brooks2023instructpix2pix} advances this direction by training on instruction--edit pairs, enabling intuitive edits from natural language instructions.
While these models enable powerful editing, their malicious use raises concerns about content authenticity and motivates image protection methods for safer, more controllable editing.

\paragraph{\textbf{\textup{Image Immunization.}}}
Early works~\cite{ltp, tafim, LaS} primarily focused on defending against generative adversarial networks (GANs).
As diffusion models~\cite{ramesh2022hierarchical, saharia2022photorealistic, rombach2022high, brooks2023instructpix2pix} advanced, research focus shifted to diffusion-based immunization.
Following PhotoGuard~\cite{raising}, which introduced encoder- and diffusion-level attacks for diffusion-based editing, subsequent studies, including Semantic Attack~\cite{semantic} and Adversarial Attention~\cite{trippodo2025immunizing} further suggested disruption of text-image alignment.
For image inpainting, DiffusionGuard~\cite{diffguard} and AdvPaint~\cite{advpaint} proposed mask-aware perturbation optimization that targets the denoising process and attention embeddings, respectively.
EditShield~\cite{editshield} and FaceLock~\cite{editaway} targeted instruction-guided image editing models, aiming to prevent unauthorized edits and facial identity privacy.
To improve robustness against perturbation purification techniques~\cite{jpeg, gridpure, diffpure}, BlurGuard~\cite{blurguard} and DCT-Shield~\cite{dct} optimize perturbations in low-frequency domains, while AntiPure~\cite{gridrobust} specifically targeted diffusion-based purification in diffusion customization.
More recently, FastProtect~\cite{nearly} and DiffVax~\cite{diffvax} reduced inference-time cost via a mixture-of-perturbation design and a learned immunizer, respectively.
While these works~\cite{nearly, diffvax} reduce inference cost over earlier optimization-based approaches~\cite{raising, semantic, advpaint, diffguard, blurguard, gridrobust}, they either show a performance gap to strong image-specific methods or require GPU acceleration for practical runtime and pretrained immunizer, limiting real-world scalability.

\paragraph{\textbf{\textup{Universal Adversarial Perturbation.}}}
Universal Adversarial Perturbations (UAPs), first introduced in~\cite{uap}, aim to generate a single, image-agnostic perturbation capable of fooling a wide range of inputs. 
UAPs can be crafted in either data-dependent~\cite{sga, GAP, spgd} or data-free settings~\cite{cosine-uap, gd-uap, trm, psp}.
Data-dependent methods rely on representative datasets to optimize perturbations across samples, while data-free methods use synthetic inputs, such as Gaussian noise or jigsawed images, making them suitable for scenarios where access to training data or the target domain is restricted. 
{Yet, UAPs remain less effective than image-specific perturbations in classification~\cite{MIfg, gapp, fsim, Deepfool}, because the challenge of deceiving all images simultaneously inevitably limits their performance.}
Furthermore, the development of universal perturbations for generative models, particularly diffusion models, remains largely underexplored.
\section{Background and Motivation}

\paragraph{\textbf{\textup{Targeted Universal Adversarial Perturbation (UAP).}}}

UAPs have been explored in the context of targeted attacks~\cite{targeted-uap}, demonstrating that a single, input-agnostic perturbation can effectively drive diverse inputs toward a specific target label $y_\text{tar}$.
The objective is defined as:
\begin{align}
    \delta^* = \arg \min_{\delta} \ \mathbb{E}_{x \sim \mathcal{D}} \left[ \mathcal{L}(f(x + \delta), y_{\text{tar}}) \right], \quad \text{s.t.} \quad \| \delta \|_{\infty}
\ \leq \epsilon,
    \label{eq:targeted_uap}
\end{align}
where $\mathcal{D}$ is the data distribution, {$x$ is an input image sampled from $\mathcal{D}$,} $f$ is the classifier, $\mathcal{L}$ is the loss function (\eg, cross-entropy), $\delta$ is the UAP, and $\epsilon$ is the perturbation budget.
By leveraging the diverse semantic of the data to encode target-aligned features into a UAP, costly per-image optimization can be avoided.

Motivated by this, we extend targeted UAPs to immunize images against diffusion-based image editing by embedding target semantics into a single, universal perturbation.
In this way, the model is encouraged to rely more on a disrupted semantic interpretation shifted toward the target, rather than on the original source content.
Unlike prior image-specific approaches~\cite{raising,semantic} that rely on slow per-image optimization, our method {trains} a UAP that generalizes across inputs, enabling efficient and scalable protection against diffusion-based editing.

\paragraph{\textbf{\textup{Text-Image Cross-Attention in Diffusion Models.}}}
{Text-to-image diffusion models~\cite{ramesh2022hierarchical, saharia2022photorealistic, rombach2022high}
generate images by progressively denoising latent variables while conditioned on textual inputs.
Their training follows the standard diffusion objective:
\begin{equation}
    \mathcal{L}(\theta)
    = \mathbb{E}_{x_0, t, \epsilon, c}
    \left[
        \left\| \epsilon_\theta(x_t, k, t) - \epsilon \right\|_2^2
    \right],
    \label{eq:diffusion_obj}
\end{equation}
where the noise prediction network $\epsilon_\theta$ learns to denoise a wide variety of images while conditioning on diverse text prompts $t$.
To achieve this, the U-Net backbone incorporates cross-attention layers at multiple spatial resolutions, allowing textual embeddings to influence the latent representation at each timestep $k$.
Let \( z \in \mathbb{R}^{N \times d} \) denote latent features and
\( t \in \mathbb{R}^{M \times d} \) the text embeddings.
The cross-attention output CA$_l$ is computed as:
\begin{equation}
    {A}_l = \mathrm{softmax}\!\big(Q_l(z) K_l(t)^\top/\sqrt{d_k}\big),~~
\text{CA}_l = {A}_l V_l(t).
\label{eq:cross_attention}
\end{equation}
where \( Q_l, K_l, V_l \) are learned projections and ${A}_l$ refers attention map at layer $l$, respectively.
 ${A}_l$ specifies the attention weights between image and text tokens, and when multiply with ${V}_l$, the resulting cross-attention output ${A}_l {V}_l$ injects textual semantics into the latent update.
This mechanism enables each spatial location in the latent image to selectively attend to semantically relevant components of the text prompt.
Recent works~\cite{semantic, advpaint} have exploited cross-attention for image-specific immunization by manipulating attention maps~\cite{semantic} or altering intermediate image representations~\cite{advpaint, silva2025attacking}.
However, they often lack a direct and scalable intervention for controlling target semantics within diffusion models.
{Distinct from previous works, our semantic injection approach guides diffusion models to interpret the immunized image itself as containing the target semantics.
To accomplish this, the proposed losses operate directly on the cross-attention outputs.
Detailed theoretical justifications and ablation results are provided in supplementary material (Sections~\textcolor{red}{A.5} and~\textcolor{red}{C.1}}).


\section{Method}
\label{sec_method}

In this section, we present our universal immunization approach for disrupting diffusion-based image editing.
Our method injects intended target semantics into images while suppressing the original content, and remains effective even in a data-free setting, enabling practical and scalable protection.

\subsection{Universal Image Immunization via Semantic Injection}
Previous image immunization methods primarily focus on image-specific approaches~\cite{raising, semantic, diffguard, advpaint}, where a distinct adversarial perturbation is generated for each source image.
While effective, these methods incur significant computational overhead to craft each immunized image, as illustrated in Figure~\ref{fig:motivation}(b).
To address this, we propose a universal immunization method that learns a universal adversarial perturbation (UAP) through one-time training, which can then be effortlessly applied to any input at inference.

Inspired by targeted UAPs in classification~\cite{targeted-uap}, our approach aims to overwrite the original semantics of source images, leading denoiser $\epsilon_\theta$ to refer to the intended target concepts through an adversarial perturbation.
To this end, we introduce two loss terms for training a single, general-purpose perturbation: the \textit{target semantic injection loss} $\mathcal{L}_{\text{inj}}$, which encourages alignment with the intended target semantics, and the \textit{source semantic suppression loss} $\mathcal{L}_{\text{sup}}$, which discourages retention of the original content.
The overall optimization objective for training UAPs in our universal immunization framework, based on the two proposed loss terms, $\mathcal{L}_{\text{inj}}$ and $\mathcal{L}_{\text{sup}}$, is defined as follows:
\begin{align}
    \delta^* = \arg \min_{\delta} \ \mathbb{E}_{(x, t) \sim \mathcal{D}_p} \left[ \mathcal{L}_\text{inj} +\mathcal{L}_\text{sup} \right], \quad \text{s.t.} \quad \| \delta \|_{\infty}
\ \leq \epsilon,
    \label{eq:our_objective}
\end{align}
where $(x,t)\sim \mathcal{D}_p$ denotes an input image and its corresponding text embeddings from the image-prompt pair data distribution.
Each loss component is described in detail below.

\paragraph{\textbf{\textup{Target Semantic Injection.}}}
Our primary objective is to train a UAP that leads the diffusion model to interpret the source image as containing target semantic content.
To achieve this, we encourage the cross-attention responses of immunized images—perturbed by the UAP and conditioned on a target prompt—to closely match those of genuine target images guided by the same prompt.
To ensure semantic alignment across multiple spatial feature levels, we aggregate cross-attention outputs from all layers of U-Net.
Formally, the proposed \textit{target semantic injection loss} is defined as follows:
\begin{align}
    \mathcal{L}_{\text{inj}} =
    \sum_{\ell=1}^{L} 
    \left\|
 \text{CA}_l(\Phi^\ell(\mathcal{E}(x + \delta)),\ t_{\text{tar}})
        - 
        \text{CA}_l(\Phi^\ell(\mathcal{E}(x_{\text{tar}})),\ t_{\text{tar}})
    \right\|_2^2,
\label{eq:semantic_injection}
\end{align}
where $\text{CA}$ denotes the cross-attention output described in Eq.~(\ref{eq:cross_attention}),  $\Phi^\ell(\cdot)$ represents an intermediate feature map after $\ell$-th intermediate block of the denoising U-Net, $L$ is the number of intermediate blocks in the U-Net, $\mathcal{E}$ is an encoder of a variational auto-encoder (VAE)~\cite{kingma2014, van2017neural}.
{Here, $x_{tar}$ and $t_{tar}$ denote a target image and CLIP~\cite{clip} text embedding of the target prompt (\eg, `Ronaldo', `tiger'), respectively.}
The resulting UAP is trained to align the internal attention outputs of diverse source images with those of the target, effectively injecting target semantics in a consistent and image-agnostic manner. 
Even though {the editing results of} the immunized image does not perfectly resemble the target, the disruption of the original semantics is typically sufficient to prevent faithful content preservation during malicious image-to-image editing.

\paragraph{\textbf{\textup{Source Semantic Suppression.}}}
To further misalign the immunized image from its original semantics, we introduce a \textit{source semantic suppression loss}.
Inspired by prior work on adversarial attacks~\cite{topk}, which improves attack effectiveness by minimizing the cross-entropy with the target label while maximizing it for non-target labels, we similarly aim to suppress  source semantics in diffusion models.
Specifically, we maximize the discrepancy between the cross-attention outputs of the original image and its UAP-perturbed counterpart. 
The loss is defined as:
\begin{align}
    \mathcal{L}_{\text{sup}} = 
    - \sum_{\ell=1}^{L} 
    \left\|
        \text{CA}_l(\Phi^\ell(\mathcal{E}(x + \delta)),\ t)
        - 
        \text{CA}_l(\Phi^\ell(\mathcal{E}(x)),\ t)
    \right\|_2^2.
    \label{eq:semantic_supp}
\end{align}
The loss $\mathcal{L}_{\text{sup}}$ is jointly optimized with the target semantic injection loss $\mathcal{L}_{\text{inj}}$, enabling stronger targeted attacks by simultaneously erasing source semantics and reinforcing the injected target semantics. 
Specifically, while $\mathcal{L}_{\text{inj}}$ encourages alignment of the adversarial image with the target's semantics, $\mathcal{L}_{\text{sup}}$ drives its deviation from the original source semantics.


\subsection{Extension to Data-free Setting}
Our target semantic injection loss is designed to overwrite an image's original semantics with an arbitrary target, regardless of the input, making it effective even in data-free settings.
While training with real data can enhance the consistency and strength of the injected semantics, we show that our method remains reliable without access to any real images.
Following prior data-free universal attack methods~\cite{psp, trm}, we sample synthetic inputs {$x_r$ from a random prior distribution $\mathcal{D}_r$} (\eg, Gaussian noise or jigsaw puzzles) and treat them as training data.
We then optimize UAP $\delta$ using only the target semantic injection loss in Eq.~(\ref{eq:semantic_injection}).
This enables our data-free  approach to generate a targeted UAP from a single target image, without requiring real-world training data, while effectively defending against malicious editing across diverse inputs. 
Additional details and examples of random prior synthesis are provided in supplementary material (Section~\textcolor{red}{A.4}).
\subsection{Overall Algorithm}
The overall training and testing procedures for our universal image immunization framework are provided in Algorithm~\textcolor{red}{1} and Algorithm~\textcolor{red}{2} of supplementary material, respectively.
We adopt the optimization strategy of the image-specific immunization method, Semantic Attack~\cite{semantic}, extending it to a universal setting by training a single perturbation across the entire dataset instead of per-test image optimization.
Once trained, the universal perturbation $\delta$ can be directly applied to any new input image at test time as $x_{\text{new}} + \delta$, enabling fast and scalable immunization.

\section{Experiments}

\label{sec_exp}

\subsection{Experimental Setup}
{Since our work presents the first universal immunization approach against diffusion-based image editing, no directly comparable methods exist under the same setting.
Therefore, we establish baselines for universal image immunization by adapting three representative image-specific methods: (i) the encoder attack from PhotoGuard~\cite{raising} (\textit{Encoder}), (ii) AdvPaint~\cite{advpaint}, which alters intermediate embeddings of self-attention and cross-attention (\textit{Embedding}), and (iii) Semantic Attack~\cite{semantic}, which manipulates cross-attention maps (\textit{Map}).
Detailed implementation setup and modifications of the universalized baselines are provided in the supplementary materials (Section~\textcolor{red}{A.6}).
To further assess the effectiveness and efficiency of our approach, we also compare against state-of-the-art image-specific immunization methods, including PhotoGuard~\cite{raising}—with its Encoder Attack (EA) and Diffusion Attack (DA)—and the Semantic Attack (SA)~\cite{semantic}, as well as a optimization-free method, FastProtect~\cite{nearly} for image editing.}

\paragraph{\textbf{\textup{Implementation Details.}}}
In our experiments, we train UAPs on the Stable Diffusion V1.5~\cite{rombach2022high} model available on Hugging Face.
{Following prior work on classifier UAPs~\cite{psp,trm}, we set $\epsilon=10/255$ to limit visible artifacts, as a single perturbation must generalize across inputs. 
For image-specific methods, we adopt the convention~\cite{raising,semantic} with $\epsilon=16/255$.}
The number of optimization iterations is set to 100 for image-specific methods, while ours is trained for 20 epochs.
The timestep parameter in Algorithm~\textcolor{red}{1} is set to $T=\{5, 10, 15, 20, 25\}$.
We adopt the default setting of Stable Diffusion V1.5 for all remaining hyperparameters.
To evaluate black-box transferability, we further test our trained UAPs on Stable Diffusion V1.4, V2.0~\cite{rombach2022high}, and instruction-based InstructPix2Pix~\cite{brooks2023instructpix2pix}.

\begin{figure}[!t]
\centering    \includegraphics[width=1\linewidth]{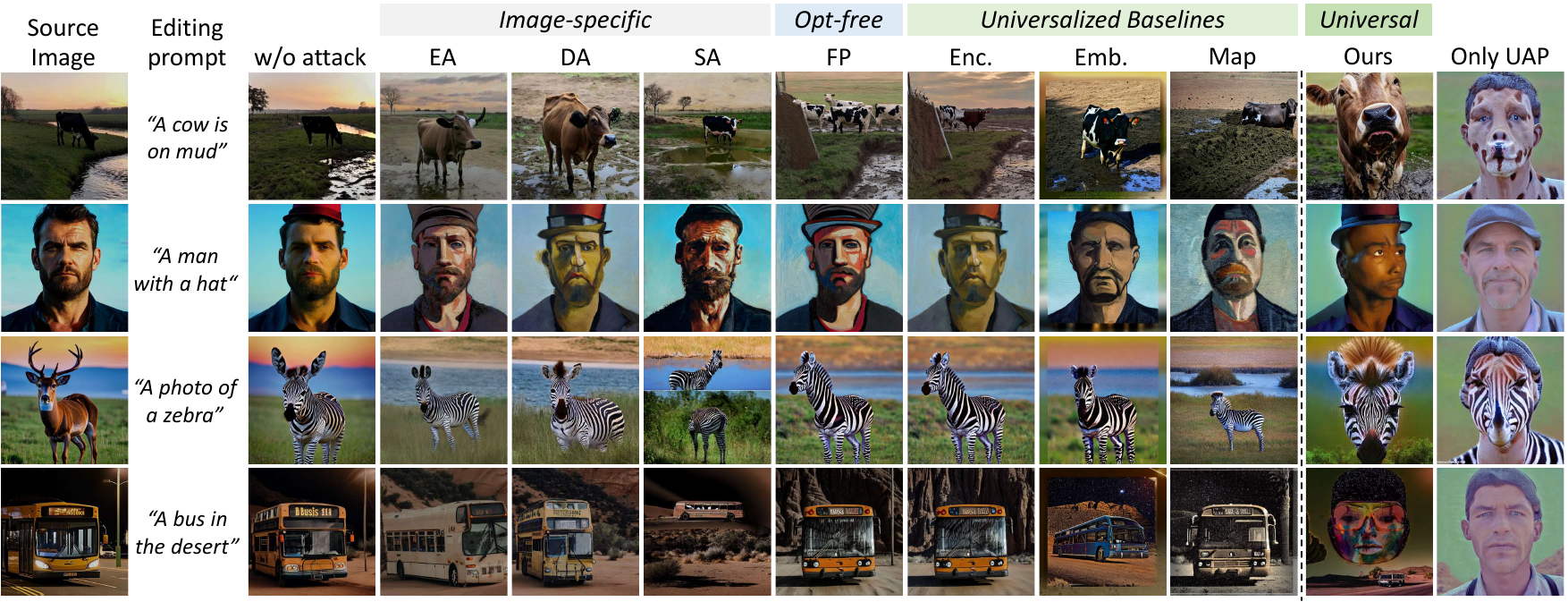}    
    \caption{
    {Qualitative comparison with universalized baselines and image-specific methods. 
    `Enc.', `Emb.', and `Map' represent the universalized baselines, obtained by adapting image-specific methods: encoder attack~\cite{raising}, intermediate representation disruption (e.g., query/key/value embeddings)~\cite{advpaint}, and attention map attack~\cite{semantic}, respectively.
    `FP' refers FastProtect~\cite{nearly}, optimization free immunization method, and `EA', `DA', and `SA' indicate image-specific immunization methods: Encoder-, Diffusion-~\cite{raising}, and Semantic-Attack~\cite{semantic}.
    We use `\textit{Ronaldo}' as the target prompt for UAP generation.}
    }
    \label{fig:qual_edit}
\end{figure}
\paragraph{\textbf{\textup{Dataset and Evaluation Protocol.}}}
To train UAPs, we randomly sample 10,000 image-prompt pairs from LAION-2B-en~\cite{laion} in the data-dependent setting.
For the data-free setting, we follow prior UAP methods~\cite{psp, gd-uap} by using randomly generated jigsaw puzzle images as training data.
For evaluation, we generate 50 images for each of 10 distinct object classes using the diffusion model~\cite{rombach2022high}, following the data generation procedure used in the evaluation protocol of Semantic Attack~\cite{semantic}.
Specifically, for each class, we define two editing prompts: one for object edits, the other for non-object edits (\ie, background).
To better assess generalization, we use a broader set of 10 object classes and a total of 500 images, compared to the 3 classes and 150 images used in~\cite{semantic}.
We further evaluate our method on the real-image dataset ImageNet-Edit~\cite{blurguard}.
Target images are defined by arbitrary prompts and generated via SD 1.5~\cite{rombach2022high},
{as in dataset construction.}
We use `\textit{tiger}' for evaluation and `\textit{Ronaldo}' for qualitative visualization. 

{We follow the evaluation protocol of prior works~\cite{raising, semantic, dct}, reporting PSNR, SSIM~\cite{ssim}, VIFp~\cite{vifp}, FSIM~\cite{fsim}, and LPIPS~\cite{lpips}.
Additionally, we assess how closely the edited image preserves the source image by using semantic alignment metrics based on CLIP~\cite{clip} feature similarity between the source and edited images.}
All results are averaged over 10 random seeds to ensure statistical robustness.
Further details on dataset construction, visualizations on various targets, and evaluation protocols are provided in supplementary material (Section~\textcolor{red}{A.1}).

\subsection{Main Results}

\paragraph{\textbf{\textup{Qualitative Results.}}}
We present qualitative comparison on image editing tasks with universalized baselines (\ie, Encoder, Embedding, and Map) along with image-specific (\ie, EA, DA, and SA) and optimization-free (FP) methods.
As shown in Figure~\ref{fig:qual_edit}, the edited outputs clearly ignore the original source content.
For instance, prompt applied to the \textit{bus} image yields unrelated human figure.
In the case of \textit{man}, \textit{cow}, and \textit{deer}, the outputs follow their prompts but deviate substantially from the originals and often appear unnatural, indicating that the model relies more on the injected semantics than on the source content.
This effect is further supported by the similarity between outputs from immunized images and those edited using the UAP alone (Only UAP), as well as the attention maps in Figure~\ref{fig:attention}(c), highlighting the dominant influence of the injected target.
More qualitative results are provided in the supplementary material (Section~\textcolor{red}{D}).

\begin{table}[t!]
\footnotesize
\centering
\caption{White-box comparison with universalized baslines on Stable Diffusion v1.5. \textit{Ours} is the full method, and \textit{Ours\textsubscript{DF}} is its data-free variant using only target semantic injection. Best and second-best results are shown in bold and underline.
}
\resizebox{0.85\linewidth}{!}{
\begin{tabular}{c|c|ccc|cc}
\Xhline{0.35mm}
\multirow{2}{*}{Method} & \multirow{2}{*}{Clean}
& \multicolumn{3}{c|}{\cellcolor{green!15}\textit{Universalized Baselines}}
& \multicolumn{2}{c}{\cellcolor{green!35}\textit{Universal}} \\
& & \cellcolor{green!15}Encoder & \cellcolor{green!15}Embedding & \cellcolor{green!15}Map
& \cellcolor{green!35}Ours$_{\text{DF}}$ & \cellcolor{green!35}Ours \\
\Xhline{0.1mm}
PSNR $\downarrow$ &  --& 16.55{\scriptsize$\pm$0.073} & 15.80{\scriptsize$\pm$0.042} & 16.16{\scriptsize$\pm$0.056} & \underline{14.68}{\scriptsize$\pm$0.035} & \textbf{14.19}{\scriptsize$\pm$0.059}  \\
SSIM $\downarrow$ &  --&0.482{\scriptsize$\pm$0.004} & \underline{0.378}{\scriptsize$\pm$0.003} & 0.468{\scriptsize$\pm$0.004} & \underline{0.378}{\scriptsize$\pm$0.002} & \textbf{0.332}{\scriptsize$\pm$0.003} \\
VIFp $\downarrow$ & --& 0.154{\scriptsize$\pm$0.002} & 0.117{\scriptsize$\pm$0.001} & 0.152{\scriptsize$\pm$0.002} & \underline{0.106}{\scriptsize$\pm$0.001} & \textbf{0.082}{\scriptsize$\pm$0.001}  \\
FSIM $\downarrow$ &  --&0.714{\scriptsize$\pm$0.005} & 0.689{\scriptsize$\pm$0.004} & 0.710{\scriptsize$\pm$0.004} & \underline{0.666}{\scriptsize$\pm$0.002} & \textbf{0.642}{\scriptsize$\pm$0.003}  \\
LPIPS $\uparrow$  & --& 0.452{\scriptsize$\pm$0.005} & 0.548{\scriptsize$\pm$0.003} & 0.465{\scriptsize$\pm$0.004} & \underline{0.557}{\scriptsize$\pm$0.002} & \textbf{0.606}{\scriptsize$\pm$0.003}  \\
Feat. Sim. (C) $\downarrow$ & 0.744 &  0.708{\scriptsize$\pm$0.004} & 0.696{\scriptsize$\pm$0.005} & 0.704{\scriptsize$\pm$0.001} & \underline{0.685}{\scriptsize$\pm$0.003} & \textbf{0.673}{\scriptsize$\pm$0.002} \\
\Xhline{0.35mm}
\end{tabular}
}
\label{white-box_universal}
\end{table}

\begin{table}[t]
\footnotesize
\setlength{\tabcolsep}{3pt}
\centering
\caption{Black-box immunization transferability on Stable Diffusion V1.4, V2.0~\cite{rombach2022high}, and InstructPix2Pix~\cite{brooks2023instructpix2pix}. 
The surrogate model is Stable Diffusion V1.5.}
\resizebox{\linewidth}{!}{
\begin{tabular}{c|c|ccc|cc|c|ccc|cc|c|ccc|cc}
\Xhline{0.35mm}
{Model}
& \multicolumn{6}{c|}{Stable Diffusion V1.4} 
& \multicolumn{6}{c|}{Stable Diffusion V2.0}
& \multicolumn{6}{c}{InstructPix2Pix} \\
\Xhline{0.1mm}
Method  
& clean & \cellcolor{green!15}Enc. & \cellcolor{green!15}Emb. & \cellcolor{green!15}Map & \cellcolor{green!35}Ours$_{\text{DF}}$ & \cellcolor{green!35}Ours
& clean & \cellcolor{green!15}Enc. & \cellcolor{green!15}Emb. & \cellcolor{green!15}Map & \cellcolor{green!35}Ours$_{\text{DF}}$ & \cellcolor{green!35}Ours
& clean & \cellcolor{green!15}Enc. & \cellcolor{green!15}Emb. & \cellcolor{green!15}Map & \cellcolor{green!35}Ours$_{\text{DF}}$ & \cellcolor{green!35}Ours \\
\Xhline{0.1mm}
PSNR $\downarrow$ 
& --&18.08 & 17.00 & 17.51 & \underline{14.66} & \textbf{14.23}
& --&14.33 & 14.25 & 13.89 & \underline{13.22} & \textbf{12.17}
& --&18.18 & 17.11 & 17.12 & \underline{16.06} & \textbf{15.36} \\
SSIM $\downarrow$ 
& --&0.634 & 0.467 & 0.584 & \underline{0.345} & \textbf{0.318}
& --&0.417 & 0.335 & 0.355 & \underline{0.314} & \textbf{0.240}
& --&0.571 & 0.521 & 0.498 & \underline{0.474} & \textbf{0.418} \\
VIFp $\downarrow$ 
& --&0.239 & 0.162 & 0.210 & \underline{0.091} & \textbf{0.075}
& --&0.122 & 0.099 & 0.107 & \underline{0.085} & \textbf{0.057}
& --&{0.205} & 0.207 & 0.204 & \underline{0.164} & \textbf{0.135} \\
FSIM $\downarrow$ 
& --&0.787 & 0.732 & 0.759 & \underline{0.656} & \textbf{0.641}
& --&0.654 & 0.644 & 0.624 & \underline{0.608} & \textbf{0.555}
& --&0.816 & 0.790 & 0.805 & \underline{0.760} & \textbf{0.713} \\
LPIPS $\uparrow$  
& --&0.348 & 0.492 & 0.399 & \underline{0.565} & \textbf{0.605}
& --&0.510 & 0.579 & 0.556 & \underline{0.597} & \textbf{0.660}
& --&{0.372} & 0.419 & 0.442 & \underline{0.464} & \textbf{0.527} \\
Feat. Sim. (C) $\downarrow$
& 0.743 & 0.707 & 0.690 & 0.702 & \underline{0.687} & \textbf{0.675}
& 0.694 & 0.652 & 0.647 & 0.643 & \underline{0.629} & \textbf{0.616}
& 0.810 & 0.787 & 0.756 & 0.769 & \underline{0.752} & \textbf{0.729} \\
\Xhline{0.35mm}
\end{tabular}
}
\label{black-box_universal}
\end{table}

\paragraph{\textbf{\textup{Comparison with Universalized Baselines.}}}
We first evaluate our method against the universal extensions of existing state-of-the-art image-specific immunization approaches.
As shown in Table~\ref{white-box_universal}, our method consistently outperforms these baselines by a clear margin.
Notably, even our data-free variant achieves competitive results, ranking second-best across most metrics, where it highlights the effectiveness and practicality of our approach.


We further evaluate our UAP's transferability across editing models, including Stable Diffusion V1.4, V2.0~\cite{rombach2022high}, {and InstructPix2Pix~\cite{brooks2023instructpix2pix}.}
As shown in Table~\ref{black-box_universal}, our method consistently outperforms universalized baselines in the black-box setting.
In particular, higher LPIPS scores and lower Feat. Sim.(C) indicate substantial perceptual and semantic shifts, likely caused by our semantic injection, which steers the editing process to be more conditioned on the injected target rather than the original content.
These results across models highlight the strong transferability of our semantic injection approach.
\begin{table}[t]
\setlength{\tabcolsep}{3pt}
\centering
\caption{Robustness evaluation against various purification methods. 
}
\resizebox{\linewidth}{!}{
\begin{tabular}{c|c|ccc|c|ccc|c|ccc|c|ccc|c}
\Xhline{0.35mm}
{Purification}
&\multirow{2}{*}{\makecell{Clean \\ Sim.}}
& \multicolumn{4}{c|}{JPEG compression~\cite{jpeg}} 
& \multicolumn{4}{c|}{GrIDPure~\cite{gridpure}}
& \multicolumn{4}{c|}{Conditional DiffPure~\cite{diffpure}}
& \multicolumn{4}{c}{Noisy Upscaling~\cite{diffpure}}\\
\Xcline{1-1}{0.1mm}
\Xcline{3-18}{0.1mm}
Method  
&  & \cellcolor{green!15}Enc. & \cellcolor{green!15}Emb. & \cellcolor{green!15}Map & \cellcolor{green!35}Ours
& \cellcolor{green!15}Enc. & \cellcolor{green!15}Emb. & \cellcolor{green!15}Map & \cellcolor{green!35}Ours
& \cellcolor{green!15}Enc. & \cellcolor{green!15}Emb. & \cellcolor{green!15}Map & \cellcolor{green!35}Ours
& \cellcolor{green!15}Enc. & \cellcolor{green!15}Emb. & \cellcolor{green!15}Map & \cellcolor{green!35}Ours\\
\Xhline{0.1mm}
PSNR $\downarrow$ & --& 17.56 & 16.59 & 17.31 & \textbf{15.88}
                  & 18.83 & 18.51 & 18.64 & \textbf{17.90}
                  & 16.64 & 16.54 & 16.40 & \textbf{14.99}
                  & 18.22 & 17.99 & 17.90 & \textbf{17.20} \\
SSIM $\downarrow$ &-- &0.541 & 0.454 & {0.526} &  \textbf{0.433}
                  & 0.593 & 0.575 & 0.585 & \textbf{0.545}
                  & 0.524 & 0.462 & 0.475 & \textbf{0.371}
                  & 0.558 & 0.551 & 0.548 & \textbf{0.482} \\
VIFp $\downarrow$ & --&0.198 & 0.148 & 0.187 & \textbf{0.129}
                  & 0.224 & 0.220 & 0.226 & \textbf{0.197}
                  & 0.173 & 0.148 & 0.165 & \textbf{0.102}
                  & 0.195 & 0.195 & 0.190 & \textbf{0.165} \\
FSIM $\downarrow$ &-- &0.750 & 0.717 & 0.741 & \textbf{0.697}
                  & 0.778 & 0.773 & 0.775 & \textbf{0.758}
                  & 0.733 & 0.717 & 0.718 & \textbf{0.664}
                  & 0.762 & 0.760 & 0.757 & \textbf{0.732} \\
LPIPS $\uparrow$  &-- &0.406 & 0.478 & 0.421 & \textbf{0.511}
                  & 0.373 & 0.382 &0.377 & \textbf{0.410}
                  & 0.423 & 0.466 & 0.455 & \textbf{0.568}
                  & 0.383 & 0.384 & 0.389 & \textbf{0.427} \\
Feat. Sim. (C) $\downarrow$
& 0.744 & 0.710 & 0.708 & 0.713 & \textbf{0.698} 
&  0.727 & 0.723 & 0.724 & \textbf{0.717}
&  0.707 & 0.707 & 0.701 & \textbf{0.683}
&  0.728 & 0.730 & 0.730 & \textbf{0.723} \\

\Xhline{0.35mm}
\end{tabular}
}
\label{robustness}
\end{table}
\begin{table}[t]
\footnotesize
\centering
\caption{Comparison with image-specific and optimization-free methods in the white-box setting on SD V1.5.
    \textit{Ours} denotes the full method, while \textit{Ours\textsubscript{DF}} uses only the target semantic injection loss in a data-free setting.}
\resizebox{\linewidth}{!}{
\begin{tabular}{c|c|c|ccc|c|cc}
\Xhline{0.35mm}
\multicolumn{2}{c|}{\multirow{2}{*}{Method}} 
& \multirow{2}{*}{Clean}
& \multicolumn{3}{>{\columncolor{gray!15}}c|}{\textit{Image-specific}} 
& \multicolumn{1}{>{\columncolor{cyan!15}}c|}{\textit{Opt-free}}
& \multicolumn{2}{>{\columncolor{green!35}}c}{\textit{Universal}} \\
\multicolumn{2}{c|}{} 
& 
& \cellcolor{gray!15}EA~\cite{raising} 
& \cellcolor{gray!15}DA~\cite{raising} 
& \cellcolor{gray!15}SA~\cite{semantic} 
& \cellcolor{cyan!15}FP~\cite{nearly} 
& \cellcolor{green!35}Ours$_{\text{DF}}$ 
& \cellcolor{green!35}Ours \\
\Xhline{0.1mm}
\multicolumn{2}{c|}{Input adaptation}
& --& \ding{51}  & \ding{51}  & \ding{51} & \ding{51}  & \ding{55} & \ding{55} \\
\Xhline{0.1mm}
\multirow{7}{*}{Metrics}

& PSNR $\downarrow$
& --& 14.75{\scriptsize$\pm$0.715}
& 14.71{\scriptsize$\pm$0.951}
& 16.28{\scriptsize$\pm$1.547}
& 16.44{\scriptsize$\pm$1.207}
& \underline{14.68}{\scriptsize$\pm$0.035}
& \textbf{14.19}{\scriptsize$\pm$0.059} \\

& SSIM $\downarrow$
& --& 0.386{\scriptsize$\pm$0.059}
& 0.382{\scriptsize$\pm$0.082}
& 0.431{\scriptsize$\pm$0.102}
& 0.472{\scriptsize$\pm$0.013}
& 0.378{\scriptsize$\pm$0.002}
& \textbf{0.332}{\scriptsize$\pm$0.003} \\

& VIFp $\downarrow$
& --& \underline{0.088}{\scriptsize$\pm$0.025}
& 0.106{\scriptsize$\pm$0.037}
& 0.138{\scriptsize$\pm$0.052}
& 0.162{\scriptsize$\pm$0.009}
& 0.106{\scriptsize$\pm$0.001}
& \textbf{0.082}{\scriptsize$\pm$0.001} \\

& FSIM $\downarrow$
& --&\textbf{0.637}{\scriptsize$\pm$0.023}
& 0.660{\scriptsize$\pm$0.037}
& 0.714{\scriptsize$\pm$0.051}
& 0.710{\scriptsize$\pm$0.010}
& 0.666{\scriptsize$\pm$0.002}
& \underline{0.642}{\scriptsize$\pm$0.003} \\

& LPIPS $\uparrow$
& --&\underline{0.584}{\scriptsize$\pm$0.034}
& 0.552{\scriptsize$\pm$0.050}
& 0.490{\scriptsize$\pm$0.064}
& 0.451{\scriptsize$\pm$0.021}
& 0.557{\scriptsize$\pm$0.002}
& \textbf{0.606}{\scriptsize$\pm$0.003} \\

& Feat. Sim. (C) $\downarrow$
& 0.744 & 0.677{\scriptsize$\pm$0.002}
& \textbf{0.662}{\scriptsize$\pm$0.002}
& 0.705{\scriptsize$\pm$0.002}
& 0.702{\scriptsize$\pm$0.005}
& 0.685{\scriptsize$\pm$0.003}
& \underline{0.673}{\scriptsize$\pm$0.002} \\

\Xhline{0.1mm}
\multirow{3}{*}{Imperceptibility}
& DISTS $\downarrow$ 
& --& 0.311 & 0.243 & \textbf{0.164} & 0.216 &\underline{0.170} & {0.181} \\

& PSNR $\uparrow$
& --&28.67 & 27.78 & \underline{30.82} & {30.35} & \textbf{31.15} & 29.41 \\

& LPIPS $\downarrow$
& --&0.471 & 0.428 & \textbf{0.301} & 0.360 & \underline{0.316} & {0.355} \\

\Xhline{0.1mm}
\multirow{2}{*}{\makecell{Test-Time \\ Cost (per image)}}
& Latency CPU/GPU (sec) 
& --&315.71/8.01 & 2706.84/212.46 & 2216.13/55.66 & 2.76/0.04 & $\sim$0/$\sim$0 & $\sim$0/$\sim$0 \\

& GPU Usage (GB)
& --&6.51 & 30.84 & 9.08 & 0.77 & 0 & 0 \\

\Xhline{0.35mm}
\end{tabular}
}
\label{tab:white-box_specific}
\end{table}
\paragraph{\textbf{\textup{Robustness Evaluation against Purification.}}}
We evaluate the robustness of our method and universalized baselines against perturbation purification techniques.
We adapt four widely used defenses for purifying adversarial perturbations, including JPEG compression~\cite{jpeg}, GrIDPure~\cite{gridpure}, Conditional DiffPure, and Noisy Upscaling~\cite{diffpure}.
{Importantly, diffusion-based purification methods such as Conditional DiffPure and Noisy Upscaling act as adaptive defenses~\cite{adaptive} against our approach. 
Specifically, while our method injects target semantics into the original image, these methods attempt to restore the original semantics by reconstructing or upscaling the content using the original image prompt. 
Thus, they can be viewed as adaptive strategies that assume awareness of our immunization mechanism.
}
As shown in Table~\ref{robustness}, our carefully designed semantic injection approach consistently outperforms the universalized baselines after purification, demonstrating its robustness against defenses.

\paragraph{\textbf{\textup{Comparison with Image-specific Methods.}}}
We compare our method with image-specific and optimization-free immunization methods on white-box scenario, evaluating performance, imperceptibility, and test-time cost in terms of effectiveness and efficiency.
As shown in Table~\ref{tab:white-box_specific}, our method significantly outperforms image-specific methods across most metrics, while also achieving better imperceptibility, despite operating in the more challenging universal setting.
Surprisingly, even under stricter constraints, such as the absence of real data and a smaller perturbation budget, Ours$_{\text{DF}}$ achieves competitive performance compared to existing image-specific methods, and further surpasses FP~\cite{nearly} despite comparable imperceptibility.

To validate robustness to unseen models, we conduct black-box experiments in Table~\ref{black-box}.
The results also show that our method outperforms the optimization-free method and image-specific methods in most metrics, often surpassing them across various diffusion models~\cite{rombach2022high, brooks2023instructpix2pix}.
Importantly, while image-specific methods~\cite{raising, semantic} rely on larger perturbation budgets and GPU-intensive image-aware fine-grained optimization, our method can serve as a practical alternative by offering reliable transferability with better imperceptibility and near-zero inference cost.


\begin{table}[!t]
\footnotesize
\setlength{\tabcolsep}{3pt}
\centering
\caption{Black-box immunization transferability on Stable Diffusion V1.4, V2.0~\cite{rombach2022high}, and InstructPix2Pix~\cite{brooks2023instructpix2pix}. 
The surrogate model for each method is Stable Diffusion V1.5.}
\label{black-box}
\resizebox{\linewidth}{!}{
\begin{tabular}{c|c|ccc|c|cc|c|ccc|c|cc|c|ccc|c|cc}
\Xhline{0.35mm}
{Model}
& \multicolumn{7}{c|}{Stable Diffusion V1.4} 
& \multicolumn{7}{c|}{Stable Diffusion V2.0}
& \multicolumn{7}{c}{InstructPix2Pix} \\
\Xhline{0.1mm}
\multirow{2}{*}{Method}  
& \multirow{2}{*}{Clean}
& \multicolumn{3}{>{\columncolor{gray!15}}c|}{\textit{Image-specific}}  
& \multicolumn{1}{>{\columncolor{cyan!15}}c|}{\textit{Opt-free}}
& \multicolumn{2}{>{\columncolor{green!35}}c|}{\textit{Universal}}   
& \multirow{2}{*}{Clean}
& \multicolumn{3}{>{\columncolor{gray!15}}c|}{\textit{Image-specific}}  
& \multicolumn{1}{>{\columncolor{cyan!15}}c|}{\textit{Opt-free}}
& \multicolumn{2}{>{\columncolor{green!35}}c|}{\textit{Universal}}   
& \multirow{2}{*}{Clean}
& \multicolumn{3}{>{\columncolor{gray!15}}c|}{\textit{Image-specific}}  
& \multicolumn{1}{>{\columncolor{cyan!15}}c|}{\textit{Opt-free}}
& \multicolumn{2}{>{\columncolor{green!35}}c}{\textit{Universal}}          \\
& & \cellcolor{gray!15}EA & \cellcolor{gray!15}DA & \cellcolor{gray!15}SA & \cellcolor{cyan!15}FP & \cellcolor{green!35}Ours$_{\text{DF}}$ & \cellcolor{green!35}Ours
& & \cellcolor{gray!15}EA & \cellcolor{gray!15}DA & \cellcolor{gray!15}SA & \cellcolor{cyan!15}FP & \cellcolor{green!35}Ours$_{\text{DF}}$ & \cellcolor{green!35}Ours
& & \cellcolor{gray!15}EA & \cellcolor{gray!15}DA & \cellcolor{gray!15}SA & \cellcolor{cyan!15}FP & \cellcolor{green!35}Ours$_{\text{DF}}$ & \cellcolor{green!35}Ours \\
\Xhline{0.1mm}

PSNR $\downarrow$ 
& -- & 15.46 & 15.54 & \underline{14.91} & 17.93 & {15.60} & \textbf{14.18}
& -- & \underline{12.48} & 12.73 & 12.79 & 14.01 & {13.22} & \textbf{12.17}
& -- & 16.02 & \underline{15.89} & 17.26 & 17.96 & {16.06} & \textbf{15.36} \\

SSIM $\downarrow$ 
& -- & 0.444 & 0.457 & \underline{0.336} & 0.584 & 0.467 & \textbf{0.332}
& -- & \underline{0.298} & 0.304 & 0.271 & 0.365 & {0.314} & \textbf{0.240}
& -- & 0.445 & \textbf{0.409} & 0.499 & 0.503 & {0.474} & \underline{0.418} \\

VIFp $\downarrow$ 
& -- & 0.119 & 0.143 & \underline{0.088} & 0.223 & {0.146} & \textbf{0.083}
& -- & 0.074 & 0.085 & \underline{0.060} & 0.118 & {0.085} & \textbf{0.057}
& -- & \textbf{0.133} & \underline{0.134} & 0.189 & 0.171 & {0.164} & 0.135 \\

FSIM $\downarrow$ 
& -- & \underline{0.664} & 0.695 & 0.666 & 0.763 & {0.702} & \textbf{0.642}
& -- & \underline{0.558} & 0.586 & 0.632 & 0.626 & {0.608} & \textbf{0.555}
& -- & \underline{0.747} & 0.755 & 0.795 & 0.788 & {0.760} & \textbf{0.713} \\

LPIPS $\uparrow$  
& -- & 0.545 & 0.507 & \underline{0.549} & 0.381 & {0.509} & \textbf{0.604}
& -- & \underline{0.634} & 0.611 & 0.626 & 0.533 & {0.597} & \textbf{0.660}
& -- & 0.509 & \textbf{0.533} & 0.433 & 0.433 & {0.464} & \underline{0.527} \\

Feat. Sim. (C) $\downarrow$
& 0.743 & \textbf{0.663} & 0.675 & 0.701 & 0.699 & 0.687 & \underline{0.675}
& 0.694 & \textbf{0.606} & 0.628 & 0.656 & 0.648 & 0.629 & \underline{0.616}
& 0.810 & 0.774 & 0.788 & 0.773 & 0.782 & \underline{0.752} & \textbf{0.729} \\

\Xhline{0.35mm}
\end{tabular}
}
\end{table}
\begin{table}[t]
\footnotesize
\setlength{\tabcolsep}{3pt}
\centering
\caption{Evaluating performance on ImageNet-Edit~\cite{blurguard} compared with
universalized baselines. The best result is shown in bold; the second-best is underlined.}
\resizebox{\linewidth}{!}{
\begin{tabular}{c|c|ccc|c|c|ccc|c|c|ccc|c|c|ccc|c}
\Xhline{0.35mm}
{Model}
& \multicolumn{5}{c|}{Stable Diffusion V1.4} 
& \multicolumn{5}{c|}{Stable Diffusion V1.5*} 
& \multicolumn{5}{c|}{Stable Diffusion V2.0}
& \multicolumn{5}{c}{InstructPix2Pix} \\
\Xhline{0.1mm}
Method  
& clean & \cellcolor{green!15}Enc. & \cellcolor{green!15}Emb. & \cellcolor{green!15}Map & \cellcolor{green!35}Ours
& clean & \cellcolor{green!15}Enc. & \cellcolor{green!15}Emb. & \cellcolor{green!15}Map & \cellcolor{green!35}Ours 
& clean & \cellcolor{green!15}Enc. & \cellcolor{green!15}Emb. & \cellcolor{green!15}Map & \cellcolor{green!35}Ours 
& clean & \cellcolor{green!15}Enc. & \cellcolor{green!15}Emb. & \cellcolor{green!15}Map & \cellcolor{green!35}Ours \\
\Xhline{0.1mm}

PSNR $\downarrow$ 
& --&16.74 & 16.38 & \underline{16.18} & \textbf{14.76}
& --&16.68 & 16.40 & \underline{16.63} & \textbf{14.71}
& --&13.73 & 14.31 & \underline{13.43} & \textbf{12.92}
& --&17.57 & \underline{15.59} & 17.08 & \textbf{13.82} \\

SSIM $\downarrow$ 
& --&0.582 & \underline{0.426} & 0.527 & \textbf{0.387}
& --&0.580 & \underline{0.423} & 0.535 & \textbf{0.376}
& --&0.396 & \underline{0.330} & 0.339 & \textbf{0.290}
& --&0.584 & \underline{0.505} & 0.545 & \textbf{0.440} \\

VIFp $\downarrow$ 
& --&0.183 & \underline{0.131} & 0.158 & \textbf{0.101}
& --&0.182 & \underline{0.129} & 0.162 & \textbf{0.097}
& --&0.110 & \underline{0.093} & 0.097 & \textbf{0.076}
& --&0.186 & \underline{0.166} & 0.168 & \textbf{0.123} \\

FSIM $\downarrow$ 
& --&0.748 & \underline{0.710} & 0.715 & \textbf{0.652}
& --&0.746 & \underline{0.711} & 0.720 & \textbf{0.650}
& --&0.624 & 0.634 & \underline{0.597} & \textbf{0.583}
& --&0.799 & \underline{0.751} & 0.779 & \textbf{0.704} \\

LPIPS $\uparrow$  
& --&0.424 & \underline{0.534} & 0.476 & \textbf{0.603}
& --&0.428 & \underline{0.535} & 0.470 & \textbf{0.609}
& --&0.551 & \underline{0.592} & 0.586 & \textbf{0.646}
& --&0.407 &\underline{0.473} & 0.448 & \textbf{0.541} \\

Feat. Sim. (C) $\downarrow$
& 0.714 &0.688 & \underline{0.678} & 0.682 & \textbf{0.660}
& 0.712 & 0.683 & \underline{0.674} & 0.684 & \textbf{0.661} 
& 0.677 &0.638 & 0.627 & \underline{0.626} & \textbf{0.605}
& 0.803 &0.764 & \underline{0.728} & 0.745 & \textbf{0.706} \\


\Xhline{0.35mm}
\end{tabular}
}
\label{imagenet-uap}
\end{table}

\begin{table}[t]
\footnotesize
\centering
\caption{Evaluating effectiveness of our method on ImageNet-Edit~\cite{blurguard} compared with image-specific and optimization free methods in the white-box setting on Stable Diffusion V1.5.
The best result is shown in bold; the second-best is underlined.}
\resizebox{\linewidth}{!}{
\begin{tabular}{c|c|c|ccc|c|c}
\Xhline{0.35mm}
\multicolumn{2}{c|}{\multirow{2}{*}{Method}} & \multirow{2}{*}{Clean}
& \multicolumn{3}{c|}{\cellcolor{gray!15}\textit{Image-specific}}
& \multicolumn{1}{c|}{\cellcolor{cyan!15}\textit{Opt-free}}
& \multicolumn{1}{c}{\cellcolor{green!35}\textit{Universal}} \\
\multicolumn{2}{c|}{} &
& \cellcolor{gray!15}EA & \cellcolor{gray!15}DA & \cellcolor{gray!15}SA & \cellcolor{cyan!15}FP & \cellcolor{green!35}Ours \\
\Xhline{0.1mm}

\multicolumn{2}{c|}{Input adaptation}
& -- & \ding{51}  & \ding{51}  & \ding{51} & \ding{51}  & \ding{55}  \\
\Xhline{0.1mm}

\multirow{7}{*}{Metrics}
& PSNR $\downarrow$            & -- 
& \underline{14.71}{\scriptsize$\pm$0.101} 
& \textbf{14.50}{\scriptsize$\pm$0.130} 
& 16.72{\scriptsize$\pm$0.128}  
& 16.76 {\scriptsize$\pm$0.216} 
& \underline{14.71}{\scriptsize$\pm$0.064} \\

& SSIM $\downarrow$            & -- 
& 0.451{\scriptsize$\pm$0.001} 
& \underline{0.441}{\scriptsize$\pm$0.008} 
& 0.466{\scriptsize$\pm$0.008} 
& 0.529 {\scriptsize$\pm$0.009} 
& \textbf{0.376}{\scriptsize$\pm$0.005}  \\

& VIFp $\downarrow$            & -- 
& \textbf{0.093}{\scriptsize$\pm$0.002} 
& 0.125{\scriptsize$\pm$0.002} 
& 0.149{\scriptsize$\pm$0.002} 
& 0.183{\scriptsize$\pm$0.010} 
& \underline{0.097}{\scriptsize$\pm$0.001}  \\

& FSIM $\downarrow$            & -- 
& \textbf{0.638}{\scriptsize$\pm$0.004} 
& 0.668{\scriptsize$\pm$0.006} 
& 0.737{\scriptsize$\pm$0.007} 
& 0.719{\scriptsize$\pm$0.007} 
& \underline{0.650}{\scriptsize$\pm$0.005}  \\

& LPIPS $\uparrow$             & -- 
& \underline{0.605}{\scriptsize$\pm$0.010} 
& 0.544{\scriptsize$\pm$0.007} 
& 0.484{\scriptsize$\pm$0.004} 
& 0.439{\scriptsize$\pm$0.014} 
& \textbf{0.609}{\scriptsize$\pm$0.004}  \\

& Feat. Sim. (C) $\downarrow$  & 0.714 
& \textbf{0.652}{\scriptsize$\pm$0.002} 
& 0.664{\scriptsize$\pm$0.004} 
& 0.678{\scriptsize$\pm$0.003} 
& 0.679{\scriptsize$\pm$0.007} 
& \underline{0.661}{\scriptsize$\pm$0.004} \\


\Xhline{0.1mm}
\multirow{3}{*}{Imperceptibility}
& DISTS $\downarrow$           & -- 
& 0.350 
& 0.300 
& \textbf{0.201} 
& 0.251 
& \underline{0.215} \\
& PSNR $\uparrow$              & -- 
& 27.20 
& 26.87 
& \textbf{30.91} 
& \underline{30.03} 
& 29.84 \\
& LPIPS $\downarrow$           & -- 
& 0.515 
& 0.475 
& \textbf{0.358} 
& \underline{0.379} 
& 0.383 \\

\Xhline{0.1mm}
\multirow{2}{*}{\makecell{Test-Time\\Cost (per image)}}
& Latency CPU/GPU (sec)                & -- 
& 315.71/8.01  
& 2706.84/212.46 
& 2216.13/55.66 
& 2.76/0.04 
& $\sim$0/$\sim$0 \\
& GPU Usage (GB)               & -- 
& 6.51  
& 30.84 
& 9.08  
& 0.77 
& 0 \\

\Xhline{0.35mm}
\end{tabular}
}
\label{tab:imagenet-edit-main}
\end{table}
\begin{table}[t!]
\footnotesize
\setlength{\tabcolsep}{3pt}
\centering
\caption{Black-box immunization performance on ImageNet-Edit compared with image-specific and optimization free methods. The best result is shown in bold; the second-best is underlined.}
\resizebox{\linewidth}{!}{
\begin{tabular}{c|c|ccc|c|c|c|ccc|c|c|c|ccc|c|c}
\Xhline{0.35mm}
{Model}
& \multicolumn{6}{c|}{Stable Diffusion V1.4} 
& \multicolumn{6}{c|}{Stable Diffusion V2.0}
& \multicolumn{6}{c}{InstructPix2Pix} \\
\Xhline{0.1mm}

\multicolumn{1}{c|}{\multirow{2}{*}{Method}}
& \multirow{2}{*}{clean}
& \multicolumn{3}{c|}{\cellcolor{gray!15}\textit{Image-specific}}
& \multicolumn{1}{c|}{\cellcolor{cyan!15}\textit{Opt-free}}
& \multicolumn{1}{c|}{\cellcolor{green!35}\textit{Universal}}

& \multirow{2}{*}{clean}
& \multicolumn{3}{c|}{\cellcolor{gray!15}\textit{Image-specific}}
& \multicolumn{1}{c|}{\cellcolor{cyan!15}\textit{Opt-free}}
& \multicolumn{1}{c|}{\cellcolor{green!35}\textit{Universal}}
                      
& \multirow{2}{*}{clean}
& \multicolumn{3}{c|}{\cellcolor{gray!15}\textit{Image-specific}}
& \multicolumn{1}{c|}{\cellcolor{cyan!15}\textit{Opt-free}}
& \multicolumn{1}{c}{\cellcolor{green!35}\textit{Universal}} \\

&  & \cellcolor{gray!15}EA & \cellcolor{gray!15}DA & \cellcolor{gray!15}SA & \cellcolor{cyan!15}FP & \cellcolor{green!35}Ours
&  & \cellcolor{gray!15}EA & \cellcolor{gray!15}DA & \cellcolor{gray!15}SA & \cellcolor{cyan!15}FP & \cellcolor{green!35}Ours
&  & \cellcolor{gray!15}EA & \cellcolor{gray!15}DA & \cellcolor{gray!15}SA & \cellcolor{cyan!15}FP & \cellcolor{green!35}Ours \\

\Xhline{0.1mm}

PSNR $\downarrow$ 
& -- & \underline{14.58} & \textbf{14.44} & 16.82 & 16.84 & 14.76
& -- & \underline{12.69} & \textbf{12.30} & 14.95 & 13.44 & 12.92
& -- & \underline{14.33} & 14.99 & 16.04 & 17.66 & \textbf{13.82} \\

SSIM $\downarrow$ 
& -- & 0.451 & \underline{0.437} & 0.476 & 0.537 & \textbf{0.387}
& -- & 0.344 & \underline{0.304} & 0.389 & 0.339 & \textbf{0.290}
& -- & \underline{0.479} & 0.460 & 0.535 & 0.538 & \textbf{0.440} \\

VIFp $\downarrow$ 
& -- & \textbf{0.095} & 0.121 & 0.154 & 0.184 & \underline{0.101}
& -- & \textbf{0.072} & 0.081 & 0.109 & 0.104 & \underline{0.076}
& -- & \textbf{0.101} & \underline{0.120} & 0.171 & 0.166 & 0.123 \\

FSIM $\downarrow$ 
& -- & \textbf{0.633} & 0.661 & 0.742 & 0.722 & \underline{0.652}
& -- & \textbf{0.553} & \underline{0.566} & 0.667 & 0.590 & 0.583
& -- & \textbf{0.692} & 0.733 & 0.770 & 0.776 & \underline{0.704} \\

LPIPS $\uparrow$  
& -- & \underline{0.602} & 0.551 & 0.478 & 0.432 & \textbf{0.603}
& -- & \textbf{0.661} & 0.628 & 0.552 & 0.568 & \underline{0.646}
& -- & \textbf{0.549} & 0.519 & 0.434 & 0.451 & \underline{0.541} \\

Feat. Sim. (C) $\downarrow$
& 0.714 & \textbf{0.652} & 0.663 & 0.689 & 0.677 & \underline{0.660}
& 0.677 & \textbf{0.595} & 0.623 & 0.641 & 0.635 & \underline{0.605}
& 0.803 & \textbf{0.700} & 0.725 & 0.737 & 0.781 & \underline{0.706} \\


\Xhline{0.35mm}
\end{tabular}
}
\label{imagenet-specific-bb}
\end{table}
\paragraph{\textbf{\textup{Evaluation on Real Images.}}}
We evaluate white-box and black-box immunization on the real-image ImageNet-Edit dataset~\cite{blurguard} to assess transferability across diverse diffusion models compared with universalized baselines(\ie Enc., Emb., Map).
As shown in Table~\ref{imagenet-uap}, our method, which is carefully designed for the UAP setting rather than naively obtained by scaling up training data from image-specific baselines, consistently outperforms universalized baselines.

Tables~\ref{tab:imagenet-edit-main} and~\ref{imagenet-specific-bb} further confirm that our UAP-based immunization framework provides competitive or superior protection on real images compared to image-specific methods in the white-box and black-box settings, respectively, while maintaining practicality and efficiency with near-zero inference cost.

\paragraph{\textbf{\textup{Limitations and Discussions.}}}
A potential concern is reverse-engineering of protections. However, our method generates diverse UAPs even for the same prompt, producing varied protection patterns that hinder reverse-engineering, as supported by Table~\ref{target}. Further analysis of limitations and discussions is provided in the supplementary material (Sections~\textcolor{red}{E} and~\textcolor{red}{F}).

\subsection{Ablation Studies}
\paragraph{\textbf{\textup{Impact of Each Proposed Components.}}}
Table~\ref{ablation} shows immunization performance under three settings to assess the effects of real data and our proposed losses: \texttt{Inj$_{\text{DF}}$} (target semantic injection, data-free), \texttt{Inj} (target semantic injection with real data), and \texttt{Inj+Sup} (additional source semantic suppression). The improvement from \texttt{Inj$_{\text{DF}}$} to \texttt{Inj} shows the benefit of real data, while suppression further improves robustness and transferability by reducing residual source semantics. Overall, the results highlight the complementary roles of target semantic injection, real data, and semantic disentanglement in black-box immunization. Stable Diffusion V1.5 is used as the surrogate model for ablation.

\begin{table}[t]
\footnotesize
\setlength{\tabcolsep}{3pt}
\centering
\caption{Ablation study for our proposed method. 
\texttt{Inj}$_{\text{\texttt{DF}}}$, \texttt{Inj}, and \texttt{Inj+Sup} denote results obtained using UAPs trained with $\mathcal{L}_\text{inj}$ in the data-free setting, data-dependent setting, and the combination of $\mathcal{L}_\text{inj}$ and $\mathcal{L}_\text{sup}$, respectively.
* refers the white-box result.}
\resizebox{\linewidth}{!}{
\begin{tabular}{c|c|ccc|c|ccc|c|ccc|c|ccc}
\Xhline{0.35mm}
{Model}
& \multicolumn{4}{c|}{Stable Diffusion V1.4}
& \multicolumn{4}{c|}{Stable Diffusion V1.5*} 
& \multicolumn{4}{c|}{Stable Diffusion V2.0}
& \multicolumn{4}{c}{InstructPix2Pix}\\
\Xhline{0.1mm}
{Method}
& Clean&\texttt{Inj}$_{\text{\texttt{DF}}}$ &\texttt{Inj}&\texttt{Inj+Sup}  
& Clean&\texttt{Inj}$_{\text{\texttt{DF}}}$ &\texttt{Inj}&\texttt{Inj+Sup}
& Clean&\texttt{Inj}$_{\text{\texttt{DF}}}$ &\texttt{Inj}&\texttt{Inj+Sup} 
& Clean&\texttt{Inj}$_{\text{\texttt{DF}}}$ &\texttt{Inj}&\texttt{Inj+Sup}  \\
\Xhline{0.1mm}
PSNR $\downarrow$ &--& 14.66 & 14.33 & 14.23 
                  &--& 14.68 & 14.41 & 14.19 
                  &--& 12.33 & 12.13 & 12.04     
                  &--& 16.06 & 15.61 & 15.36 \\  
SSIM $\downarrow$ &--& 0.345 & 0.332 & 0.318
                  &--& 0.378 & 0.367 & 0.332
                  &--& 0.253 & 0.245 & 0.235
                  &--& 0.474 & 0.433 & 0.418 \\
VIFp $\downarrow$ &--& 0.091 & 0.080 & 0.075
                  &--& 0.106 & 0.096 & 0.082
                  &--& 0.058 & 0.053 & 0.051
                  &--& 0.164 & 0.141 & 0.135 \\
FSIM $\downarrow$ &--& 0.656 & 0.646 & 0.641
                  &--& 0.666 & 0.655 & 0.642
                  &--& 0.573 & 0.567 & 0.564
                  &--& 0.760 & 0.719 & 0.713 \\
LPIPS $\uparrow$  &--& 0.565 & 0.591 & 0.605
                  &--& 0.557 & 0.585 & 0.606
                  &--& 0.643 & 0.658 & 0.665
                  &--& 0.464 & 0.514 & 0.527 \\
Feat. Sim. (C) $\downarrow$
& 0.743& 0.687 & 0.681 & 0.675
& 0.744& 0.685 & 0.680 & 0.673
& 0.694& 0.629 & 0.620 & 0.616
& 0.810& 0.752 & 0.731 & 0.729 \\

\Xhline{0.35mm}
\end{tabular}
}
\label{ablation}
\end{table}

\begin{table}[t]
\centering
\caption{Target selection ablation. * denotes the prompt used in the main experiments.}
\resizebox{0.95\linewidth}{!}{
\begin{tabular}{c|c|ccccc|cc}
\Xhline{0.35mm}
Target & Clean & Ronaldo & Tiger* & Sunflower & Peacock & Mandala & Avg. & Std. \\
\Xhline{0.1mm}
PSNR $\downarrow$ & --&14.69{\tiny$\pm$0.033} & 14.19{\tiny$\pm$0.058} & 14.32{\tiny$\pm$0.044} & 14.22{\tiny$\pm$0.055} & 14.25{\tiny$\pm$0.033} & 14.33 & 0.199 \\
SSIM $\downarrow$ & --&0.383{\tiny$\pm$0.002} & 0.332{\tiny$\pm$0.003} & 0.340{\tiny$\pm$0.002} & 0.308{\tiny$\pm$0.003} & 0.301{\tiny$\pm$0.003} & 0.332 & 0.029 \\
VIFp $\downarrow$ & --&0.095{\tiny$\pm$0.001} & 0.082{\tiny$\pm$0.001} & 0.088{\tiny$\pm$0.001} & 0.092{\tiny$\pm$0.001} & 0.078{\tiny$\pm$0.001} & 0.087 & 0.006 \\
FSIM $\downarrow$ & --&0.660{\tiny$\pm$0.001} & 0.642{\tiny$\pm$0.002} & 0.651{\tiny$\pm$0.002} & 0.639{\tiny$\pm$0.001} & 0.641{\tiny$\pm$0.001} & 0.646 & 0.008 \\
LPIPS $\uparrow$  & --&0.576{\tiny$\pm$0.002} & 0.606{\tiny$\pm$0.003} & 0.610{\tiny$\pm$0.003} & 0.598{\tiny$\pm$0.003} & 0.621{\tiny$\pm$0.003} & 0.602 & 0.015 \\
Feat. Sim. (C) $\downarrow$
& 0.744 & 0.677{\tiny$\pm$0.003} & 0.673{\tiny$\pm$0.002} & 0.679{\tiny$\pm$0.001} & 0.681{\tiny$\pm$0.002} & 0.673{\tiny$\pm$0.002} & 0.676 & 0.004 \\

\Xhline{0.35mm}
\end{tabular}
}
\label{target}

\end{table}
\paragraph{\textbf{\textup{Diverse Target Analysis.}}}
To demonstrate the target-independency of our method, we arbitrarily select five diverse targets, \eg, \textit{`Ronaldo', `Tiger', `Sunflower', `Peacock'}, and \textit{`Mandala'}, and evaluate performance across them.
As shown in Table~\ref{target}, our method consistently achieves strong defense performance with low variance, indicating robustness to target choice. We provide visualizations of the generated UAPs and qualitative editing results for each target in the supplementary material (Sections~\textcolor{red}{A.1} and~\textcolor{red}{D.1}).

\section{Conclusion}
We introduced {the first} universal image immunization framework that defends against diffusion-based editing by injecting target semantics through a targeted UAP.
By incorporating two optimization objectives--target semantic injection and source semantic suppression--the method effectively overrides the source content while suppressing the original semantics, guiding edits {conditioned on} the injected target. 
As it requires only a simple additive operation at test time, the method enables highly efficient and model-agnostic deployment. 
Our method significantly outperforms all universalized baselines adapted from prior image-specific methods, remains effective in data-free settings, 
and demonstrates strong competitiveness against image-specific and optimization-free approaches, markedly reducing the performance--cost tension and underscoring its practical applicability.

\subsection*{Acknowledgement}
This work was supported by AI Graduate School Program at POSTECH (RS-2019-II191906 (5\%)), the NRF grants (RS-2025-24535146 (20\%), RS-2026-25491789 (50\%)) and the IITP grants (RS-2022-II220926 (25\%)) funded by MSIT, Korea.







%
%
\bibliographystyle{splncs04}
\bibliography{main}
\newpage
\appendix
\setcounter{figure}{3}
\setcounter{table}{6}
\setcounter{equation}{6}
\begin{center}
  {\bfseries\LARGE Universal Image Immunization against Diffusion-based Image Editing via  Semantic Injection}\\[2mm]
  \vspace{-2mm}
  {\bfseries\large -\textit{Supplementary Materials}-}
  \vspace{-2mm}
\end{center}

\startcontents[sections]
\printcontents[sections]{l}{1}{\setcounter{tocdepth}{2}}
\clearpage

\section{Experimental Details}
\label{exp}

\subsection{Datasets and Evaluation Protocols}
\label{app:eval}
\noindent
\paragraph{\textbf{\textup{Evaluation Dataset for Image Editing (${D}_E$).}}} 
{As mentioned in Section~\textcolor{red}{5} of the main manuscript, we constructed evaluation datasets for image editing, by following the setup of prior work~\cite{raising, semantic}. 
Specifically, we generated 50 samples for each of 10 distinct object classes using Stable Diffusion V3~\cite{sd3}. 
Compared to the previous setup~\cite{semantic}, which used only 3 object classes (50$\times$3 classes = 150 images), our expanded dataset allows for a more comprehensive evaluation of the generalization capability of the universal adversarial perturbation (UAP) across a broader range of visual concepts.
Generated samples of each class are visualized in Figure~\ref{fig:dataset_vis}. 
Furthermore, we designed two editing prompts for each class, each describing the manipulation of either a specific object or the background, as shown in Table~\ref{dataset}, following~\cite{semantic}.
For convenience, we refer to this entire image-text pair dataset (50$\times$10 concepts = 500 images) as $D_E$.}


\noindent
\paragraph{\textbf{\textup{Evaluation Dataset for Image Inpainting (${D}_I$).}}}
To verify whether our method extends beyond immunization for image editing to image inpainting,
we compare it with three image inpainting methods: DiffusionGuard (DG)~\cite{diffguard}, AdvPaint (AP)~\cite{advpaint}, and DiffVax~\cite{diffvax}.
For evaluation, we construct the inpainting dataset $D_I$ by selecting 3 images from each of the 10 object classes in $D_E$ and generating a binary mask for each image following the DG's inpainting protocol.
The white regions in the binary masks are then filled by the inpainting models.
\begin{figure}[h!]
\centering    \includegraphics[width=\linewidth]{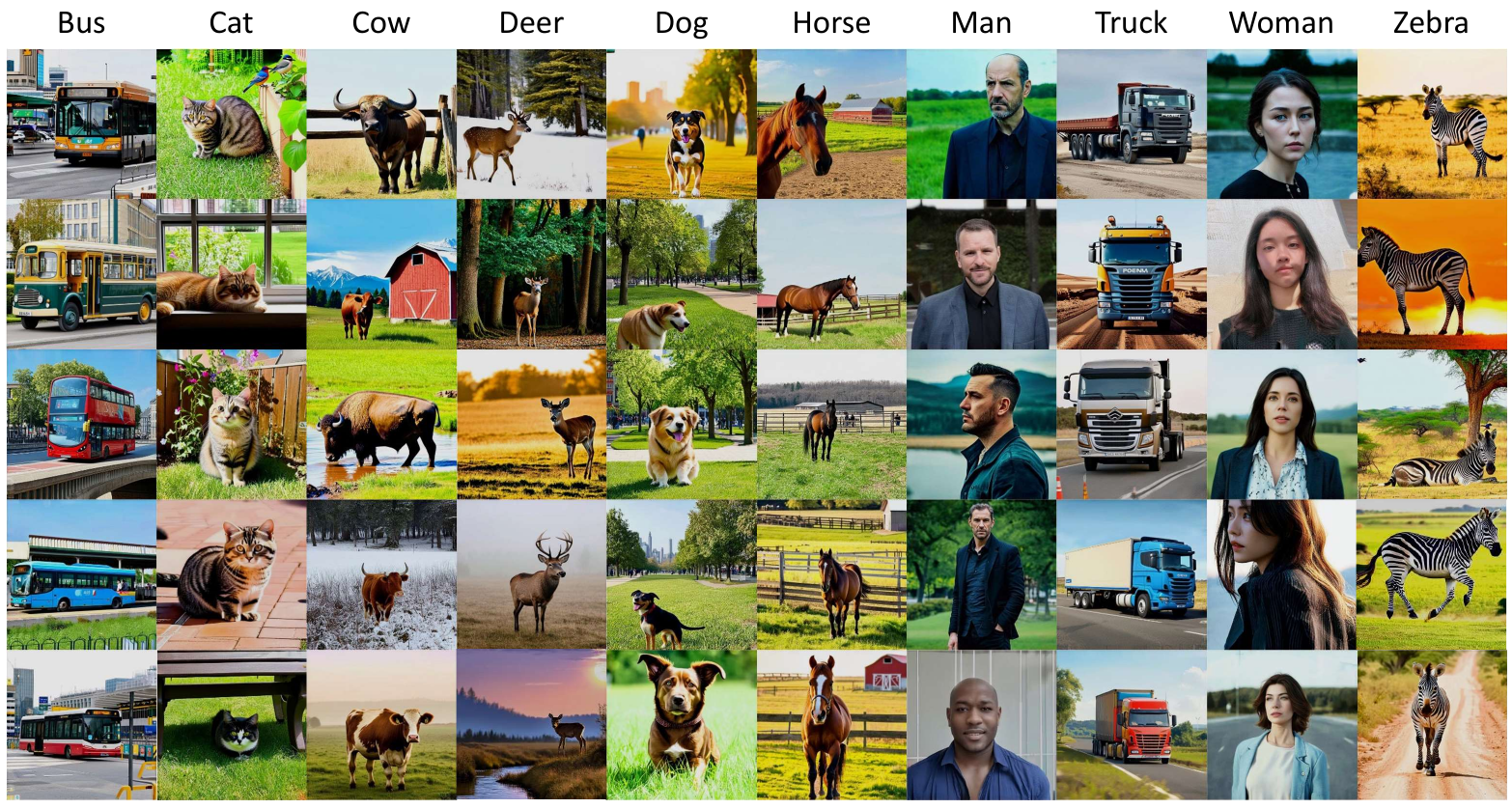}  
\vspace{-0.5cm}
    \caption{Examples of our generated evaluation dataset.
    }
    \vspace{-0.3cm}
    \label{fig:dataset_vis}
\end{figure}

\begin{table}[H]
\footnotesize
\setlength{\tabcolsep}{6pt}
\centering
\caption{Selected object classes and prompts for the evaluation dataset.
Each object class is associated with two editing prompts.
}
\begin{tabular}{c|c|c}
\Xhline{0.35mm}
{Object Class} & {Prompt 1} & {Prompt 2} \\
\Xhline{0.1mm}
bus     & \textit{``A photo of a truck"}       & \textit{``A bus in the desert"} \\
truck   & \textit{``A photo of a bus"}         & \textit{``A truck in the park"} \\
cat     & \textit{``A photo of a dog"}         & \textit{``A cat on the sand"} \\
dog     & \textit{``A photo of a cat"}        & \textit{``A dog with a ball"} \\
zebra   & \textit{``A photo of a deer"}       & \textit{``A zebra with a horse"} \\
deer    & \textit{``A photo of a zebra"}      & \textit{``A deer in the city"} \\
man     & \textit{``A photo of a woman"}       & \textit{``A man with a hat"} \\
woman   & \textit{``A photo of a man" }        & \textit{``A woman in the beach"} \\
cow     &\textit{``A photo of a horse"}       & \textit{``A cow is on mud"} \\
horse   & \textit{``A photo of a cow"}       & \textit{``A horse and a sheep"} \\
\Xhline{0.35mm}
\end{tabular}
\label{dataset}
\vspace{-0.7cm}
\end{table}

\noindent
\paragraph{\textbf{\textup{Evaluation Protocols.}}}
We use a diverse metrics to quantitatively evaluate our method.
Following the main manuscript, we report widely used image quality and perceptual metrics for image immunization~\cite{dct, semantic}: PSNR, SSIM~\cite{ssim}, VIFp~\cite{vifp}, FSIM~\cite{fsim}, and LPIPS~\cite{lpips}.
These metrics are computed between the editing results of immunized and non-immunized images to measure how strongly immunization alters the final edited outputs.
We also report {Feat. Sim. (C)}, a CLIP-based feature similarity metric, to measure how much the edited result still depends on the source image, since our goal is to induce editing results that do {not} rely on the original source content.
\vspace{-0cm}
\begin{itemize}[label={$\bullet$}]
    \item {PSNR} measures pixel-level fidelity based on mean squared error. {Higher} PSNR indicates greater pixel-level similarity to the original editing result, whereas successful immunization typically {lowers} PSNR by inducing larger deviations from that result.
    \item {SSIM}~\cite{ssim} evaluates structural similarity in terms of luminance, contrast, and texture. {Higher} SSIM implies stronger structural similarity to the clean result, while effective immunization generally leads to {lower} SSIM by disrupting its structure.
    \item {VIFp}~\cite{vifp} measures the amount of visual information preserved using perceptual modeling. {Higher} VIFp indicates better perceptual fidelity to the original result, whereas successful immunization tends to {reduce} VIFp by weakening such fidelity.
    \item {FSIM}~\cite{fsim} assesses feature-level similarity using phase congruency and gradient magnitude. {Higher} FSIM suggests closer feature-level similarity to the clean edited image, while strong immunization typically yields {lower} FSIM by altering salient features.
    \item {LPIPS}~\cite{lpips} compares deep feature representations to measure perceptual distance. {Lower} LPIPS indicates higher perceptual similarity to the original editing result, whereas effective immunization is reflected by {higher} LPIPS due to larger perceptual deviation from that result.
    \item {Feat. Sim. (C)} computes the cosine similarity between CLIP~\cite{clip} features of the \emph{source image} and the \emph{edited image}. {Higher} values indicate greater similarity, while successful immunization corresponds to {lower} values.
\end{itemize}

We follow the default settings of each pipeline for evaluation.
For the image-to-image pipeline of Stable Diffusion V1.5, we set the number of inference steps to 50, the strength to 0.8, and the guidance scale to 7.5.
For Stable Diffusion V1.5 inpainting, we use an inference step of 50, a guidance scale of 7.5, and a strength of 1.0.
To account for the stochastic nature of diffusion models, we randomly select 10 seeds and report the average of the quantitative results.

\noindent
\paragraph{\textbf{\textup{Target Generation}}}
To generate target images, we use Stable Diffusion~\cite{rombach2022high} with the prompt ``a photo of \textit{\textless target\textgreater}.''
As target concepts, we generate five representative examples—\textit{Ronaldo}, \textit{Tiger}, \textit{Sunflower}, \textit{Peacock}, and \textit{Mandala}—and evaluate our method on them.
{However, our approach is not limited to this particular set and can, in principle, be applied to any target concept, where multiple target images may also be considered for the same concept.}
Figure~\ref{fig:tar_uap_vis} shows examples of target images and their corresponding UAPs for five randomly selected target objects.
{Figure~\ref{fig:tigers} further visualizes UAPs trained with different target images generated with the same target prompt, `tiger.'}

\begin{figure}[h!]
\centering    \includegraphics[width=0.8\linewidth]{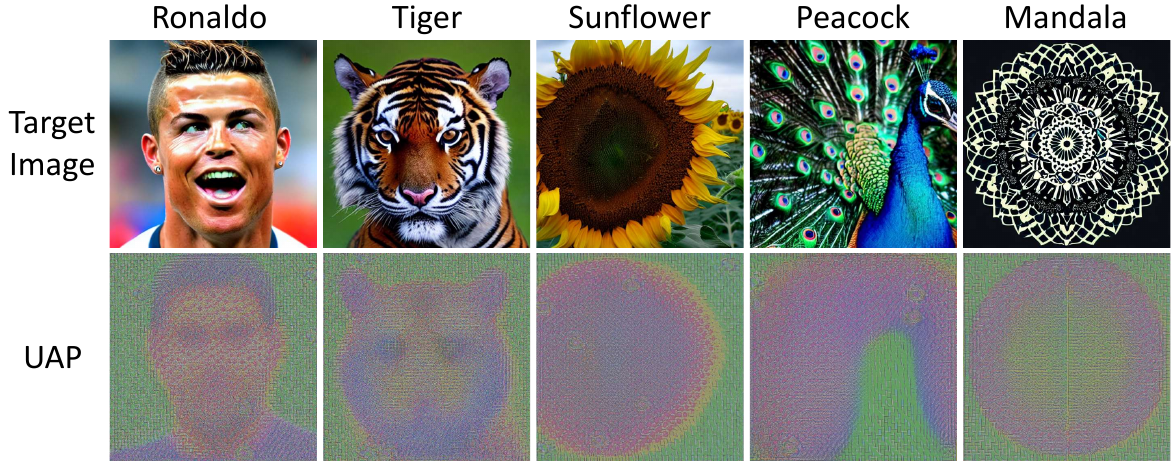}    
    \vspace{-0.2cm}\hspace{0.8cm}\caption{Visualization for targets and their corresponding UAPs. UAPs are scaled for visualization. 
    }
    \vspace{-0.3cm}
    \label{fig:tar_uap_vis}
\end{figure}

\begin{figure}[h!]
\centering    \includegraphics[width=0.8\linewidth]{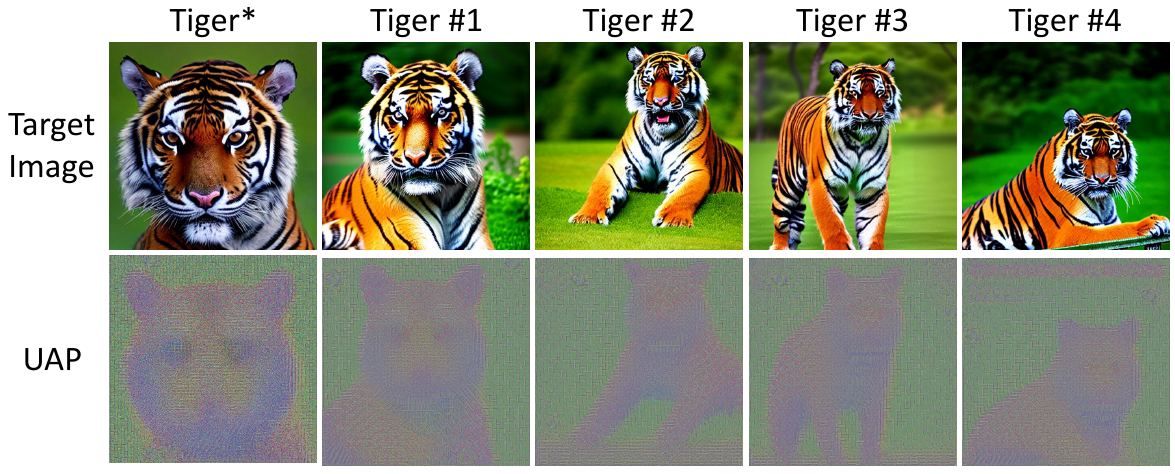}    
    \vspace{-0.2cm}\hspace{0.8cm}\caption{Visualization for diverse target images on the same prompt (`tiger') and their corresponding UAPs. * indicates the target image used in our experiments. UAPs are scaled for visualization. 
    }
    \vspace{-0.3cm}
    \label{fig:tigers}
\end{figure}



\subsection{Training Setup}
\label{app:train}
For training our UAP, we set the attack stepsize $s=1/255$, constraint parameter $\epsilon=10/255$, timestep set $K=\{5, 10, 15, 20, 25 \}$, inference steps to 50, and the guidance scale in diffusion models to 7.5. 
For image-specific methods~\cite{raising, semantic}, we adopt the same diffusion process settings, including 50 inference steps and a guidance scale of 7.5. 
The number of attack iterations is set to 100, and all other configurations follow those specified in the original implementations.

\begin{algorithm}[t]
\caption{UAP Training}
\label{alg:algo_ori}
\footnotesize
\textbf{Input:} Training image set $X$, its corresponding prompt token embedding set $T$, target image $x_{tar}$, target prompt token embedding $t_{tar}$, surrogate diffusion model $\mathcal{M}$, diffusion timestep set $K=\{k_1, k_2, ..., k_n\}$, training epoch $N$, attack stepsize $s$, perturbation budget $\epsilon$ \\
\textbf{Output:} Universal adversarial perturbation $\delta$.\\
\vspace{-0.4cm}
\begin{algorithmic}[1]
    \STATE Initialize $\delta \sim \mathcal{U}(-\epsilon, \epsilon)$
    \FOR{$i = 1, \dots, N$}
        \FOR{$x \in X$ and $t \in T$}
        \STATE $grad\_total\leftarrow 0$
        \STATE $x_{adv}\leftarrow \text{clip}(x + \delta, 0, 1)$
        \FOR{$k$ in $K$}
        \STATE Add the forward-process noise to both clean and immunized images 
        \\ $x^k \leftarrow x+\text{noise\_scheduler}(k)$
        \\ $x_{adv}^k \leftarrow x_{adv}+\text{noise\_scheduler}(k)$ 
        \\ $x_{\text{tar}}^k \leftarrow x_{\text{tar}} + \text{noise\_scheduler}(k)$\hspace{-0.2cm}
        \STATE Compute loss using Eq.~(5) and (6)
        \\ $\mathcal{L} \leftarrow  \mathcal{L}_{\text{inj}} + \mathcal{L}_{\text{sup}}$
        \STATE Update the total gradient 
        \\ $grad\_total \leftarrow grad\_total+\nabla \mathcal{L}$
        \ENDFOR
    \STATE Update and clip perturbation $\delta$
    \\ $\delta \leftarrow \delta - s \cdot sign(grad \_ total)$
    \\ $\delta \leftarrow  \text{min}(\epsilon, \text{max}(\delta, -\epsilon))$
    \ENDFOR
    \ENDFOR
    \RETURN $\delta$
\end{algorithmic}
\end{algorithm}

\begin{algorithm}[!t]
\caption{Test-time Immunization}
\label{alg:algo_test}
\footnotesize
\textbf{Input:} New image $x$, pre-computed UAP $\delta$\\
\textbf{Output:} immunized image ${x}_{adv}$.\\
\vspace{-0.4cm}
\begin{algorithmic}[1]
    \STATE ${x}_{adv} \leftarrow x+\delta$
    \RETURN ${x}_{adv}$
\end{algorithmic}
\end{algorithm}
\subsection{Algorithms}
Algorithm~\ref{alg:algo_ori} and Algorithm~\ref{alg:algo_test} present the training and testing procedures of our universal image immunization framework, respectively.
We extend the optimization strategy of Semantic Attack~\cite{semantic} from image-specific immunization to universal immunization by learning a single dataset-level perturbation, instead of optimizing perturbations independently for each test image.
Once learned, the universal perturbation $\delta$ is directly applied to an unseen image as $x_{\text{new}} + \delta$, enabling fast and scalable protection.

\vspace{-3mm}
\subsection{Data-free Universal Immunization}
\label{app:data-free}
In this subsection, we describe how to train a targeted universal adversarial perturbation (UAP) in a fully data-free manner against diffusion-based editing models.
Following the strategy of TRM-UAP~\cite{trm}, we generate random jigsaw puzzles and apply a mean filter to smooth the edges between regions, resembling real-world training samples.
We also adopt a curriculum learning scheme that starts with simple artificial images and gradually introduces more complex ones, controlled by a distribution parameter and the training iteration. 
Figure~\ref{fig:jigsaw} demonstrates examples of random jigsaw puzzles, where the complexity of the artificial images increases from left to right.
The overall training procedure for our data-free targeted UAP is presented in Algorithm~\ref{alg:algo_df}.
\begin{algorithm}[H]
\caption{Data-free Targeted UAP Training}
\label{alg:algo_df}
\footnotesize
\textbf{Input:} Random data prior $\mathcal{D}_r$, target image $x_{tar}$, target prompt token embedding $t_{tar}$, surrogate diffusion model $\mathcal{M}$, diffusion timestep set $K=\{k_1, k_2, ..., k_n\}$, training epoch $N$, attack stepsize $s$, constraint parameter $\epsilon$ \\
\textbf{Output:} Universal adversarial perturbation $\delta$.\\
\vspace{-0.4cm}
\begin{algorithmic}[1]
    \STATE Initialize $\delta \sim \mathcal{U}(-\epsilon, \epsilon)$
    \WHILE{$i < N$}
        \STATE $i = i + 1$
        \FOR{each training sample}
            
                \STATE Sample $x_r \sim \mathcal{D}_r$  
                \STATE $grad\_total \leftarrow 0$
                \STATE $x_{adv} \leftarrow clip(x_r + \delta, 0, 1)$
                \FOR{$k\ \text{in}\ K$}
                    \STATE // Inject forward noise to both target and immunized images \\
                    $x_{\text{tar}}^k \leftarrow x_{\text{tar}} + noise\_scheduler(k)$\\
                    $x_{adv}^k \leftarrow x_{adv} + noise\_scheduler(k)$
                    \STATE // Compute target semantic injection loss using Eq.~(4) \\
                    $\mathcal{L} \leftarrow \mathcal{L}_{inj}$ 
                    \STATE $grad\_total \leftarrow grad\_total + \nabla \mathcal{L}$
                \ENDFOR
            
             \STATE // Update and clip perturbation $\delta$
    \\ $\delta \leftarrow \delta - s \cdot sign(grad \_ total)$
    \\ $\delta \leftarrow  \text{min}(\epsilon, \text{max}(\delta, -\epsilon))$
        \ENDFOR
    \ENDWHILE
    \RETURN $\delta$
\end{algorithmic}
\end{algorithm}

\begin{figure}[h!]
\centering    \includegraphics[width=0.7\linewidth]{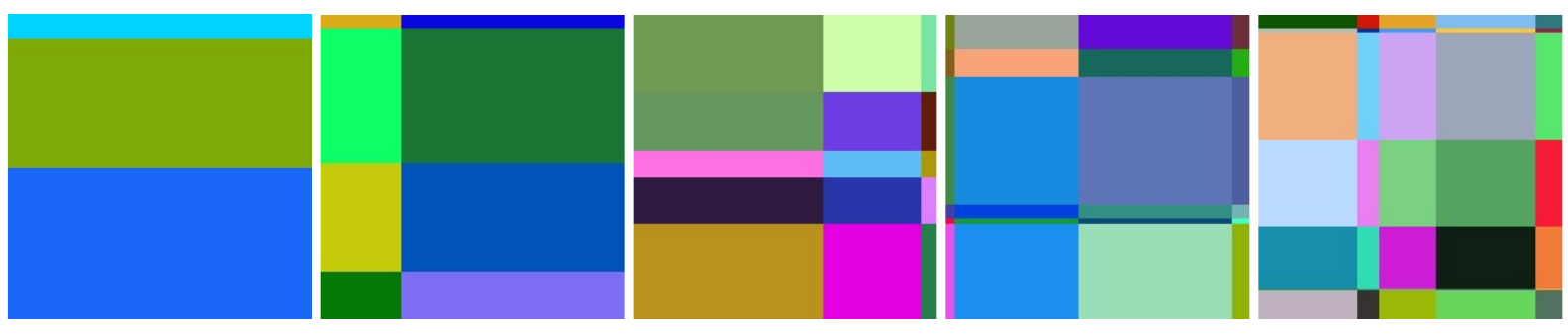} 
    \vspace{-0.2cm}
    \caption{Examples of jigsaw puzzle images.
    }
    \vspace{-0.2cm}
    \label{fig:jigsaw}
\end{figure}

\subsection{Theoretical Justification for Utilizing Cross-attention Output}
\label{sec:justification}
In latent-diffusion training, the denoiser $\epsilon_\theta(x_t,k,t)$ is guided by the conditioning prompt $t$, causing the cross-attention pathway to continuously steer latent features toward prompt-aligned semantics throughout the denoising trajectory. 
At layer $l$, a cross-attention block computes:
\begin{equation}
    Q_l = h_k^{(l)} W_l^Q,~~K_l = Y W_l^K,~~V_l = Y W_l^V,
\end{equation}
where $Y$ denotes the prompt embedding.
The attention map $\mathcal{A}_l$ and the output $CA_l$ are given by:
\begin{equation}
    \mathcal{A}_l = \mathrm{softmax}\!\big(Q_l K_l^\top/\sqrt{d_k}\big),~~
    CA_l = \mathcal{A}_l V_l,
\end{equation}
and the latent update follows:
\begin{equation}
    h_k^{(l+1)} = h_k^{(l)} + G_l(h_k^{(l)}, k) + W_l^{CA} CA_l,
    \label{eq:residual_update}
\end{equation}
where $G_l$ collects non–cross-attention operations (\eg, convolution or self-attention) and $W_l^{O}$ is the output projection.

Our goal is to inject target semantics so as to induce the diffusion model to misinterpret the input, while suppressing the original semantics.
From Eq.~(\ref{eq:residual_update}), the conditioning signal affects the latent \emph{only through the residual term} $W_l^{CA} CA_l$.
Therefore, to directly modulate the semantic information actually written into $h_k^{(l+1)}$, the most principled intervention point is the cross-attention output $CA_l$.

In contrast, the attention map $\mathcal{A}_l$ primarily determines \emph{where} to attend (token weighting) and does not by itself encode the semantic content being injected, which is carried by $V_l$ and propagated through $CA_l=\mathcal{A}_lV_l$.
Moreover, since $\mathcal{A}_l$ depends only on $Q_l$ and $K_l$, it is invariant to any invertible linear transformation of $V_l$ and thus cannot uniquely characterize the conditioning effect on the latent.
Consequently, operating on $\mathcal{A}_l$ alone can change token selection without reliably controlling the injected semantics.

Accordingly, we formulate both \textit{target semantic injection} and \textit{source semantic suppression} losses on $CA_l$ to influence the semantic interpretation of the denoiser $\epsilon_\theta(x_t,k,t)$.
This choice is further supported by the ablations in Section~\ref{app:ca-map} (Figure~\ref{fig:null_2} and Table~\ref{R3_W2}), where manipulating $CA_l$ yields substantially stronger semantic effects than operating on $\mathcal{A}_l$, leading to more effective image immunization.

\subsection{Implementation Details for Universalized Baselines}
\label{app:universal_baselines}
{We construct universalized baselines by adapting existing image-specific methods to the UAP setting with their original loss functions, ensuring a fair comparison, as we are the first to propose a universal immunization framework against diffusion-based image editing.
Specifically, we adopt three loss functions from prior image-specific studies: Encoder Attack~\cite{raising}, AdvPaint~\cite{advpaint}, and Semantic Attack~\cite{semantic}, as follows:}
\begin{table}[t]
\footnotesize
\centering
\caption{Imperceptibility comparison between our method and universal baselines. For fair comparison, we set larger perturbation budget (22/255) than universalized baselines and our method (10/255) for FastProtect (FP)~\cite{nearly}.}
\vspace{-0.2cm}
\resizebox{0.5\linewidth}{!}{
\begin{tabular}{c|cccc|cc}
\Xhline{0.35mm}
{Method} & Enc$_{10}$ & Emb$_{10}$ & Map$_{10}$ & Ours$_{10}$ & FP$_{10}$ & FP$_{22}$\\
\Xhline{0.1mm}
DISTS $\downarrow$ 
& 0.199 & 0.180 & {0.175} & 0.181 &  0.105 & {0.216} \\

PSNR $\uparrow$
& 28.97 & 29.28 & {29.23} & 29.41 & 37.07 & {30.35} \\

LPIPS $\downarrow$
& 0.345 & 0.342 & {0.335} & 0.355 & 0.170 & {0.360} \\

\Xhline{0.35mm}
\end{tabular}
}
\label{tab:fastprotect_imper}
\vspace{-0.2cm}
\end{table}
{
\begin{itemize}
    \item \textbf{Encoder Attack (Encoder)}: following the original Encoder Attack’s strategy, we minimize the $L_2$ distance between the latent representation of an uninformative black reference image and that of the perturbed training image by UAP.
    \item \textbf{Embedding Attack (Embedding)}: we maximize an $L_2$ loss between the intermediate representations (\ie query embedding in cross-attention layers and query/key/value embeddings in self-attention layers) of the perturbed image and those of the original, encouraging them to deviate.
    \item \textbf{Attention Map Attack (Map)}: following Semantic Attack~\cite{semantic}, we minimize the alignment between attention maps and the original image semantics, thereby disrupting semantic consistency. 
\end{itemize}
All three baselines are trained on the same dataset as ours, with identical hyperparameters, including 20 training epochs, step size $s=1/255$, and perturbation budget $\epsilon=10/255$.

\subsection{Implementation Details for FastProtect}
\label{sec:FP_imple}
The optimization-free image protection method FastProtect (FP)~\cite{nearly} employs an LPIPS-based scaling strategy to improve imperceptibility.
Accordingly, in the original FastProtect paper, FP and the image-specific baselines are assigned different perturbation budgets and calibrated to achieve comparable immunization performance, after which imperceptibility is reported.
In our setting, we fix the perturbation budget of our method and all universalized baselines to $10/255$ and first measure their imperceptibility using DISTS~\cite{dists}, PSNR, and LPIPS~\cite{lpips} under this common constraint.
We then determine a separate perturbation budget ($\epsilon=22/255$) for FP, reported in Table~\ref{tab:fastprotect_imper}. 
This procedure respects the design choices and contribution of the FastProtect while enabling a fair comparison against our universal defense.

}

\section{Additional Experimental Results}

\begin{table}[t]
\scriptsize
\setlength{\tabcolsep}{4pt}
\centering
\caption{Examples of complex editing prompts.}
\vspace{-2mm}
\begin{tabular}{>{\centering\arraybackslash}m{1.5cm}|>{\centering\arraybackslash}m{3.3cm}|m{6cm}}
\Xhline{0.35mm}
Object Class & Original & \centering\arraybackslash Complex Variation\\
\Xhline{0.1mm}
bus &
\textit{``A photo of a truck''} &
\textit{``A photo of a truck in daylight with sturdy wheels, metallic panels, clear windows, and realistic details.''} \\
\Xhline{0.05mm}
man &
\textit{``A man with a hat''} &
\textit{``A photo of a man wearing a hat in a softly lit outdoor setting with warm tones and blurred foliage.''} \\
\Xhline{0.05mm}
woman &
\textit{``A woman on the beach''} &
\textit{``A photo of a woman on a bright beach with soft waves, warm sunlight, and a calm coastal horizon.''} \\
\Xhline{0.35mm}
\end{tabular}
\label{complex_example}
\end{table}
\begin{table}[t!]
\footnotesize
\setlength{\tabcolsep}{3pt}
\centering
\caption{Evaluating the robustness on complex prompt editing compared with universalized baselines. The best result is shown in bold; the second-best is underlined.}
\vspace{-0.2cm}
\resizebox{\linewidth}{!}{
\begin{tabular}{c|c|ccc|c|c|ccc|c|c|ccc|c|c|ccc|c}
\Xhline{0.35mm}
{Model}
& \multicolumn{5}{c|}{Stable Diffusion V1.4} 
& \multicolumn{5}{c|}{Stable Diffusion V1.5*} 
& \multicolumn{5}{c|}{Stable Diffusion V2.0}
& \multicolumn{5}{c}{InstructPix2Pix} \\
\Xhline{0.1mm}
Method  
& clean & \cellcolor{green!15}Enc. & \cellcolor{green!15}Emb. & \cellcolor{green!15}Map & \cellcolor{green!35}Ours
& clean & \cellcolor{green!15}Enc. & \cellcolor{green!15}Emb. & \cellcolor{green!15}Map & \cellcolor{green!35}Ours 
& clean & \cellcolor{green!15}Enc. & \cellcolor{green!15}Emb. & \cellcolor{green!15}Map & \cellcolor{green!35}Ours 
& clean & \cellcolor{green!15}Enc. & \cellcolor{green!15}Emb. & \cellcolor{green!15}Map & \cellcolor{green!35}Ours \\
\Xhline{0.1mm}

PSNR $\downarrow$ 
& --&17.75 & 16.70 & 17.12 & \textbf{14.88}
& --&17.73 & 16.70 & 17.19 & \textbf{15.66}
& --&14.82 & 14.70 & 14.35 & \textbf{13.31}
& --&17.27 & 14.95 & 16.56 & \textbf{14.01} \\

SSIM $\downarrow$ 
& --&0.607 & 0.449 & 0.555 & \textbf{0.400}
& --&0.608 & 0.447 & 0.562 & \textbf{0.422}
& --&0.426 & 0.349 & 0.365 & \textbf{0.314}
& --&0.595 & 0.496 & 0.547 & \textbf{0.472} \\

VIFp $\downarrow$ 
& --&0.223 & 0.150 & 0.190 & \textbf{0.108}
& --&0.221 & 0.148 & 0.192 & \textbf{0.121}
& --&0.123 & 0.102 & 0.105 & \textbf{0.079}
& --&0.211 & 0.180 & 0.190 & \textbf{0.155} \\

FSIM $\downarrow$ 
& --&0.774 & 0.722 & 0.745 & \textbf{0.671}
& --&0.774 & 0.720 & 0.747 & \textbf{0.687}
& --&0.666 & 0.658 & 0.635 & \textbf{0.611}
& --&0.805 & 0.742 & 0.779 & \textbf{0.717} \\

LPIPS $\uparrow$  
& --&0.368 & 0.509 & 0.422 & \textbf{0.579}
& --&0.369 & 0.511 & 0.417 & \textbf{0.548}
& --&0.503 & 0.572 & 0.551 & \textbf{0.620}
& --&0.351 & 0.451 & 0.400 & \textbf{0.480} \\

Feat. Sim. (C) $\downarrow$
& 0.740 &0.714 & 0.702 & 0.708 & \textbf{0.685}
& 0.739 &0.711 & 0.702 & 0.707 & \textbf{0.689}
& 0.706 &0.666 & 0.665 & 0.659 & \textbf{0.633}
& 0.755 &0.727 & 0.691 & 0.719 & \textbf{0.684} \\


\Xhline{0.35mm}
\end{tabular}
}
\label{complex-universal}
\end{table}

\begin{table}[t!]
\footnotesize
\setlength{\tabcolsep}{3pt}
\centering
\caption{Evaluating the robustness on complex prompt editing compared with image-specific and optimization-free methods. Stable Diffusion V1.5 is the white-box model. The best result is shown in bold; the second-best is underlined.}
\vspace{-0.2cm}
\resizebox{\linewidth}{!}{
\begin{tabular}{c|c|ccc|c|c|c|ccc|c|c|c|ccc|c|c|c|ccc|c|c}
\Xhline{0.35mm}
{Model}
& \multicolumn{6}{c|}{Stable Diffusion V1.4} 
& \multicolumn{6}{c|}{Stable Diffusion V1.5*} 
& \multicolumn{6}{c|}{Stable Diffusion V2.0}
& \multicolumn{6}{c}{InstructPix2Pix} \\
\Xhline{0.1mm}
\multirow{2}{*}{Method}
& \multirow{2}{*}{clean}
& \multicolumn{3}{c|}{\cellcolor{gray!15}\textit{Image-specific}}
& \multicolumn{1}{c|}{\cellcolor{cyan!15}\textit{Opt-free}}
& \multicolumn{1}{c|}{\cellcolor{green!35}\textit{Univ.}}
& \multirow{2}{*}{clean}
& \multicolumn{3}{c|}{\cellcolor{gray!15}\textit{Image-specific}}
& \multicolumn{1}{c|}{\cellcolor{cyan!15}\textit{Opt-free}}
& \multicolumn{1}{c|}{\cellcolor{green!35}\textit{Univ.}}
& \multirow{2}{*}{clean}
& \multicolumn{3}{c|}{\cellcolor{gray!15}\textit{Image-specific}}
& \multicolumn{1}{c|}{\cellcolor{cyan!15}\textit{Opt-free}}
& \multicolumn{1}{c|}{\cellcolor{green!35}\textit{Univ.}}
& \multirow{2}{*}{clean}
& \multicolumn{3}{c|}{\cellcolor{gray!15}\textit{Image-specific}}
& \multicolumn{1}{c|}{\cellcolor{cyan!15}\textit{Opt-free}}
& \multicolumn{1}{c}{\cellcolor{green!35}\textit{Univ.}} \\
&  & \cellcolor{gray!15}EA & \cellcolor{gray!15}DA & \cellcolor{gray!15}SA & \cellcolor{cyan!15}FP & \cellcolor{green!35}Ours
&  & \cellcolor{gray!15}EA & \cellcolor{gray!15}DA & \cellcolor{gray!15}SA & \cellcolor{cyan!15}FP & \cellcolor{green!35}Ours
&  & \cellcolor{gray!15}EA & \cellcolor{gray!15}DA & \cellcolor{gray!15}SA & \cellcolor{cyan!15}FP & \cellcolor{green!35}Ours
&  & \cellcolor{gray!15}EA & \cellcolor{gray!15}DA & \cellcolor{gray!15}SA & \cellcolor{cyan!15}FP & \cellcolor{green!35}Ours \\
\Xhline{0.1mm}

PSNR $\downarrow$ 
& --& \underline{15.08} & 15.14 & 17.31 & 17.54 & \textbf{14.88}
& --& \underline{15.18} & \textbf{15.13} & 17.36 & 17.58 & 15.66
& --& \underline{13.16} & \textbf{13.15} & 15.32 & 14.48 & 13.31
& -- & \underline{14.98}&15.44 & 15.07 & 17.22 & \textbf{14.01} \\

SSIM $\downarrow$ 
& -- & \underline{0.413} & 0.419 & 0.504 & 0.552 & \textbf{0.400} 
& --& \textbf{0.420} & \underline{0.422} & 0.506 & 0.555 & \underline{0.422}
& --& 0.318 & \textbf{0.308} & 0.411 & 0.374 & \underline{0.314}
& -- & 0.483 &\underline{0.481} & 0.495 & 0.546 & \textbf{0.472} \\

VIFp $\downarrow$ 
& --& \textbf{0.105} & 0.129 & 0.172 & 0.209 & \underline{0.108}
& --& \textbf{0.111} & 0.128 & 0.173 & 0.208 & \underline{0.121}
& --& \textbf{0.075} & 0.085 & 0.120 & 0.116 & \underline{0.079}
& -- & \textbf{0.136}&\underline{0.150}  & 0.174 & 0.190 & 0.155 \\

FSIM $\downarrow$ 
& --& \textbf{0.643} & 0.673 & 0.747 & 0.746 & \underline{0.671}
& --& \textbf{0.660} & \underline{0.675} & 0.748 & 0.748 & 0.687
& --& \textbf{0.579} & \underline{0.596} & 0.680 & 0.638 & 0.611
& -- & \underline{0.736} &0.755 & 0.756 & 0.788 & \textbf{0.717} \\

LPIPS $\uparrow$  
& --& \underline{0.570} & 0.536 & 0.451  & 0.399 & \textbf{0.579}
& --& \textbf{0.558} & 0.536 & 0.451 & 0.397 & \underline{0.548}
& --& \textbf{0.624} & 0.607 & 0.530 & 0.528 & \underline{0.620}
& -- & \underline{0.478} &0.467 & 0.445 & 0.395 & \textbf{0.480} \\

Feat. Sim. (C) $\downarrow$
& 0.740 & \textbf{0.667} & 0.686 & 0.710 & 0.710 & \underline{0.685}
& 0.739 & \textbf{0.674} & 0.693 & 0.710 & 0.711 & \underline{0.689}
& 0.706 & \textbf{0.619} & 0.641 & 0.670 & 0.664 & \underline{0.633}
& 0.755 &\underline{0.700} & 0.718  & 0.703 & 0.725& \textbf{0.684} \\


\Xhline{0.35mm}
\end{tabular}
}
\vspace{-0.1cm}
\label{complex_specific}
\end{table}




\subsection{Evaluation with Complex Editing Prompts}
We conduct additional editing experiments to verify that our method maintains strong immunization performance even under more elaborate editing prompts.
Specifically, we use GPT-4~\cite{gpt4} to rewrite the prompts in Table~\ref{dataset} into more detailed and complex versions.
Examples of these complex prompts are provided in Table~\ref{complex_example}.
As shown in Table~\ref{complex-universal}, our method clearly outperforms universalized baselines under these complex prompts, demonstrating robustness to prompt elaboration.

We further compare our method with image-specific and optimization-free immunization methods in Table~\ref{complex_specific}.
Despite operating under more restrictive constraints, including a lower perturbation budget, stronger imperceptibility, and universality (as shown in Table~\textcolor{red}{4} of the manuscript), our method achieves performance comparable or even superior to image-specific and optimization-free methods while incurring \emph{near-zero inference cost without GPU acceleration}, further demonstrating its robustness and practicality.

\begin{table}[t!]
\footnotesize
\setlength{\tabcolsep}{5pt}
\renewcommand{\arraystretch}{0.78}
\centering
\caption{Evaluation on noun-free background-editing prompts. Stable Diffusion V1.5 is used as the white-box model. The best result is shown in bold; the second-best is underlined.}
\vspace{-0.2cm}
\resizebox{0.6\linewidth}{!}{
\begin{tabular}{c|c|ccc|c}
\Xhline{0.35mm}
\multirow{2}{*}{Metric}
& \multirow{2}{*}{Clean}
& \multicolumn{3}{c|}{\cellcolor{gray!15}\textit{Image-specific}}
& \multicolumn{1}{c}{\cellcolor{green!35}\textit{Universal}} \\
&
& \cellcolor{gray!15}EA
& \cellcolor{gray!15}DA
& \cellcolor{gray!15}SA
& \cellcolor{green!35}Ours \\
\Xhline{0.1mm}

PSNR $\downarrow$
& --
& \textbf{15.46}
& 15.56
& 17.35
& \underline{15.52} \\

SSIM $\downarrow$
& --
& \underline{0.421}
& 0.430
& 0.486
& \textbf{0.419} \\

VIFp $\downarrow$
& --
& \textbf{0.126}
& 0.132
& 0.167
& \underline{0.128} \\

FSIM $\downarrow$
& --
& \underline{0.674}
& \textbf{0.665}
& 0.739
& 0.678 \\

LPIPS $\uparrow$
& --
& \underline{0.555}
& 0.545
& 0.476
& \textbf{0.559} \\

Feat. Sim. (C) $\downarrow$
& 0.668
& \underline{0.557}
& 0.564
& 0.602
& \textbf{0.555} \\

\Xhline{0.35mm}
\end{tabular}
}
\vspace{-0.15cm}
\label{tab:noun_free}
\end{table}

\begin{table}[h]
\footnotesize
\centering
\caption{Black-box immunization performance on Stable Diffusion V3~\cite{sd3} for image editing using $D_E$. 
All perturbations are generated using Stable Diffusion V1.5~\cite{rombach2022high}.}

\resizebox{0.8\linewidth}{!}{
\begin{tabular}{c|c|ccc|c|ccc|c}
\Xhline{0.35mm}
\multirow{2}{*}{Method}
& \multirow{2}{*}{Clean}
& \multicolumn{3}{c|}{\cellcolor{gray!15}\textit{Image-specific}} 
& \multicolumn{1}{c|}{\cellcolor{cyan!15}\textit{Opt-free}} 
& \multicolumn{3}{c|}{\cellcolor{green!15}\textit{Universalized}}
& \multicolumn{1}{c}{\cellcolor{green!35}\textit{Universal}} \\

& &  \cellcolor{gray!15}EA & \cellcolor{gray!15}DA & \cellcolor{gray!15}SA
& \cellcolor{cyan!15}FP& \cellcolor{green!15}Enc. & \cellcolor{green!15}Emb. & \cellcolor{green!15}Map & \cellcolor{green!35}Ours \\
\Xhline{0.1mm}
PSNR $\downarrow$ & --& 19.24 & \underline{18.71} & 20.01 & 20.65 & 19.96 & 19.39 & 18.92 & \textbf{17.63} \\
SSIM $\downarrow$ & --& 0.653 & \textbf{0.620} & 0.715 & 0.740 & 0.695 & 0.653 & 0.636 & \underline{0.624} \\
VIFp $\downarrow$ & --& 0.234 & \underline{0.230} & 0.301 & 0.315 & 0.284 & 0.264 & 0.246 & \textbf{0.227} \\
FSIM $\downarrow$ & --& 0.783 & \underline{0.772} & 0.806 & 0.824 & 0.804 & 0.786 & 0.777 & \textbf{0.756} \\
LPIPS $\uparrow$  & --& 0.452 & \underline{0.465} & 0.384 & 0.370 & 0.405 & 0.428 & 0.448 & \textbf{0.467} \\
Feat. Sim. (C) & 0.768 & 0.733 & \underline{0.723} & 0.738  &  0.746 & 0.743 & 0.731 & 0.734 &\textbf{0.720}\\
\Xhline{0.35mm}
\end{tabular}

}
\label{tab:sd3}
\vspace{-6mm}
\end{table}

\subsection{Evaluation with Noun-free Editing Prompts}
We additionally evaluate the robustness of our framework under different textual formats for background editing.
While the main experiments use noun-explicit prompts, \eg, “A woman at the beach” and “A cat on a rainy day,” we further consider noun-omitted prompts, \eg, “at the beach” and “on a rainy day.”
As shown in Table~\ref{tab:noun_free}, our method consistently achieves strong protection performance under these prompt variations, demonstrating that the proposed universal perturbation generalizes beyond a single prompt format.

\subsection{Black-box Evaluation on DiT-based Diffusion Model}
To evaluate the robustness of our method across different model architectures, we perform experiments using diffusion transformer~(DiT) model.
Specifically, we adopt Stable Diffusion V3~\cite{sd3}, configured to generate images at a resolution of $512 \times 512$, consistent with our experimental setup, for the image editing task.
{All universal and image-specific perturbations are generated on Stable Diffusion V1.5~\cite{rombach2022high}.}


In Table~\ref{tab:sd3}, while all methods, including ours, tend to perform worse on transformer-based diffusion models compared to U-Net-based ones, our method achieves superior performance for image immunization compared to other methods.
Specifically, compared with the universalized baselines, our method achieves significantly stronger immunization performance across all metrics, clearly demonstrating the effectiveness of our approach.
Even when compared with image-specific and optimization-free methods such as EA, DA~\cite{raising}, SA~\cite{semantic}, and FP~\cite{nearly} our method attains the highest performance in every metric except SSIM, highlighting its robustness beyond the universal setting.
These results confirm that our method not only provides superior immunization on U-Net-based diffusion models, but also exhibits stronger cross-architecture transferability than universalized baselines, as well as image-specific and optimization-free methods.
\begin{table}[t!]
\footnotesize
\centering
\caption{Image immunization performance on Stable Diffusion V1.5, V2.0~\cite{rombach2022high}, and FLUX 1.0~\cite{flux} for image inpainting, evaluated on $D_I$.
* denotes white-box surrogate model for DG, AP, and DiffVax, while our method is evaluated in the black-box setting. $\dagger$ indicates methods specifically designed for inpainting immunization, which requires masks to generate perturbations, whereas \textit{Ours} uses a single perturbation across all images without mask-based training.}
\vspace{-0.2cm}
\resizebox{1\linewidth}{!}{
\begin{tabular}{c|cc|c|c|cc|c|c|cc|c|c}
\Xhline{0.35mm}
Model
& \multicolumn{4}{c|}{Stable Diffusion 1.5*}
& \multicolumn{4}{c|}{Stable Diffusion 2.0}
& \multicolumn{4}{c}{FLUX} \\
\Xhline{0.1mm}
\multirow{2}{*}{Method}
& \multicolumn{2}{c|}{\cellcolor{gray!15}\textit{Image-specific}}
& \multicolumn{1}{c|}{\cellcolor{cyan!15}\textit{Opt-free}}
& \multicolumn{1}{c|}{\cellcolor{green!35}\textit{Universal}}
& \multicolumn{2}{c|}{\cellcolor{gray!15}\textit{Image-specific}}
& \multicolumn{1}{c|}{\cellcolor{cyan!15}\textit{Opt-free}}
& \multicolumn{1}{c|}{\cellcolor{green!35}\textit{Universal}}
& \multicolumn{2}{c|}{\cellcolor{gray!15}\textit{Image-specific}}
& \multicolumn{1}{c|}{\cellcolor{cyan!15}\textit{Opt-free}}
& \multicolumn{1}{c}{\cellcolor{green!35}\textit{Universal}} \\
& \cellcolor{gray!15}DG$^\dagger$ & \cellcolor{gray!15}AP$^\dagger$ & \cellcolor{cyan!15}DiffVax$^\dagger$ & \cellcolor{green!35}Ours
& \cellcolor{gray!15}DG$^\dagger$ & \cellcolor{gray!15}AP$^\dagger$ & \cellcolor{cyan!15}DiffVax$^\dagger$ & \cellcolor{green!35}Ours
& \cellcolor{gray!15}DG$^\dagger$ & \cellcolor{gray!15}AP$^\dagger$ & \cellcolor{cyan!15}DiffVax$^\dagger$ & \cellcolor{green!35}Ours \\
\Xhline{0.1mm}
PSNR $\downarrow$
& 18.10 & \textbf{15.16} & 17.64 & \underline{17.51}
& 18.48 & \textbf{15.27} & \underline{16.42} & {17.24}
& 25.84 & {22.67} & \textbf{18.35} & \underline{20.60} \\
SSIM $\downarrow$
& 0.615 & \textbf{0.499} & 0.648 & \underline{0.528}
& 0.643 & \textbf{0.519} & 0.585 & \underline{0.534}
& 0.810 & {0.691} & \underline{0.623} & \textbf{0.598} \\
VIFp $\downarrow$
& 0.309 & \textbf{0.238} & 0.423 & \underline{0.239}
& 0.325 & \underline{0.246} & 0.386 & \textbf{0.239}
& 0.446 & \underline{0.344} & 0.375 & \textbf{0.258} \\
FSIM $\downarrow$
& 0.816 & \textbf{0.761} & 0.821 & \underline{0.805}
& 0.825 & \textbf{0.762} & 0.781 & \underline{0.791}
& 0.923 & {0.883} & \textbf{0.828} & \underline{0.859} \\
LPIPS $\uparrow$
& 0.448 & \textbf{0.556} & 0.357 & \underline{0.488}
& 0.431 & \textbf{0.553} & 0.413 & \underline{0.498}
& 0.317 & \underline{0.447} & 0.413 & \textbf{0.489} \\
\Xhline{0.35mm}
\end{tabular}
}
\label{tab:inpaint_final}
\vspace{-0cm}
\end{table}
\begin{table}[t!]
\footnotesize
\centering

\caption{Performance comparison evaluated on DiffusionGuard~\cite{diffguard} benchmark.
* denotes white-box surrogate model for DG, AP, and DiffVax, while our method is evaluated in the black-box setting. $\dagger$ indicates methods specifically designed for inpainting immunization, which requires masks to generate perturbations, whereas \textit{Ours} uses a single perturbation across all images without mask-based training.}
\vspace{-0.2cm}
\resizebox{\linewidth}{!}{
\begin{tabular}{c|cc|c|c|cc|c|c|cc|c|c}
\Xhline{0.35mm}
Model
& \multicolumn{4}{c|}{Stable Diffusion 1.5*}
& \multicolumn{4}{c|}{Stable Diffusion 2.0}
& \multicolumn{4}{c}{FLUX} \\
\Xhline{0.1mm}
\multirow{2}{*}{Method}
& \multicolumn{2}{c|}{\cellcolor{gray!15}\textit{Image-specific}}
& \multicolumn{1}{c|}{\cellcolor{cyan!15}\textit{Opt-free}}
& \multicolumn{1}{c|}{\cellcolor{green!35}\textit{Universal}}
& \multicolumn{2}{c|}{\cellcolor{gray!15}\textit{Image-specific}}
& \multicolumn{1}{c|}{\cellcolor{cyan!15}\textit{Opt-free}}
& \multicolumn{1}{c|}{\cellcolor{green!35}\textit{Universal}}
& \multicolumn{2}{c|}{\cellcolor{gray!15}\textit{Image-specific}}
& \multicolumn{1}{c|}{\cellcolor{cyan!15}\textit{Opt-free}}
& \multicolumn{1}{c}{\cellcolor{green!35}\textit{Universal}} \\
& \cellcolor{gray!15}DG$^\dagger$ & \cellcolor{gray!15}AP$^\dagger$ & \cellcolor{cyan!15}DiffVax$^\dagger$ & \cellcolor{green!35}Ours
& \cellcolor{gray!15}DG$^\dagger$ & \cellcolor{gray!15}AP$^\dagger$ & \cellcolor{cyan!15}DiffVax$^\dagger$ & \cellcolor{green!35}Ours
& \cellcolor{gray!15}DG$^\dagger$ & \cellcolor{gray!15}AP$^\dagger$ & \cellcolor{cyan!15}DiffVax$^\dagger$ & \cellcolor{green!35}Ours \\
\Xhline{0.1mm}
PSNR $\downarrow$
& 15.81 & \textbf{12.69} & \underline{15.57} & 16.00 
& 16.82 & \textbf{13.62} & \underline{15.72} & 16.21 
& 23.28 & 20.71 & \textbf{18.55} & \underline{19.37} \\
SSIM $\downarrow$
& 0.495 & \textbf{0.336} & 0.531 & \underline{0.432} 
& 0.564 & \textbf{0.426} & 0.564 & \underline{0.490}
& 0.751 & \underline{0.624} & 0.660 & \textbf{0.476} \\
VIFp $\downarrow$
& \underline{0.169} & \textbf{0.127} & 0.217 & 0.172 
& 0.202 & \textbf{0.161} & 0.237 & \underline{0.186} 
& 0.328 & \underline{0.241} & 0.337 & \textbf{0.196} \\
FSIM $\downarrow$
& 0.743 & \textbf{0.656} & \underline{0.737} & 0.741 
& 0.776 & \textbf{0.700} & \underline{0.743} & 0.751 
& 0.876 & 0.820 & \textbf{0.779} & \underline{0.780} \\
LPIPS $\uparrow$
& 0.553 & \textbf{0.664} & 0.493 & \underline{0.555}  
& 0.516 & \textbf{0.621} & 0.471 & \underline{0.542}
& 0.351 & \underline{0.468} & 0.409 & \textbf{0.536} \\
\Xhline{0.35mm}
\end{tabular}
}
\label{tab:inpaint_diffguard}
\end{table}
\subsection{Comparison with Image Inpainting Immunization}
We train our UAP on the Stable Diffusion V1.5 image-to-image pipeline to prevent unauthorized image editing. However, to validate its robustness on the image inpainting task, we conduct experiments on inpainting pipelines including Stable Diffusion inpainting V1.5, V2.0~\cite{rombach2022high}, and Dit-based FLUX 1.0~\cite{flux}.
To establish baselines, we consider prior image-specific methods, DiffusionGuard (DG)~\cite{diffguard} and AdvPaint (AP)~\cite{advpaint}, together with the optimization-free method DiffVax~\cite{diffvax}, which employs a pretrained immunizer tailored to inpainting models.
In particular, these baselines~\cite{diffguard, advpaint, diffvax} have a \emph{white-box advantage} on the task itself, as they use training masks and directly optimize perturbations~\cite{diffguard, advpaint} and pretrain immunizer~\cite{diffvax} with inpainting models, whereas our method is trained on an image-to-image pipeline without any masks.
We evaluate on our inpainting dataset $D_I$ and additionally on the real-image inpainting benchmark provided by DiffusionGuard~\cite{diffguard}.

As shown in Table~\ref{tab:inpaint_final}, our method achieves comparable performance even without prior knowledge of the target models or masks, including cases where the baselines operate in a white-box model (Stable Diffusion V1.5 Inpainting).
A similar performance trend is also observed on the real-image benchmark in Table~\ref{tab:inpaint_diffguard}.
Notably, on the DiT-based FLUX 1.0~\cite{flux}, our method can outperform existing inpainting-specific methods, including image-specific and optimization-free.
Table~\ref{tab:test-cost} further shows that the effectiveness of our method is achieved with negligible test-time overhead, unlike prior approaches, highlighting the practical scalability of a single pretrained UAP.
These results demonstrate that our method offers strong task transferability to inpainting models and robust cross-architecture transferability, highlighting its practicality for real-world deployment.
\begin{table}[t!]
\footnotesize
\centering
\caption{Comparison of test-time computational cost between image-specific immunization methods and our universal approach. 
$\dagger$ indicates methods specifically designed for image inpainting immunization, requiring masks to generate perturbations.}
\vspace{-0.2cm}
\resizebox{0.75\linewidth}{!}{
\begin{tabular}{c|ccccc|c|c}
\Xhline{0.35mm}
\multirow{2}{*}{Method} 
& \multicolumn{5}{c|}{\cellcolor{gray!15}\textit{Image-specific}}
& \multicolumn{1}{c|}{\cellcolor{cyan!15}\textit{Opt-free}}
& \multicolumn{1}{c}{\cellcolor{green!35}\textit{Universal}} \\ 
& \cellcolor{gray!15}EA & \cellcolor{gray!15}DA & \cellcolor{gray!15}SA & \cellcolor{gray!15}DG{$^\dagger$} & \cellcolor{gray!15}AP{$^\dagger$}
& \cellcolor{cyan!15}DiffVax{$^\dagger$} & \cellcolor{green!35}Ours \\
\Xhline{0.1mm}
Latency (sec)    & 8.01 & 212.46 & 55.66 & 156.97 & 183.27 & 0.08 & $\sim$0 \\
GPU Usage (GB)   & 6.51 & 30.84  & 9.08  & 5.13   & 19.59  & 5.45 & 0 \\
Requires mask    & \ding{55}
                 &\ding{55}
                 & \ding{55}
                 & \ding{51}
                 & \ding{51}
                  & \ding{51}
                 & \ding{55}\\
Per-image opt. & \ding{51} & \ding{51} & \ding{51} & \ding{51} & \ding{51} &\ding{55} &\ding{55}\\
\Xhline{0.35mm}
\end{tabular}
}
\label{tab:test-cost}
\end{table}

\begin{table}[h]
\footnotesize
\centering
\caption{Performance comparison on IP2P~\cite{brooks2023instructpix2pix} with EditShield (ES)~\cite{editshield} which is specified for instruction-guided image editing model.
All perturbations are generated using Stable Diffusion V1.5.}

\resizebox{0.7\linewidth}{!}{
\begin{tabular}{c|cccc|ccc|c}
\Xhline{0.35mm}
\multirow{2}{*}{Method}
& \multicolumn{4}{c|}{\cellcolor{gray!15}\textit{Image-specific}}
& \multicolumn{3}{c|}{\cellcolor{green!15}\textit{Universalized}}
& \multicolumn{1}{c}{\cellcolor{green!35}\textit{Universal}}\\
& \cellcolor{gray!15}EA & \cellcolor{gray!15}DA & \cellcolor{gray!15}SA & \cellcolor{gray!15}ES
& \cellcolor{green!15}Enc. & \cellcolor{green!15}Emb. & \cellcolor{green!15}Map & \cellcolor{green!35}Ours \\
\Xhline{0.1mm}
PSNR $\downarrow$ & 16.12 & 16.14 & 19.60& \textbf{14.42} & 19.13 & 17.47 & 18.64 & \underline{15.49} \\
SSIM $\downarrow$ & 0.513 & 0.476 & 0.612 & \textbf{0.400} & 0.555 & 0.500 & 0.503 & \underline{0.457} \\
VIFp $\downarrow$ & \underline{0.145} & \underline{0.145} & 0.266 & \textbf{0.148} & 0.202 & 0.203 & 0.197 & 0.166 \\
FSIM $\downarrow$ & \underline{0.730} & 0.756 & 0.823 & \textbf{0.692} & 0.811 & 0.780 & 0.795 & 0.735 \\
LPIPS $\uparrow$  & 0.554 & 0.558 & 0.425 & \textbf{0.588} & 0.489 & 0.511 & 0.516 & \underline{0.568} \\
Feat. Sim. (C) $\downarrow$ & 0.717 & 0.757 & 0.769 & \textbf{0.688} & 0.801 & 0.734 & 0.768 & \underline{0.707} \\
\Xhline{0.35mm}
\end{tabular}

}
\label{tab:editshield}
\vspace{-6mm}
\end{table}

\subsection{Comparison with EditShield on Instruction-guided Editing}
We compare our universal immunization method against EditShield~\cite{editshield}, specified for instruction-guided image editing, on InstructPix2Pix (IP2P)~\cite{brooks2023instructpix2pix}.
For evaluation, we sample 100 image and instruction pairs on MagicBrush~\cite{magic}.
Since EditShield~\cite{editshield} specifically targets IP2P~\cite{brooks2023instructpix2pix}, EditShield (ES) shows remarkable performance against instruction-guided editing pipeline in Table~\ref{tab:editshield}.
However, as shown in Table~\ref{edit_black}, in black-box settings on Stable Diffusion V1.4 and V2.0, while EditShield exhibits immunization transferability, our method consistently maintains superior performance on black-box settings.
Although strong immunization against instruction-guided editing is an important advantage, robustness to unseen models is also a crucial practical requirement, as the defense should transfer across diverse diffusion editing models.
\begin{table}[t!]
\footnotesize
\setlength{\tabcolsep}{3pt}
\centering
\caption{Black-box immunization transferability on Stable Diffusion V1.4 and V2.0~\cite{rombach2022high}. 
The surrogate model for each method is Stable Diffusion V1.5.}
\label{edit_black}

\resizebox{\linewidth}{!}{
\begin{tabular}{c|c|cccc|c|c|cccc|c}
\Xhline{0.35mm}
{Model}
& \multicolumn{6}{c|}{Stable Diffusion V1.4} 
& \multicolumn{6}{c}{Stable Diffusion V2.0} \\
\Xhline{0.1mm}
\multirow{2}{*}{Method}  
& \multirow{2}{*}{Clean}
& \multicolumn{4}{c|}{\cellcolor{gray!15}\textit{Image-specific}}  
& \multicolumn{1}{c|}{\cellcolor{green!35}\textit{Universal}}   
& \multirow{2}{*}{Clean}
& \multicolumn{4}{c|}{\cellcolor{gray!15}\textit{Image-specific}}  
& \multicolumn{1}{c}{\cellcolor{green!35}\textit{Universal}} \\
& & \cellcolor{gray!15}EA & \cellcolor{gray!15}DA & \cellcolor{gray!15}SA & \cellcolor{gray!15}ES & \cellcolor{green!35}Ours
& & \cellcolor{gray!15}EA & \cellcolor{gray!15}DA & \cellcolor{gray!15}SA & \cellcolor{gray!15}ES & \cellcolor{green!35}Ours \\

\Xhline{0.1mm}
PSNR $\downarrow$ 
& -- & 15.46 & 15.54 & \underline{14.91} & 16.24 & \textbf{14.18}
& -- & \underline{12.48} & 12.73 & 12.79 & 13.74 & \textbf{12.17} \\
SSIM $\downarrow$ 
& -- & 0.444 & 0.457 & \underline{0.336} & 0.408 & \textbf{0.332}
& -- & \underline{0.298} & 0.304 & 0.271 & 0.317 & \textbf{0.240} \\
VIFp $\downarrow$ 
& -- & 0.119 & 0.143 & \underline{0.088} & 0.141 & \textbf{0.083}
& -- & 0.074 & 0.085 & \underline{0.060} & 0.092 & \textbf{0.057} \\
FSIM $\downarrow$ 
& -- & \underline{0.664} & 0.695 & 0.666 & 0.691 & \textbf{0.642}
& -- & \underline{0.558} & 0.586 & 0.632 & 0.608 & \textbf{0.555} \\
LPIPS $\uparrow$  
& -- & 0.545 & 0.507 & \underline{0.549} & 0.510 & \textbf{0.604}
& -- & \underline{0.634} & 0.611 & 0.626 & 0.593 & \textbf{0.660} \\
Feat. Sim. (C) $\downarrow$
& 0.743 & \textbf{0.663} & 0.675 & 0.701 & 0.704 & \underline{0.675}
& 0.694 & \textbf{0.606} & 0.628 & 0.656 & 0.651 & \underline{0.616} \\
\Xhline{0.35mm}
\end{tabular}
}
\end{table}
\begin{table}[t]
\setlength{\tabcolsep}{3pt}
\centering
\caption{Robustness evaluation against various purification methods compared with image-specific and optimization-free methods.}
\vspace{-0.2cm}
\resizebox{\linewidth}{!}{
\begin{tabular}{c|c|ccc|c|c|ccc|c|c|ccc|c|c|ccc|c|c}
\Xhline{0.35mm}
{Purification}
& \multirow{3}{*}{\makecell{Clean \\ Sim.}}
& \multicolumn{5}{c|}{JPEG compression} 
& \multicolumn{5}{c|}{GrIDPure}
& \multicolumn{5}{c|}{Conditional DiffPure}
& \multicolumn{5}{c}{Noisy Upscaling} \\
\Xcline{1-1}{0.1mm}
\Xcline{3-22}{0.1mm}

\multirow{2}{*}{Method}
& 
& \multicolumn{3}{c|}{\cellcolor{gray!15}\textit{Image-specific}} & \cellcolor{cyan!15}\textit{Opt-free} & \cellcolor{green!35}\textit{Universal}
& \multicolumn{3}{c|}{\cellcolor{gray!15}\textit{Image-specific}} & \cellcolor{cyan!15}\textit{Opt-free} & \cellcolor{green!35}\textit{Universal}
& \multicolumn{3}{c|}{\cellcolor{gray!15}\textit{Image-specific}} & \cellcolor{cyan!15}\textit{Opt-free} & \cellcolor{green!35}\textit{Universal}
& \multicolumn{3}{c|}{\cellcolor{gray!15}\textit{Image-specific}} & \cellcolor{cyan!15}\textit{Opt-free} & \cellcolor{green!35}\textit{Universal} \\

& 
& \cellcolor{gray!15}EA & \cellcolor{gray!15}DA & \cellcolor{gray!15}SA & \cellcolor{cyan!15}FP & \cellcolor{green!35}Ours
& \cellcolor{gray!15}EA & \cellcolor{gray!15}DA & \cellcolor{gray!15}SA & \cellcolor{cyan!15}FP & \cellcolor{green!35}Ours
& \cellcolor{gray!15}EA & \cellcolor{gray!15}DA & \cellcolor{gray!15}SA & \cellcolor{cyan!15}FP & \cellcolor{green!35}Ours
& \cellcolor{gray!15}EA & \cellcolor{gray!15}DA & \cellcolor{gray!15}SA & \cellcolor{cyan!15}FP & \cellcolor{green!35}Ours \\
\Xhline{0.1mm}

PSNR $\downarrow$
& --
& \underline{15.95} & 15.96 & 17.40 & 17.14 & \textbf{15.88}
& \textbf{17.71} & \underline{17.85} & 18.39 & 19.22 & 17.90
& 15.83 & \underline{15.72} & 16.68 & 16.69 & \textbf{14.99}
& 17.67 & \underline{17.53} & 17.99 & 17.97 & \textbf{17.20} \\

SSIM $\downarrow$
& --
& \underline{0.448} & 0.449 & 0.507 & 0.575 & \textbf{0.433}
& \underline{0.537} & \textbf{0.532} & 0.573 & 0.614 & 0.545
& 0.435 & \underline{0.423} & 0.482 & 0.478 & \textbf{0.371}
& 0.534 & \underline{0.533} & 0.552 & 0.550 & \textbf{0.482} \\

VIFp $\downarrow$
& --
& \textbf{0.128} & 0.140 & 0.175 & 0.220 & \underline{0.129}
& \textbf{0.191} & \underline{0.194} & 0.215 & 0.250 & 0.197
& \underline{0.125} & 0.128 & 0.155 & 0.156 & \textbf{0.102}
& \underline{0.180} & 0.184 & 0.191 & 0.192 & \textbf{0.165} \\

FSIM $\downarrow$
& --
& \textbf{0.687} & 0.703 & 0.740 & 0.763 & \underline{0.697}
& \textbf{0.753} & \underline{0.757} & 0.773 & 0.791 & 0.758
& \underline{0.683} & 0.687 & 0.722 & 0.715 & \textbf{0.664}
& \underline{0.749} & 0.751 & 0.758 & 0.758 & \textbf{0.732} \\

LPIPS $\uparrow$
& --
& \underline{0.502} & 0.484 & 0.429 & 0.377 & \textbf{0.511}
& 0.407 & \textbf{0.417} & 0.377 & 0.340 & \underline{0.410}
& \underline{0.506} & 0.500 & 0.453 & 0.438 & \textbf{0.568}
& \underline{0.400} & 0.398 & 0.387 & 0.382 & \textbf{0.427} \\

Feat. Sim. (C) $\downarrow$
& 0.744
& \underline{0.696} & \textbf{0.692} & 0.713 & 0.717 & 0.698
& 0.719 & \textbf{0.711} & 0.729 & 0.733 & \underline{0.717}
& \underline{0.691} & 0.693 & 0.708 & 0.707 & \textbf{0.683}
& \underline{0.726} & \textbf{0.723} & 0.729 & 0.727 & \textbf{0.723} \\


\Xhline{0.35mm}
\end{tabular}
}
\label{robustness-specific}
\vspace{-0.2cm}
\end{table}
\subsection{Robustness Evaluation against Purification with Image-specific and Optimization-free Methods}
We further evaluate the robustness against perturbation purification techniques comparison with image-specific and optimization-free methods.
As shown in Table~\ref{robustness-specific}, our UAP demonstrates competitive robustness overall, and in particular, it remains highly effective under diffusion-based purification, achieving comparable or stronger performance than image-specific and optimization-free approaches.
These results suggest that, although our method uses a single universal perturbation, it achieves competitive robustness even when compared with input-specific, fine-grained, and more visually perceptible (in Table~\textcolor{red}{4} of manuscript) methods.
\subsection{Analysis for Imperceptibility}
{Universal adversarial perturbations generate a single perturbation to deceive multiple images, which makes them more susceptible to visible artifacts compared to image-specific approaches that leverage individual target images.
To examine the trade-off between immunization performance and imperceptibility, we further evaluate different perturbation budgets ($\epsilon=8,10,16$), measuring imperceptibility with DISTS~\cite{dists}, PSNR, and LPIPS~\cite{lpips}, alongside standard immunization evaluation metrics.
}
As shown in Table~\ref{tab:tradeoff}, our method achieves a comparable level of imperceptibility to image-specific baselines at smaller budgets (\eg, $\epsilon=8/255$ and $\epsilon=10/255$), while at the same time delivering superior immunization performance.
These results highlight that our approach effectively balances perceptual quality and robustness without incurring additional test-time costs.

\begin{table}[t]
\footnotesize
\centering
\caption{Trade-off between immunization performance and imperceptibility.}
\vspace{-0.2cm}
\resizebox{0.7\linewidth}{!}{
\begin{tabular}{c|c|c|ccc|ccc}
\Xhline{0.35mm}
\multirow{2}{*}{Category} & \multirow{2}{*}{Method} & \multirow{1}{*}{Clean}
& \multicolumn{3}{c|}{\cellcolor{gray!15}\textit{Image-specific}}
& \multicolumn{3}{c}{\cellcolor{green!35}\textit{Universal}} \\
& & &
\cellcolor{gray!15}EA & \cellcolor{gray!15}DA & \cellcolor{gray!15}SA
& \cellcolor{green!35}Ours$_8$ & \cellcolor{green!35}Ours$_{10}$ & \cellcolor{green!35}Ours$_{16}$ \\
\Xhline{0.1mm}
\multirow{6}{*}{Performance} 
 & PSNR $\downarrow$ & --&14.75 & 14.71 & 16.28 & 14.58 & \underline{14.19} & \textbf{13.31} \\
 & SSIM $\downarrow$ & --&0.386 & 0.382 & 0.431 & 0.363 & \underline{0.332} & \textbf{0.277}\\
 & VIFp $\downarrow$ & --&0.088 & 0.106 & 0.138 & 0.095 & \underline{0.082} & \textbf{0.061} \\
 & FSIM $\downarrow$ & --&\underline{0.637} & 0.660 & 0.714 & 0.656 & \underline{0.642} & \textbf{0.612} \\
 &  LPIPS $\uparrow$  & --&0.584 & 0.552 & 0.490 & 0.574 & \underline{0.606} & \textbf{0.677} \\
& Feat. sim.(C) $\downarrow$ &0.744& 0.677& 0.662& 0.705 & 0.681& 0.673& 0.658\\
\Xhline{0.1mm}
\multirow{3}{*}{Imperceptibility} 
 & DISTS $\downarrow$ & --& 0.311 & 0.243 & \underline{0.164}  & \textbf{0.157} & 0.181 & 0.246 \\
 & PSNR $\uparrow$    & --& 28.67 & 27.78 & \underline{30.82}  & \textbf{31.37} & 29.41 & 25.29 \\
 & LPIPS $\downarrow$    & --& 0.471 & 0.428 & \textbf{0.301} & \underline{0.306} & 0.355 & 0.463 \\
\Xhline{0.35mm}
\end{tabular}
}
\label{tab:tradeoff}
\vspace{-0.4cm}
\end{table}

\begin{figure}[t!]
\centering    \includegraphics[width=\linewidth]{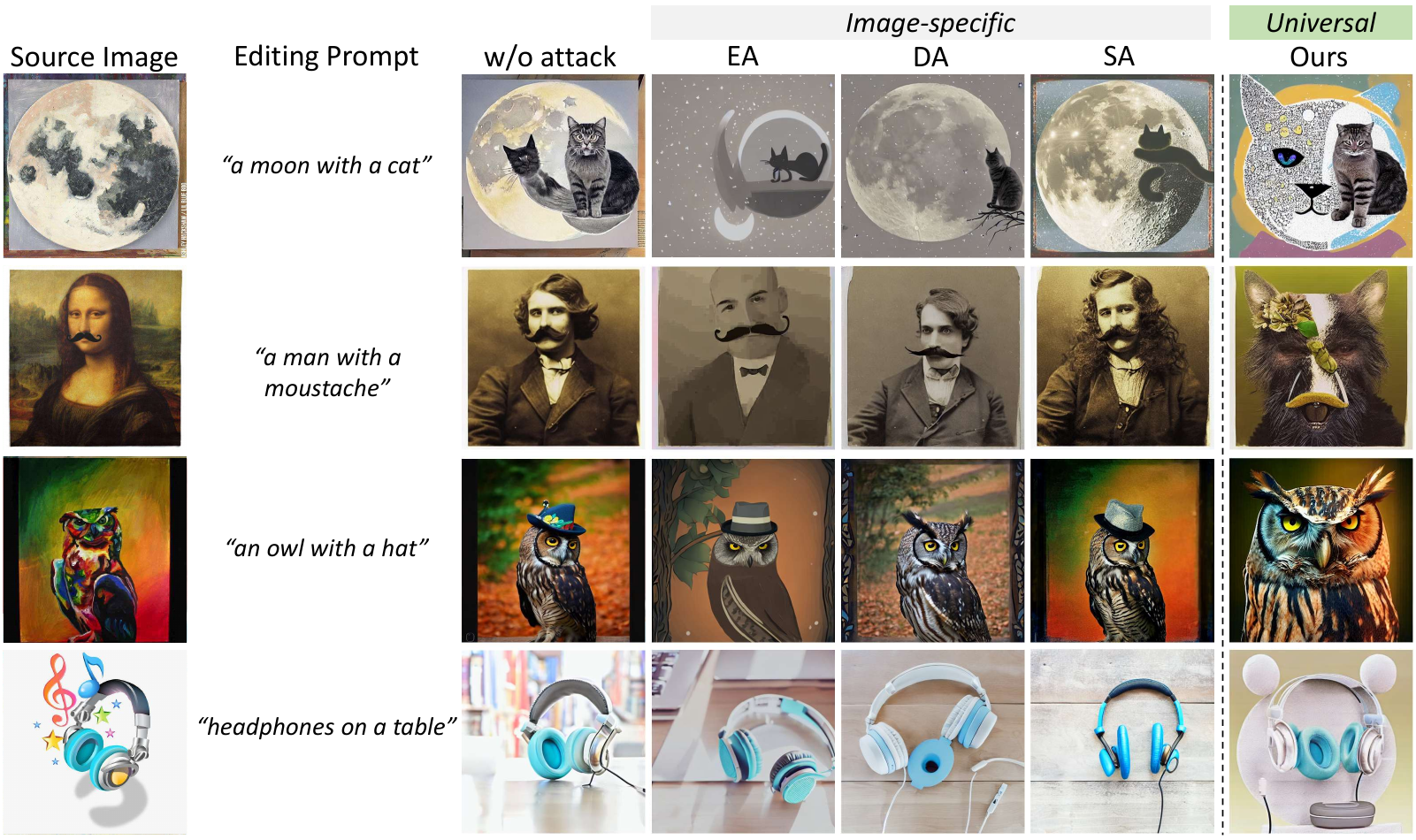}    
    \caption{Editing examples of DomainNet images used in the user study after applying immunization across different methods.
    }
    \vspace{0.4cm}
    \label{fig:ood}

\centering    \includegraphics[width=\linewidth]{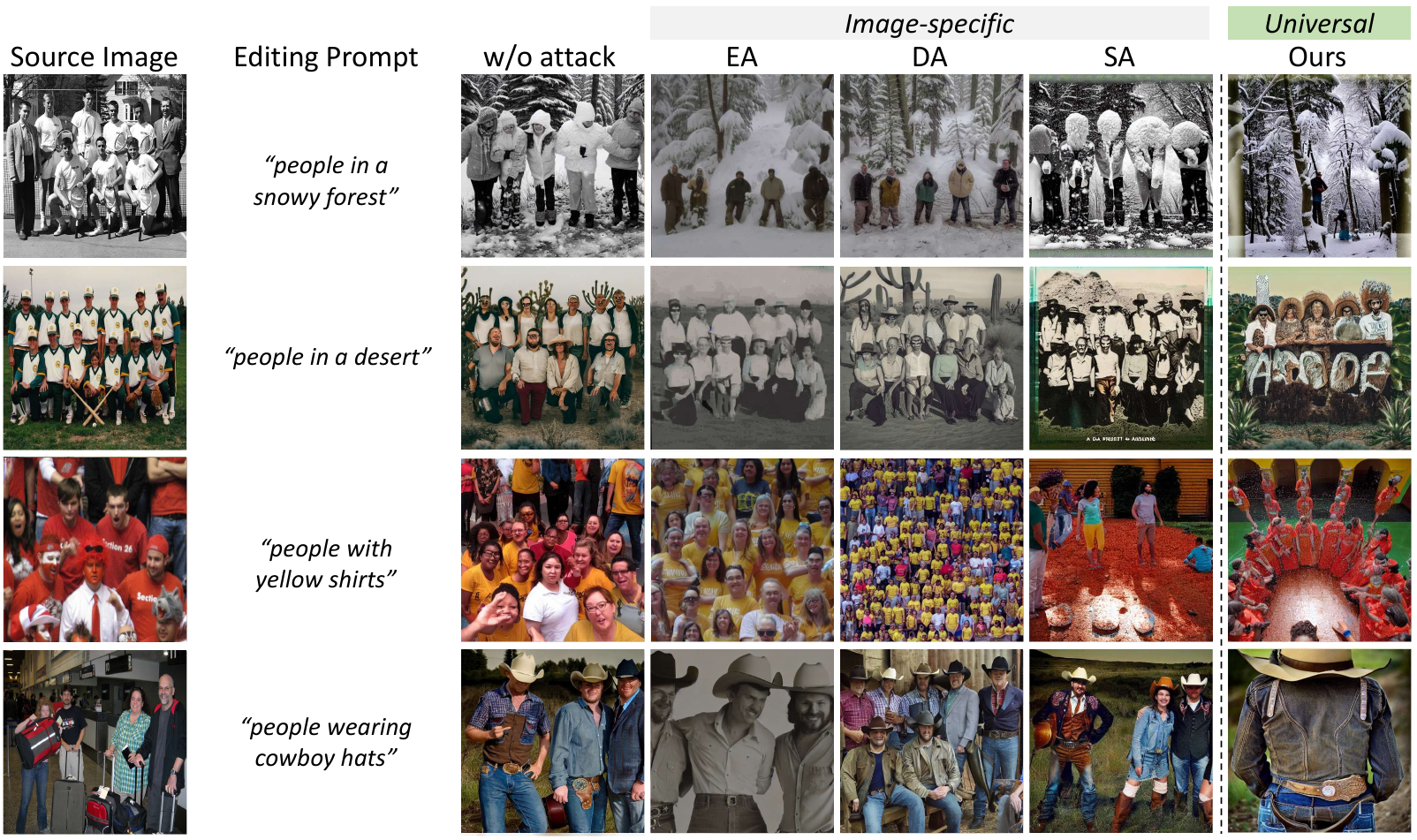}    
    \caption{Editing examples of MS-COCO images used in the user study after applying immunization across different methods.
    }
    \vspace{-0.3cm}
    \label{fig:complex}
\end{figure}

\subsection{Human Evaluation}
{To practically and comprehensively evaluate the utility of our method, we conducted a user study comparing it with image-specific approaches similar to previous works~\cite{trimotion, trippodo2025immunizing}.
Since we are the first to propose a universal immunization strategy, these serve as the appropriate baselines. 
The evaluation was performed across three dimensions: imperceptibility, effectiveness, and practicality.}
The study uses 20 representative images: 10 sampled from DomainNet~\cite{domain} (painting and cartoon domains, representing out-of-distribution cases) and 10 from MS-COCO~\cite{mscoco} (complex, multi-object natural scenes).
Each image is perturbed using five methods: Encoder Attack (EA), Diffusion Attack (DA), Semantic Attack (SA), and our approach with perturbation budgets of $\epsilon=10/255$ and $\epsilon=16/255$.
A total of 20 participants were recruited via Amazon Mechanical Turk (MTurk), with each participant evaluating all perturbed samples in randomized order.
For imperceptibility, participants were asked to identify the samples that appeared most visually conspicuous or artifact-heavy, where a higher score indicates that perturbations were more noticeable and thus less imperceptible.
For effectiveness, participants ranked the perturbed outputs according to how well they disrupted the intended semantic edit, using a scale from 1 to 5, where higher ranks correspond to stronger immunization.
Finally, for practicality, participants rated their willingness to adopt each method in real-world security or safety contexts on a 5-point Likert scale ranging from 1 (not likely) to 5 (very likely), reflecting their subjective judgment of deployability.
\vspace{-0.2cm}
\begin{table}[h!]
\vspace{-4mm}
\footnotesize
\centering
\caption{User study results on imperceptibility and effectiveness. 
Imperceptibility is measured as the percentage of votes (lower is better), 
and effectiveness is measured as the mean ranking score (higher is better).}
\vspace{-1mm}
\resizebox{0.9\linewidth}{!}{
\begin{tabular}{l|ccc|cc}
\Xhline{0.35mm}
\multirow{2}{*}{Method}
& \multicolumn{3}{c|}{\cellcolor{gray!15}\textit{Image-specific}}
& \multicolumn{2}{c}{\cellcolor{green!35}\textit{Universal}} \\
& \cellcolor{gray!15}EA & \cellcolor{gray!15}DA & \cellcolor{gray!15}SA
& \cellcolor{green!35}Ours$_{10}$ & \cellcolor{green!35}Ours$_{16}$ \\
\Xhline{0.1mm}
Imperceptibility (\% voting $\downarrow$) & 46.75 & 44.25 & \textbf{27.00} & \underline{38.00} & 47.00 \\
Effectiveness (mean score $\uparrow$) & 2.49$\pm$1.33 & 2.23$\pm$1.02 & 2.25$\pm$1.12 & \underline{3.86$\pm$0.98} & \textbf{4.20$\pm$1.13} \\
\Xhline{0.35mm}
\end{tabular}
}
\label{tab:user_1}
\vspace{-5mm}
\end{table}
\vspace{-0.5cm}
\begin{table}[h!]
\footnotesize
\centering
\caption{User study results on practicality. 
Participants rated their willingness to adopt our method in security contexts on a 5-point Likert scale.}
\vspace{-2mm}
\resizebox{0.4\linewidth}{!}{
\begin{tabular}{lccccc}
\toprule
Score & 1 & 2 & 3 & 4 & 5 \\
\midrule
Practicality (\% voting) & 9 & 27 & 10 & 25 & 29 \\
\bottomrule
\vspace{-1cm}
\end{tabular}
}
\label{tab:user_2}

\end{table}


\begin{table}[h!]
\footnotesize
\centering

\caption{Quantitative results on MS-COCO and DomainNet. The best result is shown in bold; the second-best is underlined.}
\resizebox{0.9\linewidth}{!}{
\begin{tabular}{c|c|ccc|c|c|ccc|c}
\Xhline{0.35mm}
 
{Dataset}& \multicolumn{5}{c|}{MS-COCO} 
& \multicolumn{5}{c}{DomainNet} \\
\Xhline{0.1mm}
\multirow{2}{*}{Method}
& \multirow{2}{*}{Clean}
& \multicolumn{3}{c|}{\cellcolor{gray!15}\textit{Image-specific}}
& \multicolumn{1}{c|}{\cellcolor{green!35}\textit{Universal}}
& \multirow{2}{*}{Clean}
& \multicolumn{3}{c|}{\cellcolor{gray!15}\textit{Image-specific}}
& \multicolumn{1}{c}{\cellcolor{green!35}\textit{Universal}}\\
& & \cellcolor{gray!15}EA & \cellcolor{gray!15}DA & \cellcolor{gray!15}SA & \cellcolor{green!35}Ours & & \cellcolor{gray!15}EA & \cellcolor{gray!15}DA & \cellcolor{gray!15}SA & \cellcolor{green!35}Ours\\
\Xhline{0.1mm}
PSNR $\downarrow$ & --& 14.98 & \underline{14.59} & 16.10 & \textbf{14.16}   
                  & --& 14.63 & \textbf{14.24} & 16.66 & \underline{14.55} \\
SSIM $\downarrow$ & --& 0.352 & \underline{0.327} & 0.366 & \textbf{0.267}   
                  & --& 0.443 & \underline{0.419} & 0.444 & \textbf{0.361}  \\
VIFp $\downarrow$ & --& \underline{0.066} & 0.078 & 0.115 & \textbf{0.060} 
                  & --& \textbf{0.079} & 0.111 & 0.130 & \underline{0.083} \\
FSIM $\downarrow$ & --& \textbf{0.615} & 0.659 & 0.722 & \underline{0.648}  
                  & --& \textbf{0.626} & 0.650 & 0.720 & \underline{0.628}  \\
LPIPS $\uparrow$  & --& \textbf{0.676}  & 0.603 & 0.558 & \underline{0.650}  
                  & --& \underline{0.618} & 0.596 & 0.539 & \textbf{0.638}  \\
Feat. Sim. (C) $\downarrow$ & 0.580 & 0.505 & 0.522 & \textbf{0.487} & \underline{0.491} 
& 0.622 & \textbf{0.581} & \underline{0.591}  & 0.607 & \underline{0.591}\\
\Xhline{0.35mm}

\end{tabular}
}
\label{tab:coco_domainnet}

\end{table}

The human evaluation result in Table~\ref{tab:user_1} reflects a trade-off between invisibility and effectiveness.
Interestingly, a similar trend is also observed in the quantitative measurements reported in Table~\ref{tab:tradeoff}.
Notably, at $\epsilon=10/255$, our method strikes a strong balance, demonstrating both high immunization effectiveness and improved imperceptibility as confirmed by human evaluation compared to image-specific baselines.
It also receives favorable practicality scores in Table~\ref{tab:user_2}, suggesting strong potential for real-world applicability.
We additionally provide a visualization of the examples used in the human evaluation in Figure~\ref{fig:ood} and~\ref{fig:complex}.

\subsection{Robustness on Diverse Domains}

We evaluate the generality of our method across diverse image types using MS-COCO and DomainNet.
MS-COCO contains complex multi-object scenes, while DomainNet serves as an out-of-distribution dataset since it was not seen during UAP training.
In Table~\ref{tab:coco_domainnet}, our method achieves superior performance even when compared to immunization processes that directly leverage source images with a larger perturbation budget of $\epsilon=16/255$ than ours ($\epsilon=10/255$).
This highlights the robustness of our approach across diverse data domains.



\subsection{Convergence Behavior and Quantitative Analysis}
We provide training curves for target semantic injection loss and source semantic suppression loss across training epochs to verify optimization stability.
As shown in Figure~\ref{fig:loss}, both losses converge stably during training without interfering with each other.
\begin{figure}[h!]
\centering    \includegraphics[width=\linewidth]{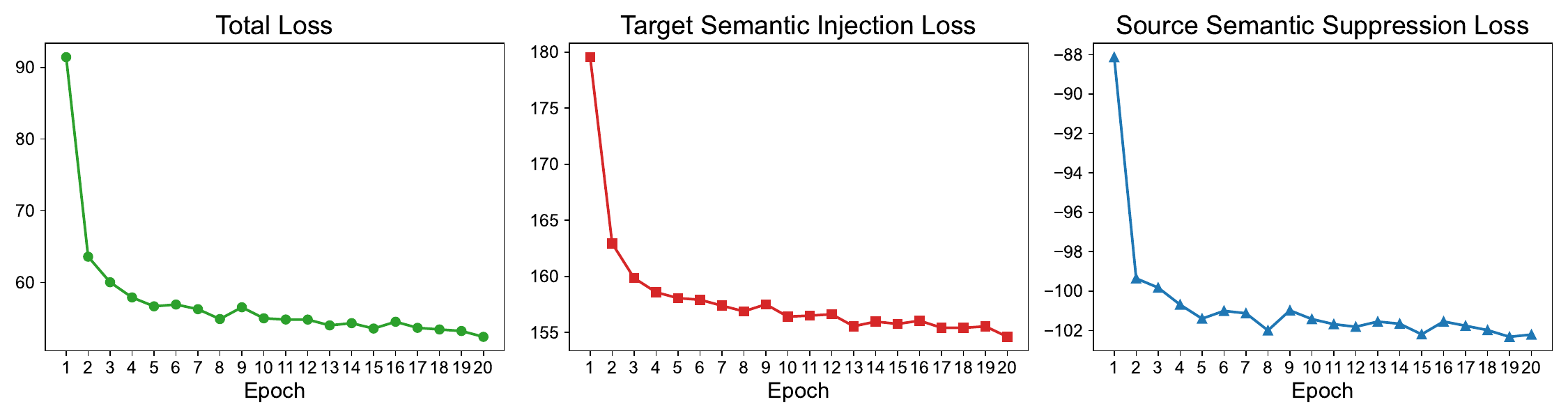}  
    \caption{Visualization of loss values across training epochs.
    }
    \vspace{-0.3cm}
    \label{fig:loss}
\end{figure}

We also report the ablation results in Table~\ref{R1_W4}, including variants trained with each loss individually and with their combination.
The results show that the suppression loss alone provides a meaningful degree of immunization by weakening source semantics.
However, the primary strength of our universal approach stems from semantic injection: the injection loss yields stronger immunization, and combining both losses leads to the best overall performance.
This is because the injection loss promotes the target semantics, whereas the suppression loss reduces the source semantics of the training data, making the two objectives complementary rather than contradictory.

\begin{table}[t!]
\footnotesize
\centering
\caption{Ablation study for loss functions. Best results are shown in bold; second-best are underlined.}
\resizebox{0.5\linewidth}{!}{
\begin{tabular}{c|c|ccc}
\Xhline{0.35mm}
Loss Function & Clean & $\mathcal{L}_\text{sup}$ & $\mathcal{L}_\text{inj}$ & $\mathcal{L}_\text{inj}+ \mathcal{L}_\text{sup}$ \\
\Xhline{0.1mm}

PSNR $\downarrow$
& --          
& 16.11
& \underline{14.41}
& \textbf{14.19} \\

SSIM $\downarrow$
& --          
& 0.420
& \underline{0.367}
& \textbf{0.332} \\

VIFp $\downarrow$
& --          
& 0.140
& \underline{0.096}
& \textbf{0.082} \\

FSIM $\downarrow$
& --          
& 0.697
& \underline{0.655}
& \textbf{0.642} \\

LPIPS $\uparrow$
& --          
& 0.490
& \underline{0.585}
& \textbf{0.606} \\

Feat. Sim. (CLIP) $\downarrow$
& 0.744      
& 0.688
& \underline{0.680}
& \textbf{0.673} \\


\Xhline{0.35mm}
\end{tabular}
}
\label{R1_W4}
\end{table}

\section{Additional Ablation Study}

\begin{figure}[t!]
\centering    \includegraphics[width=\linewidth]{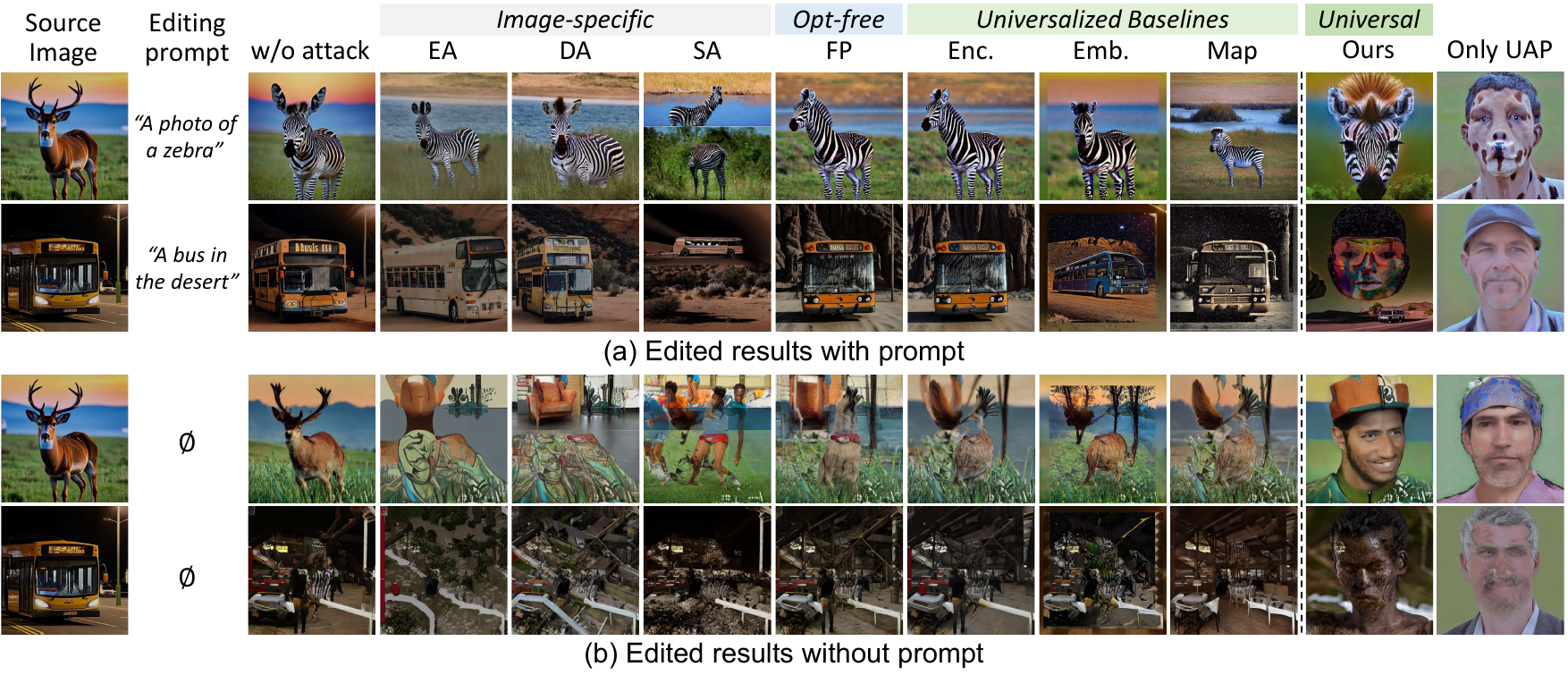} 
    \caption{Visualization of how perturbations from image-specific, optimization-free, and universalized baselines affect edited results with and without prompt guidance. (a) Editing results with the prompt. (b) Editing results under a null prompt ($\phi$). Our targeted UAP generates target-like content even without text guidance.
    }
    \vspace{0.1cm}
    \label{fig:null_1}

\centering    \includegraphics[width=\linewidth]{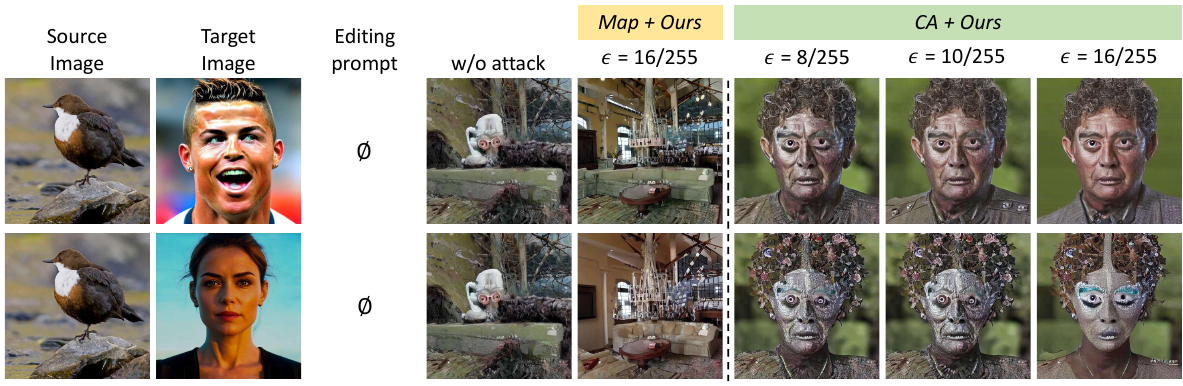} 
    \caption{Qualitative comparison under null-prompt editing between applying semantic injection/suppression losses to the attention map (\textit{Map+Ours}) and to the cross-attention output (\textit{CA}). CA-based injection restores the target concept without text guidance, whereas Map-based injection merely introduces distortion and fails to reliably induce target content.
    }

    \label{fig:null_2}
\end{figure}

\subsection{Effect of Utilizing Cross-attention Outputs}
\label{app:ca-map}
We conduct experiments to comprehensively analyze how our UAP influences diffusion-based editing behavior.
Specifically, as shown in Figure~\ref{fig:null_1}, to verify whether the UAP successfully injects target semantics, we perform editing without prompts that could alter the injected target semantics.
In Figure~\ref{fig:null_1}(a), the prompt intent is reflected in the edited results.
However, without text guidance, the diffusion model fails to reliably reconstruct the source image (\eg, \textit{w/o attack}), often producing only partial recovery (\eg, deer) or even an unrelated image (\eg, bus), as shown in Figure~\ref{fig:null_1}(b).
Under immunization, existing image-specific methods, optimization-free methods, and universalized baselines mainly produce distortions relative to results of \textit{w/o attack}.
Meanwhile, our target-semantic UAP still recovers the target content, indicating that it encodes dominant target semantics and steers the editing dynamics away from the source contents, even without text guidance.
These results further demonstrate that our method enables effective immunization at inference time without requiring a target prompt.
\begin{table}[t!]
\footnotesize
\centering
\caption{Comparison between using attention map and cross-attention with different loss functions. * denotes our proposed method.}
\resizebox{0.65\linewidth}{!}{
\begin{tabular}{c|c|ccc|ccc}
\Xhline{0.35mm}
\multirow{2}{*}{Method} & \multirow{2}{*}{Clean} 
& \multicolumn{3}{c|}{Map} 
& \multicolumn{3}{c}{CA*} \\
\cline{3-8}
& 
& \texttt{Sup} & \texttt{Inj} & \texttt{Inj+Sup} 
& \texttt{Sup} & \texttt{Inj} & \texttt{Inj+Sup}* \\
\Xhline{0.1mm}

PSNR $\downarrow$
& -- 
&16.16
& 15.59
& 15.45
&16.11
& \underline{14.41}
& \textbf{14.19} \\

SSIM $\downarrow$
& --
&0.468
& 0.358
& \underline{0.335}
&0.420
& 0.367
& \textbf{0.332} \\

VIFp $\downarrow$
& -- 
&0.152
& 0.121
& 0.115
&0.140
& \underline{0.096}
& \textbf{0.082} \\

FSIM $\downarrow$
& --
&0.710
& 0.686
& 0.679
&0.697
& \underline{0.655}
& \textbf{0.642} \\

LPIPS $\uparrow$
& --
&0.465
& 0.537
& 0.550
&0.490
& \underline{0.585}
& \textbf{0.606} \\

Feat. Sim. (CLIP) $\downarrow$
& 0.744
&0.708
& 0.702
& 0.696
&0.688
& \underline{0.680}
& \textbf{0.673} \\


\Xhline{0.35mm}
\end{tabular}
}
\label{R3_W2}
\end{table}

Also, as described in Section~\textcolor{red}{4.3} and~\ref{sec:justification}, a cross-attention block responsible for image-prompt alignment consists of an attention map (MAP) and a cross-attention output (CA), and our method applies semantic injection/suppression to the cross-attention output.
To validate this design choice, we additionally apply Eq.~(\textcolor{red}{5}) and (\textcolor{red}{6}) to the attention map and examine whether target semantic injection is successfully achieved.
As shown in Figure~\ref{fig:null_2}, the $\mathrm{Map}$ variant distorts the edited results compared to the non-immunized case, but does not induce the target-like content.
In contrast, our $\mathrm{CA}$ successfully restores the target concept (\eg, man or woman), even under a smaller perturbation budget ($\epsilon=8/255$).
Furthermore, Table~\ref{R3_W2} shows that applying semantic injection and suppression losses to the cross-attention output consistently yields stronger immunization performance than applying them to the attention map (Map).
These qualitative and quantitative results together demonstrate that applying target semantic injection and suppression to the cross-attention output is a key component for achieving our semantic injection objective.


\subsection{Impact of Diverse Constraint Parameter}
\begin{figure}[h]
\centering \includegraphics[width=\linewidth]{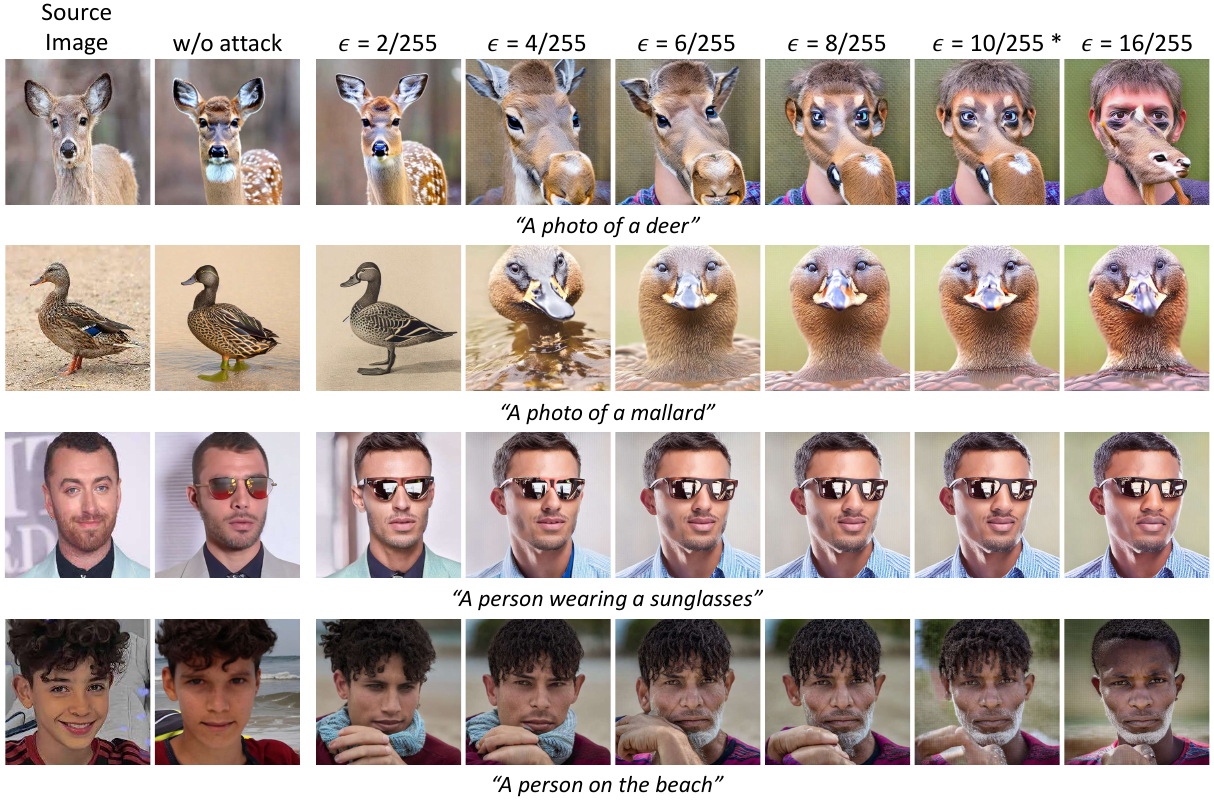}
    \caption{Qualitative comparison between different constraint parameters. $\epsilon=10/255$* is used for our main experiments.
    }
    \label{fig:epsilon_quali}
\end{figure}

\noindent
To investigate the effect of the $\ell_\infty$-norm constraint parameter $\epsilon$ used in adversarial perturbations on immunization performance, we conduct experiments with $\epsilon=8/255$ and 16/255.
{As shown in Table~\ref{tab:tradeoff}, even with a tighter budget of $\epsilon=8/255$, our method achieves competitive or superior performance compared to image-specific baselines that use a larger budget of $\epsilon=16/255$, along with higher imperceptibility, thereby striking a favorable balance between immunization performance and visual quality.

We also visualize the editing results in Figure~\ref{fig:epsilon_quali} by slightly increasing $\epsilon$ from 2/255 to 16/255, where the UAP is trained on the target `\textit{Ronaldo}'.
Notably, from around $\epsilon=4/255$, our method already produces sufficiently source-agnostic results, indicating that the diffusion process is conditioned more on the injected target semantics than on the original source content.
Interestingly, when the editing prompts (\eg, `\textit{person}') do not directly conflict with the target concept (\eg, `\textit{Ronaldo}'),
the edited outputs often exhibit target-related traits (in Figure~\ref{fig:tar_uap_vis}), such as tanned skin or the appearance of a middle-aged man.
This suggests that our UAP captures not only the shape or structural cues of the target image used during training, but also attribute-level information such as age and skin tone.




\vspace{-3mm}
\subsection{Ablation Studies for Diverse Targets}
\label{app:diverse_quan}
We conduct ablation studies across diverse targets using three variants of the proposed immunization approach: target injection in data-free setting (\texttt{Inj$_\text{DF}$}), target semantic injection with real data (\texttt{Inj}), and injection combined with source semantic suppression (\texttt{Inj}+\texttt{Sup}), across various target contents.
As shown in Table~\ref{ronaldo},~\ref{sunflower},~\ref{peacock}, and~\ref{mandala_}, the results consistently show that \texttt{Inj} outperforms \texttt{Inj$_\text{DF}$}, and that \texttt{Inj}+\texttt{Sup} further improves performance over \texttt{Inj}.
Importantly, despite relying on pre-computed UAPs with zero test-time cost, our method remains competitive with image-specific baselines.
This demonstrates that our \textit{semantic injection loss}, Eq.~(\textcolor{red}{5}), and \textit{source semantic suppression loss}, Eq.~(\textcolor{red}{6}), are not only effective across specific examples but also robust and generalizable to arbitrary and diverse target object classes.
{Furthermore, \texttt{Inj}+\texttt{Sup} consistently outperforms the universalized baselines in Table~\textcolor{red}{1}, and the comparable or even superior performance of the data-free variant \texttt{Inj$_\text{DF}$} further highlights both the effectiveness of semantic injection and its practical advantage of not requiring real-world data.}

\begin{table}[t!]
\centering
\caption{Ablation study of loss components on black-box transferability with diverse target selections. * denotes the white-box model.}
\begin{subtable}{\linewidth}

\footnotesize
\centering
\caption{Ablation study for the ``Ronaldo'' target across different diffusion models.}
\resizebox{\linewidth}{!}{
\begin{tabular}{c|c|ccc|c|ccc|c|ccc|c|ccc}
\Xhline{0.35mm}
{Model}
& \multicolumn{4}{c|}{Stable Diffusion V1.4}
& \multicolumn{4}{c|}{Stable Diffusion V1.5*} 
& \multicolumn{4}{c|}{Stable Diffusion V2.0}
& \multicolumn{4}{c}{InstructPix2Pix}\\
\Xhline{0.1mm}
Method & Clean & Inj$_{\text{DF}}$ & Inj & Inj+Sup 
       & Clean &Inj$_{\text{DF}}$ & Inj & Inj+Sup 
       & Clean &Inj$_{\text{DF}}$ & Inj & Inj+Sup 
       & Clean &Inj$_{\text{DF}}$ & Inj & Inj+Sup \\
\Xhline{0.1mm}
PSNR $\downarrow$ & --& 15.42 & 14.98 & {14.78}
                  & --& 15.56 & 14.99 & {14.69}
                  & --& 12.97 & 12.65 & {12.50}
                  & --& 16.13 & 15.34 & {15.03} \\
SSIM $\downarrow$ & --& 0.403 & 0.388 & {0.364}
                  & --& 0.447 & 0.421 & {0.383}
                  & --& 0.294 & 0.288 & {0.273}
                  & --& 0.530 & 0.474 & {0.452} \\
VIFp $\downarrow$ & --& 0.110 & 0.098 & {0.086}
                  & --& 0.126 & 0.117 & {0.095}
                  & --& 0.065 & 0.061 & {0.057}
                  & --& 0.182 & 0.154 & {0.141} \\
FSIM $\downarrow$ & --& 0.682 & 0.666 & {0.657}
                  & --& 0.685 & 0.677 & {0.660}
                  & --& 0.597 & 0.587 & {0.581}
                  & --& 0.777 & 0.737 & {0.721} \\
LPIPS $\uparrow$  & --& 0.535 & 0.553 & {0.576}
                  & --& 0.492 & 0.541 & {0.576}
                  & --& 0.605 & 0.639 & {0.649}
                  & --& 0.439 & 0.482 & {0.502} \\
Feat. Sim. (C) $\downarrow$
& 0.743& 0.690 & 0.688 & 0.685
& 0.744& 0.699 & 0.681 & 0.677
& 0.694& 0.635 & 0.629 & 0.622
& 0.810& 0.756 & 0.746 & 0.735 \\

\Xhline{0.35mm}
\end{tabular}
}
\label{ronaldo}
\end{subtable}

\vspace{3mm}

\begin{subtable}{\linewidth}

\footnotesize
\centering
\caption{Ablation study for the ``Sunflower'' target across different diffusion models.}
\resizebox{\linewidth}{!}{
\begin{tabular}{c|c|ccc|c|ccc|c|ccc|c|ccc}
\Xhline{0.35mm}
{Model}
& \multicolumn{4}{c|}{Stable Diffusion V1.4}
& \multicolumn{4}{c|}{Stable Diffusion V1.5*} 
& \multicolumn{4}{c|}{Stable Diffusion V2.0}
& \multicolumn{4}{c}{InstructPix2Pix}\\
\Xhline{0.1mm}
Method & Clean & Inj$_{\text{DF}}$ & Inj & Inj+Sup 
       & Clean &Inj$_{\text{DF}}$ & Inj & Inj+Sup 
       & Clean &Inj$_{\text{DF}}$ & Inj & Inj+Sup 
       & Clean &Inj$_{\text{DF}}$ & Inj & Inj+Sup \\
\Xhline{0.1mm}
PSNR $\downarrow$ & --&14.91 & 14.68 & {14.43}
                  & --&14.95 & 14.60 & {14.32}
                  & --&12.61 & 12.57 & {12.35}
                  & --&16.22 & 15.96 & {15.42} \\
SSIM $\downarrow$ & --&0.367 & 0.358 & {0.334}
                  & --&0.413 & 0.369  & {0.340}
                  & --&0.284 & 0.283 & {0.267}
                  & --&0.495 & 0.476 & {0.454} \\
VIFp $\downarrow$ & --&0.098 & 0.094 & {0.083}
                  & --&0.117 & 0.103 & {0.088}
                  & --&0.064 & 0.063 & {0.058}
                  & --&0.179 & 0.168 & {0.158} \\
FSIM $\downarrow$ & --&0.666 & 0.660 & {0.651}
                  & --&0.679 & 0.672 & {0.651}
                  & --&0.588 & 0.586 & {0.581}
                  & --&0.776 & 0.764 & {0.742} \\
LPIPS $\uparrow$  & --&0.553 & 0.572 & {0.599}
                  & --&0.544 & 0.561 & {0.610}
                  & --&0.648 & 0.643 & {0.660}
                  & --&0.449 & 0.485 & {0.508} \\
Feat. Sim. (C) $\downarrow$
& 0.743& 0.691 & 0.685 & 0.677
& 0.744& 0.688 & 0.682 & 0.679
& 0.694& 0.628 & 0.627 & 0.624
& 0.810& 0.752 & 0.742 & 0.733 \\

\Xhline{0.35mm}
\end{tabular}
}
\vspace{-1mm}
\label{sunflower}

\end{subtable}

\vspace{3mm}

\begin{subtable}{\linewidth}

\footnotesize
\centering
\caption{Ablation study for the ``Peacock'' target across different diffusion models.}
\resizebox{\linewidth}{!}{
\begin{tabular}{c|c|ccc|c|ccc|c|ccc|c|ccc}
\Xhline{0.35mm}
{Model}
& \multicolumn{4}{c|}{Stable Diffusion V1.4}
& \multicolumn{4}{c|}{Stable Diffusion V1.5*} 
& \multicolumn{4}{c|}{Stable Diffusion V2.0}
& \multicolumn{4}{c}{InstructPix2Pix}\\
\Xhline{0.1mm}
Method & Clean & Inj$_{\text{DF}}$ & Inj & Inj+Sup 
       & Clean &Inj$_{\text{DF}}$ & Inj & Inj+Sup 
       & Clean &Inj$_{\text{DF}}$ & Inj & Inj+Sup 
       & Clean &Inj$_{\text{DF}}$ & Inj & Inj+Sup \\
\Xhline{0.1mm}
PSNR $\downarrow$ & --&14.58 & 14.45 & {14.21}
                  & --&14.75 & 14.26 & {14.22}
                  & --&12.36 & 12.28 & {12.06}
                  & --&16.11 & 15.58 & {15.39} \\
SSIM $\downarrow$ & --&0.308 & 0.304 & {0.284}
                  & --&0.356 & 0.339 & {0.308}
                  & --&0.243 & 0.244 & {0.226}
                  & --&0.460 & 0.430 & {0.420} \\
VIFp $\downarrow$ & --&0.090 & 0.087 & {0.079}
                  & --&0.106 &{0.083} &  {0.088}
                  & --&0.059 & 0.058 & {0.054}
                  & --&0.166 & 0.160 & {0.155} \\
FSIM $\downarrow$ & --&0.646 & 0.643 & {0.634}
                  & --&0.662 & 0.653 & {0.651}
                  & --&0.572 & 0.573 & {0.563}
                  & --&0.773 & 0.759 & {0.754} \\
LPIPS $\uparrow$  & --&0.562 & 0.589 & {0.608}
                  & --&0.548 & 0.608 & {0.610}
                  & --&0.645 & 0.661 & {0.673}
                  & --&0.477 & 0.502 & {0.512} \\
Feat. Sim. (C) $\downarrow$
& 0.743& 0.684 & 0.682 & 0.675
& 0.744& 0.686 & 0.683 & 0.681
& 0.694& 0.631 & 0.624 & 0.622
& 0.810& 0.748 & 0.734 & 0.732 \\

\Xhline{0.35mm}
\end{tabular}
}
\label{peacock}
\end{subtable}

\vspace{3mm}

\begin{subtable}{\linewidth}
\footnotesize
\centering
\caption{Ablation study for the ``Mandala'' target across different diffusion models.}
\resizebox{\linewidth}{!}{
\begin{tabular}{c|c|ccc|c|ccc|c|ccc|c|ccc}
\Xhline{0.35mm}
{Model}
& \multicolumn{4}{c|}{Stable Diffusion V1.4}
& \multicolumn{4}{c|}{Stable Diffusion V1.5*} 
& \multicolumn{4}{c|}{Stable Diffusion V2.0}
& \multicolumn{4}{c}{InstructPix2Pix}\\
\Xhline{0.1mm}
Method & Clean & Inj$_{\text{DF}}$ & Inj & Inj+Sup 
       & Clean &Inj$_{\text{DF}}$ & Inj & Inj+Sup 
       & Clean &Inj$_{\text{DF}}$ & Inj & Inj+Sup 
       & Clean &Inj$_{\text{DF}}$ & Inj & Inj+Sup \\
\Xhline{0.1mm}
PSNR $\downarrow$ & --&14.79 & 14.40 & {14.19}
                  & --&15.20 & 14.89 & {14.25}
                  & --&12.48 & 12.21 & {12.17}
                  & --&15.94 & 15.31& {15.12} \\
SSIM $\downarrow$ & --&0.337 & 0.302 & {0.279}
                  & --&0.342 & 0.313 & {0.301}
                  & --&0.249 & 0.226 & {0.215}
                  & --&0.476 & 0.409 & {0.393} \\
VIFp $\downarrow$ & --&0.088 & 0.074 & {0.068}
                  & --&0.092 & {0.077} & {0.078}
                  & --&0.055 & 0.050 & {0.049}
                  & --&0.166 & 0.143 & {0.136} \\
FSIM $\downarrow$ & --&0.658 & 0.639 & {0.634}
                  & --&0.651 & 0.645 & {0.641}
                  & --&0.576 & {0.562} & 0.564
                  & --&0.769 & 0.735 & {0.723} \\
LPIPS $\uparrow$  & --&0.558 & 0.602 & {0.621}
                  & --&0.587 & 0.615 & {0.621}
                  & --&0.641 & 0.665 & {0.674}
                  & --&0.455 & 0.517 & {0.534} \\
Feat. Sim. (C) $\downarrow$
& 0.743& 0.682 & 0.670 & 0.668
& 0.744& 0.678 & 0.674 & 0.673
& 0.694& 0.630  & 0.618 & 0.616
& 0.810& 0.750 & 0.730 & 0.726 \\

\Xhline{0.35mm}
\end{tabular}
}
\vspace{-1mm}
\label{mandala_}
\end{subtable}
\label{mandala}
\vspace{3mm}

\end{table}
\begin{table}[t!]
\footnotesize
\centering
\caption{Comparison of different data sources for source semantic suppression loss. 
Best results are shown in bold; second-best are underlined.}
\resizebox{0.65\linewidth}{!}{
\begin{tabular}{c|c|cc}
\Xhline{0.35mm}
Method & Clean &  Laion$_\text{sup}$ &ImageNet-Excluded$_\text{sup}$ \\
\Xhline{0.1mm}

PSNR $\downarrow$
& --
& \underline{16.11}
& \textbf{16.06} \\

SSIM $\downarrow$
& --
& \textbf{0.420}
& \underline{0.440} \\

VIFp $\downarrow$
& --
& \textbf{0.140}
& \underline{0.142} \\

FSIM $\downarrow$
& --
& \textbf{0.697}
& \underline{0.700} \\

LPIPS $\uparrow$
& --
& \textbf{0.490}
& \underline{0.485}\\

Feat. Sim. (CLIP) $\downarrow$
& 0.744
& \textbf{0.688}
& \underline{0.699} \\


\Xhline{0.35mm}
\end{tabular}
}
\label{R3_W3}
\end{table}

\vspace{-0.3cm}
\subsection{Robustness of Semantic Suppression Loss}
While the main manuscript already demonstrates the robustness of target semantic injection via our data-free variant (Table~\textcolor{red}{1} and~\textcolor{red}{2} in manuscript), we conduct additional experiments to isolate and assess the robustness of the source semantic suppression loss.
Specifically, we build an ``ImageNet-Excluded'' training set by removing all 136 classes semantically related to the 10 inference categories (including their WordNet superclasses and subclasses) from the ImageNet evaluation set, and then randomly sampling 10,000 images from the remaining classes to train the UAP.
This setting tests whether source semantic suppression remains effective when the training data excludes semantics related to the inference categories.
As shown in Table~\ref{R3_W3}, the UAP trained on unseen semantics (ImageNet-Excluded) achieves performance comparable to the original version trained on LAION.
This indicates that the source semantic suppression loss is not tightly coupled to the semantics seen during training, but instead transfers effectively to unseen categories and domains, supporting robust protection under distribution shift.

\vspace{-0.3cm}
\subsection{Impact of Timestep Set}
\vspace{-0.1cm}
In our experiments, we set the timestep set to $K=\{5, 10, 15, 20, 25\}$.
To analyze the impact of the selected timesteps, we further conduct ablation studies using two alternative timestep sets: $K_1 = \{2, 4, 6, 8, 10\}$ and $K_2 = \{30, 35, 40, 45, 50\}$.
Table~\ref{tab:timestep_abl} demonstrates that our method maintains consistently strong performance across different choices of timestep sets, suggesting that it is relatively invariant to the selection of diffusion timesteps.
\begin{table}[h]
\footnotesize
\centering
\caption{Ablation study on the timestep set.}
\resizebox{0.45\linewidth}{!}{
\begin{tabular}{c|c|ccc}
\Xhline{0.35mm}
Timestep Set & Clean &  $K_1$ & $K_2$ & $K$ (Ours) \\
\Xhline{0.1mm}
PSNR $\downarrow$ & --& 14.42 & 14.37 & 14.19 \\
SSIM $\downarrow$ & --& 0.354 & 0.370 & 0.332 \\
VIFp $\downarrow$ & --& 0.089 & 0.092 & 0.082 \\
FSIM $\downarrow$ & --& 0.654 & 0.652 & 0.642 \\
LPIPS $\uparrow$  & --& 0.600 & 0.597 & 0.606 \\
Feat. Sim.(C)$\downarrow$ & 0.744 & 0.675 & 0.678 & 0.673 \\
\Xhline{0.35mm}
\end{tabular}
}
\label{tab:timestep_abl}
\end{table}


\subsection{Effect of Number of Training Samples}
We analyze the effect of training set size to investigate how the robustness of our method scales with data availability.
As shown in Table~\ref{R1_W3}, immunization performance consistently improves as the number of training samples increases, suggesting that more training data helps the semantic injection and suppression objectives learn a UAP that encodes target semantics more robustly.
These results further suggest that leveraging more diverse training data could lead to even more effective protection.
\begin{table}[h!]
\footnotesize
\centering
\caption{Performance comparison across different numbers of training samples. 
Best results are in bold, second-best are underlined.}
\resizebox{0.75\linewidth}{!}{
\begin{tabular}{c|c|ccc|ccccc}
\Xhline{0.35mm}

\multirow{2}{*}{Metric}
& \multirow{2}{*}{Clean}
& \multicolumn{3}{c|}{\textbf{Image-specific}}
& \multicolumn{5}{c}{\textbf{Universal}} \\
&  & EA & DA & SA 
& \#10 & \#100 & \#1000 & \#10000 
& Ours$_{DF}$ \\
\Xhline{0.1mm}

PSNR $\downarrow$
& -- 
& 14.75 & 14.71 & 16.28 
& 15.14 & 14.52 & \underline{14.37} & \textbf{14.19}
& 14.68 \\

SSIM $\downarrow$
& -- 
& 0.386 & 0.382 & 0.431 
& 0.408 & 0.373 & \underline{0.359} & \textbf{0.332}
& 0.378 \\

VIFp $\downarrow$
& -- 
& \underline{0.088} & 0.106 & 0.138 
& 0.124 & 0.100 & 0.093 & \textbf{0.082}
& 0.106 \\

FSIM $\downarrow$
& -- 
& \textbf{0.637} & 0.660 & 0.714 
& 0.681 & 0.658 & {0.652} & \underline{0.642}
& 0.666 \\

LPIPS $\uparrow$
& -- 
& \underline{0.584} & 0.552 & 0.490 
& 0.517 & 0.564 & 0.575 & \textbf{0.606}
& 0.557 \\

Feat. Sim. (CLIP) $\downarrow$
& 0.744 
& 0.677 & \textbf{0.662} & 0.705 
& 0.696 & 0.686 & 0.682 & \underline{0.673}
& 0.685 \\

\Xhline{0.35mm}
\end{tabular}}
\label{R1_W3}
\end{table}









At the same time, as demonstrated in Tables~\textcolor{red}{1},~\textcolor{red}{2}, and~\textcolor{red}{4} of the main manuscript, even in the data-free setting, our method substantially outperforms universalized baselines trained with data and remains competitive with image-specific methods.
These results highlight the practicality of our approach by showing strong robustness across both data-rich and data-free scenarios.

\vspace{4mm}
\section{More Qualitative Results}
\label{sec:quali_sec}
\vspace{4mm}
\subsection{Impact of Diverse Targets}
\label{app:diverse_target}
\begin{figure}[h]
\centering    
\includegraphics[width=\linewidth]{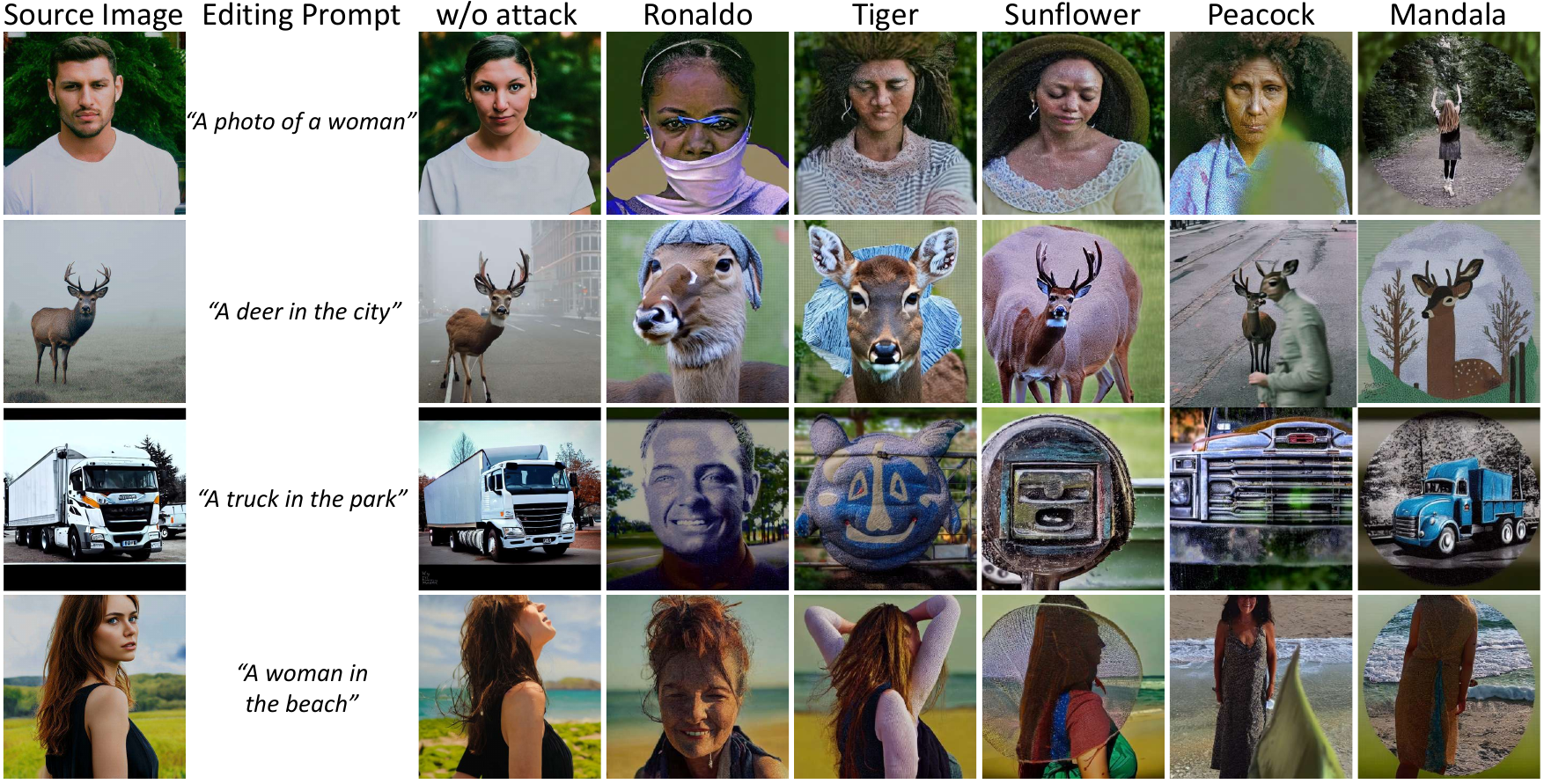}  
    \caption{Qualitative results for diverse target contents.
    }
    \label{fig:qual_target}
\end{figure}

We visualize the variation in editing outcomes across different targets in Figure~\ref{fig:qual_target}.
Across different target choices, the edited outputs consistently exhibit reduced reliance on the original source image and are instead conditioned on the injected target semantics (in Figure~\ref{fig:tar_uap_vis}).
This indicates that the immunized image is semantically misinterpreted as the target content rather than the source content.
Interestingly, target-related traits are occasionally observed in the edited outputs.
For example, the target `\textit{Ronaldo}' sometimes leads to human-like attributes, `\textit{Tiger}' may induce animal-like structures such as ear-like features, `\textit{Sunflower}' can produce circular patterns, and `\textit{Peacock}' may yield new structures at consistent spatial regions.
In contrast, when using a more abstract target such as `\textit{Mandala}', the edited outputs do not preserve the target pattern itself, but deviate substantially from the original source images.
These observations suggest that the tendency is more pronounced when the target is semantically distinctive, as its semantic characteristics are more likely to persist even after prompt-driven editing.

\subsection{Qualitative Comparison with Universalized Baselines}
We present a more qualitative comparison with universalized baselines (\ie Encoder, Embedding, and Map) on U-Net-based Stable Diffusion V1.5~\cite{rombach2022high} and DiT-based Stable Diffusion V3~\cite{sd3}.
In Figure~\ref{fig:attn_vis}, our method consistently disrupts the reference to the original image’s semantics, thereby causing the editing process to be conditioned on the target semantic (\eg, tiger) rather than source, regardless of the input image.
In the DiT-based Stable Diffusion V3 shown in Figure~\ref{fig:sd3_universal}, although the edited outputs do not clearly reflect the injected semantics, our method still exhibits reduced reliance on the original source images, indicating effective immunization transferability across different architectures.

\begin{figure}[t!]
\centering    \includegraphics[width=\linewidth]{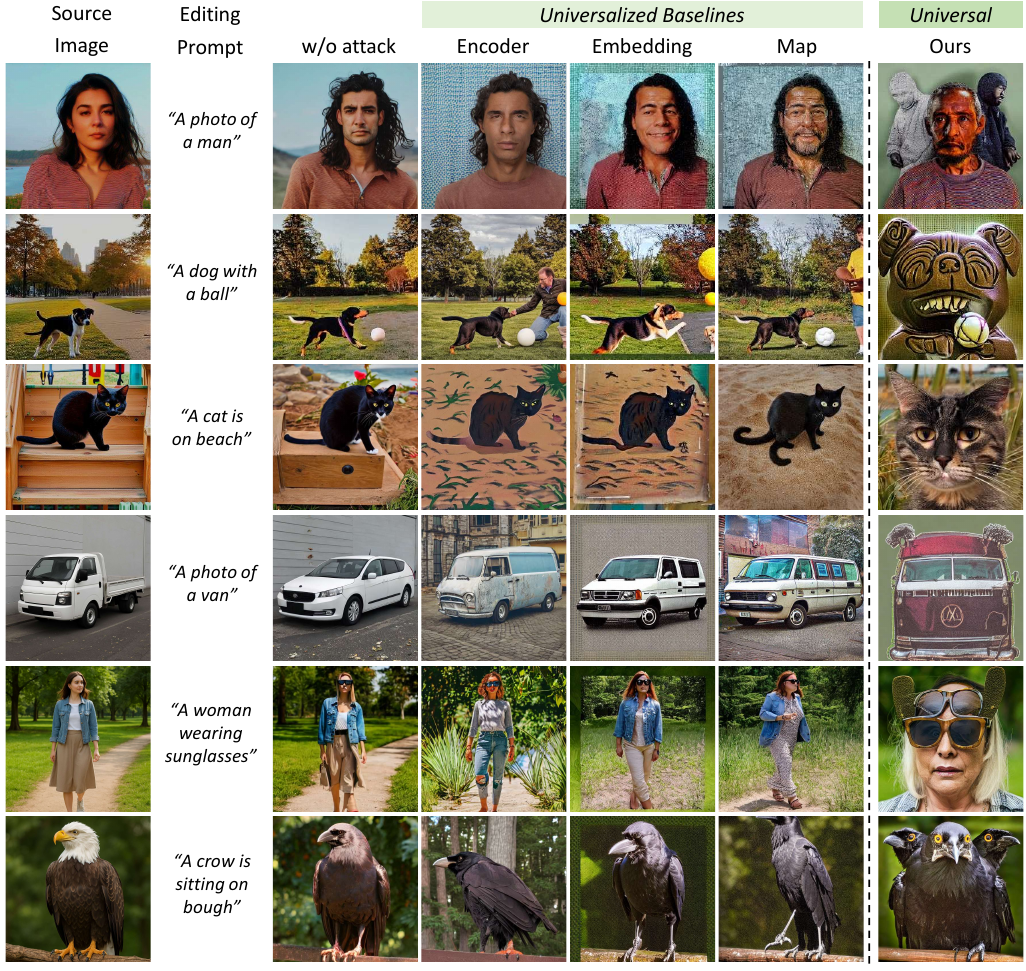}    
\vspace{-0.75cm}
    \caption{Qualitative comparison with universalized baselines on Stable Diffusion V1.5. 
    }
    \label{fig:attn_vis}
\vspace{0.3cm}
\centering    \includegraphics[width=\linewidth]{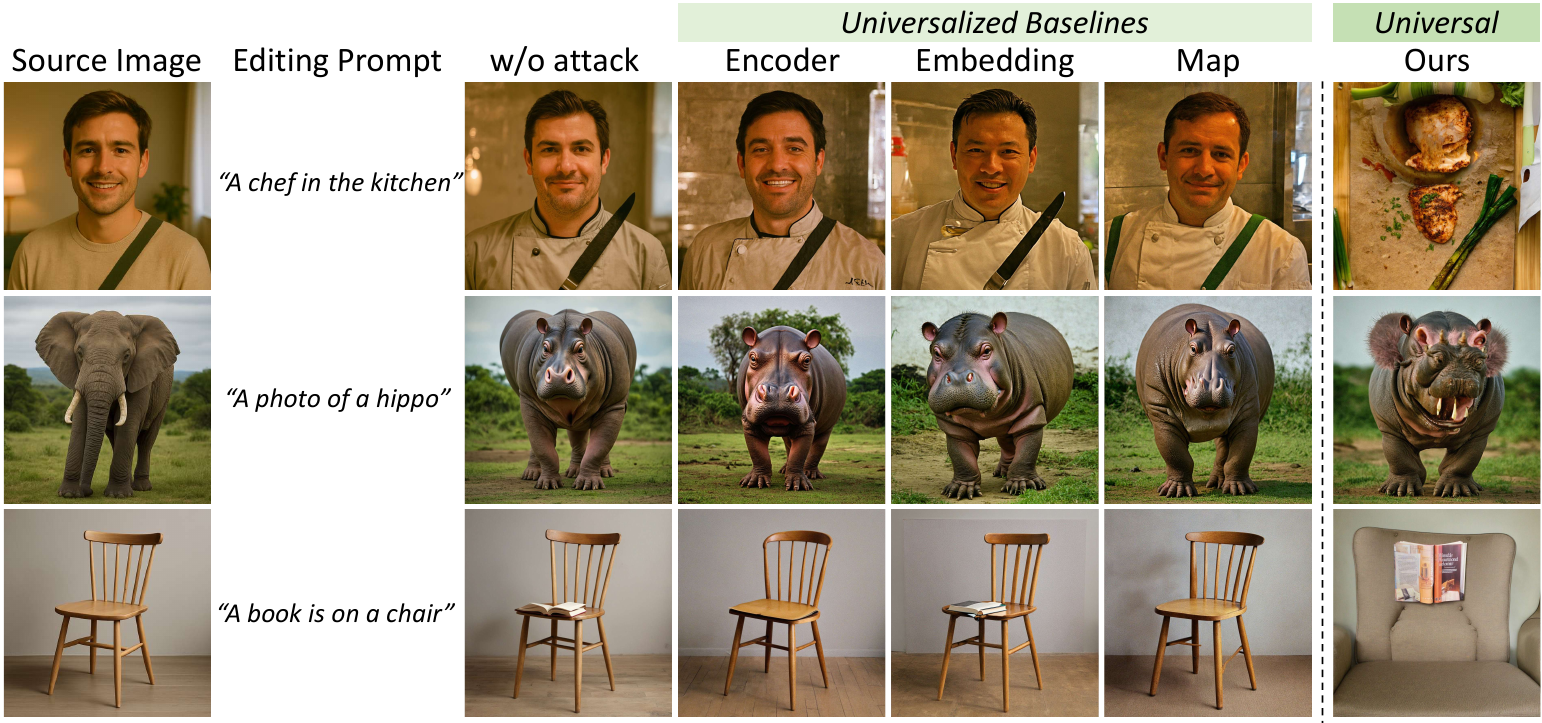}    
    \vspace{-0.7cm}
    \caption{Qualitative comparison with universalized baselines on Stable Diffusion V3~\cite{sd3}. 
    }
    \label{fig:sd3_universal}
\end{figure}

\subsection{Qualitative Comparison with Image-specific Methods}
We provide additional image manipulation results compared with image-specific baselines, EA, DA~\cite{raising}, SA~\cite{semantic}, and FP~\cite{nearly} using a U-Net-based (Stable Diffusion V1.5~\cite{rombach2022high}) and DiT-based (Stable Diffusion V3~\cite{sd3})generative models.
In the experiments, we use the UAP generated with `\textit{tiger}' as the target concept.
In Figure~\ref{fig:sd15_edit-specific}, under U-Net-based editing models, our method consistently produces editing outputs that are weakly conditioned on the original source image, indicating strong disruption of source-content preservation compared to image-specific immunization approaches.
We attribute these results to the stronger semantic shift induced by our targeted UAP.
As shown in Figure~\ref{fig:SD3_quali}, although our semantic injection is optimized on U-Net backbones and transfer to DiT-based manipulation is inherently challenging, our method still provides visually reliable protection on DiT-based model, demonstrating cross-architecture transferability.

\vspace{-0.1cm}

\begin{figure}[t!]
\centering    \includegraphics[width=\linewidth]{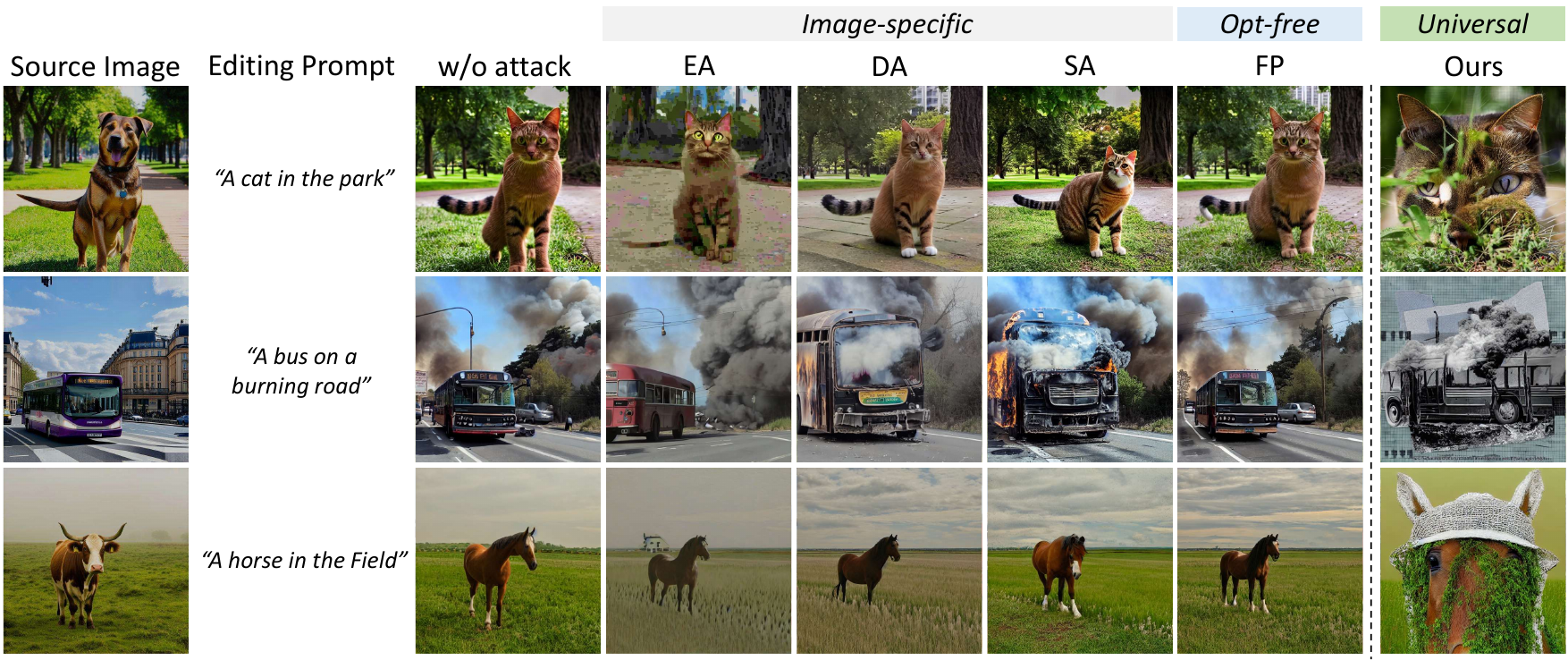}   
\vspace{-0.5cm}
    \caption{Qualitative comparison for image editing with image-specific and optimization-free baselines on Stable Diffusion V1.5~\cite{rombach2022high}. The target of UAP is `\textit{tiger}.
    }
    \label{fig:sd15_edit-specific}
\end{figure}

\begin{figure}[t!]
\centering    \includegraphics[width=\linewidth]{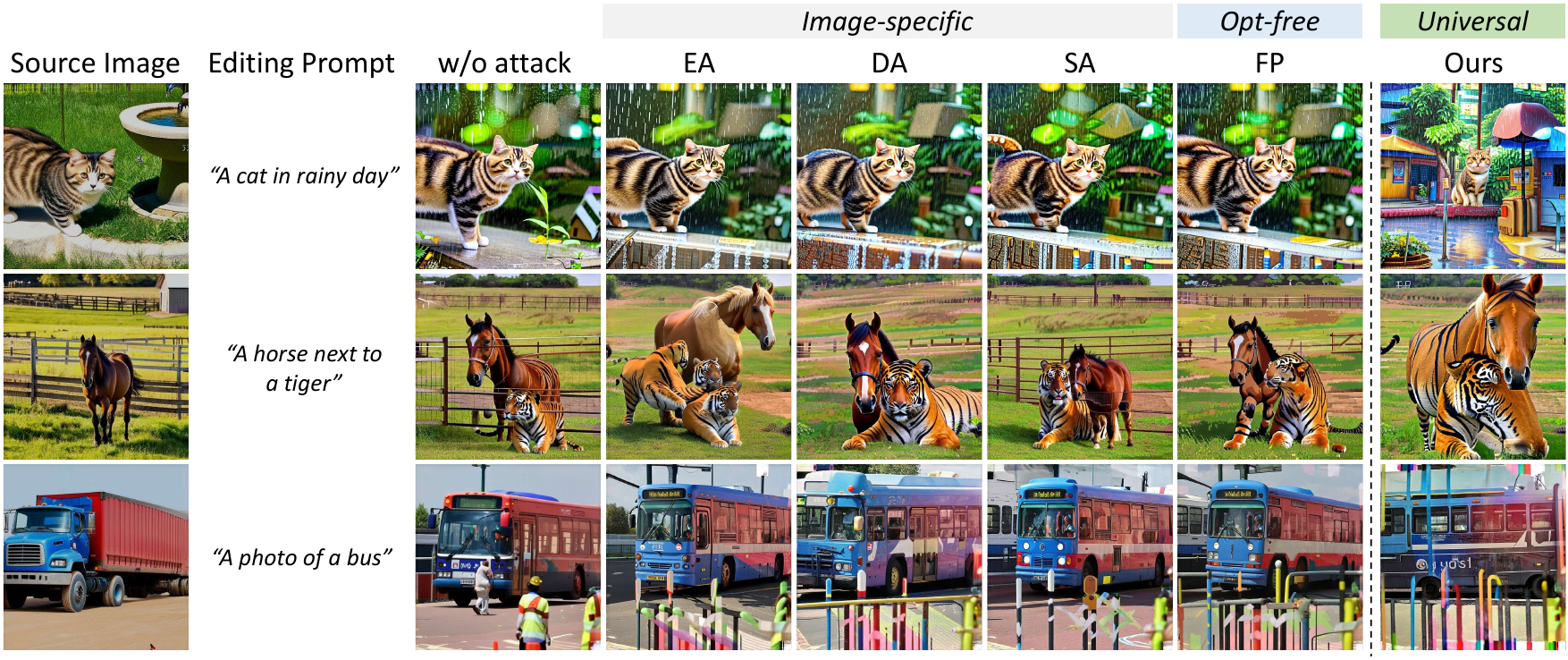}    
\vspace{-0.5cm}
    \caption{Qualitative comparison for image editing with image-specific and optimization-free baselines for image editing with Stable Diffusion V3~\cite{sd3}. The target is `\textit{tiger}.'
    }
    \label{fig:SD3_quali}
\end{figure}

\begin{figure}[t!]
\centering    \includegraphics[width=\linewidth]{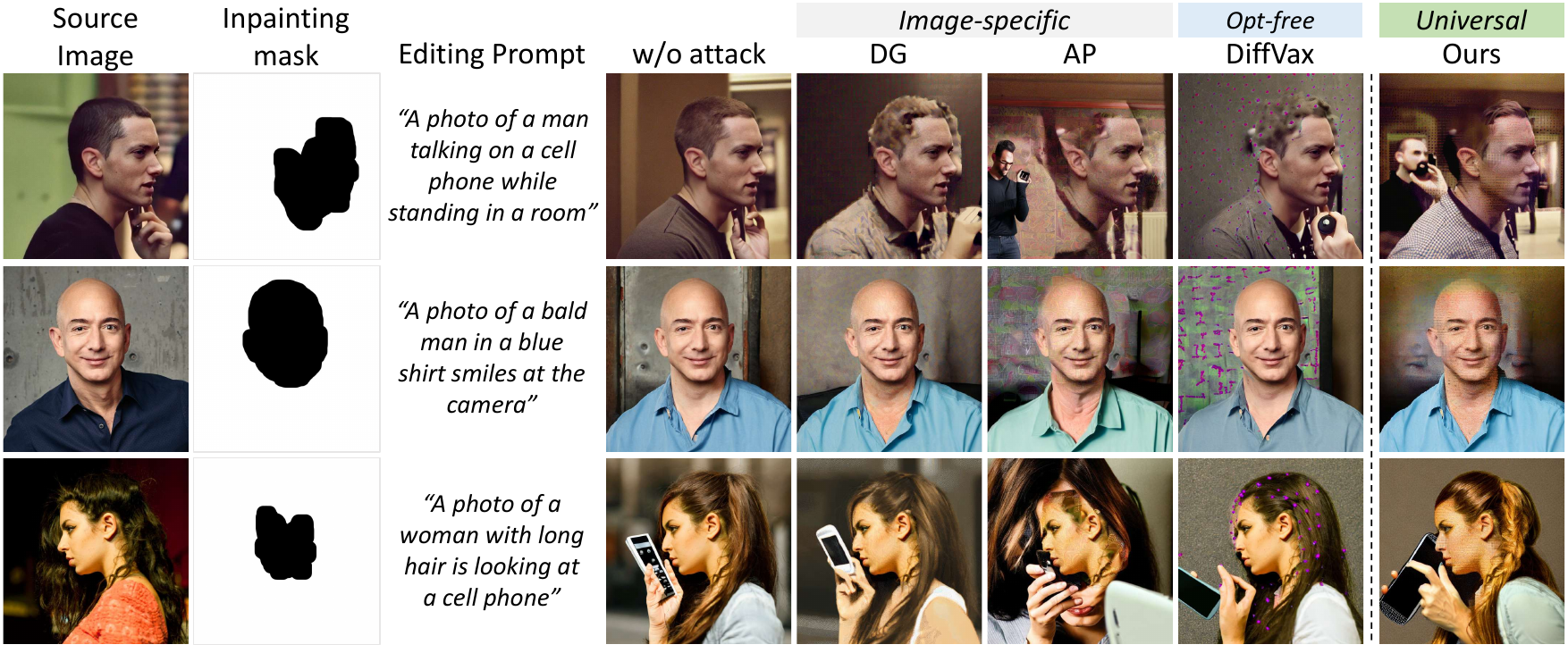}    

\vspace{-0.3cm}
    \caption{Qualitative results for image inpainting with Stable Diffusion Inpainting V2.0~\cite{rombach2022high}. The inpainting is applied to the white regions of the mask and the target of UAP is `\textit{tiger}.'
    }
    \vspace{-0.1cm}
    \label{fig:sd15_inpaint_quali}
\vspace{-0.8cm}
\centering    \includegraphics[width=\linewidth]{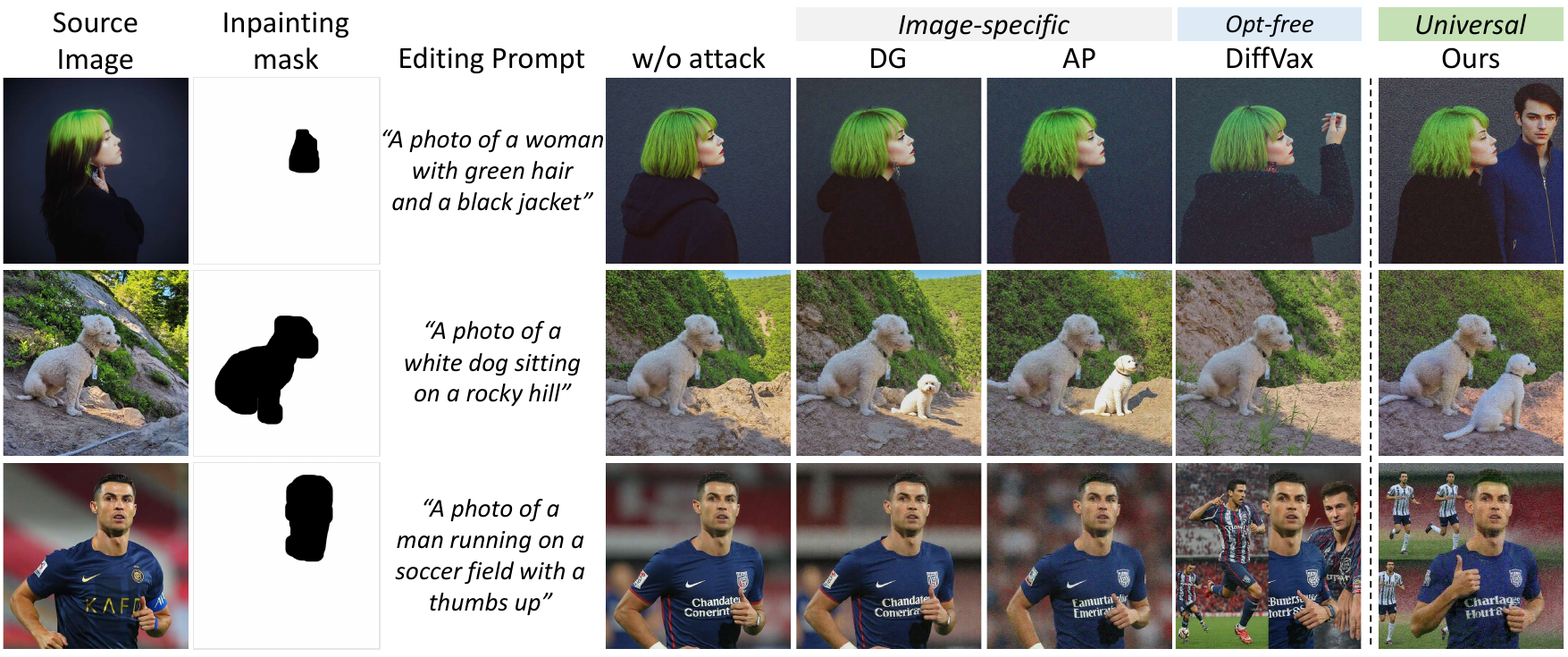}    
\vspace{-3.1cm}
    \caption{Qualitative results for image inpainting with FLUX~\cite{flux}. White regions in the mask indicate areas to be inpainted. The target of UAP is `\textit{tiger}.'
    }
    \vspace{-0.6cm}
    \label{fig:flux_quali}
\end{figure}

\subsection{Qualitative Results for Image Inpainting}
\vspace{-0.2cm}
To assess the generalization of our method, we additionally perform qualitative evaluations on both U-Net– and DiT-based inpainting models, using Stable Diffusion V2.0~\cite{rombach2022high} (black-box setting for all methods) and FLUX 1.0~\cite{flux}, respectively, with the target concept `\textit{tiger},' consistent with the image editing experiments.
DG and AP denote DiffusionGuard~\cite{diffguard} and AdvPaint~\cite{advpaint}, respectively, both specifically designed for inpainting immunization.
As shown in Figure~\ref{fig:sd15_inpaint_quali} and~\ref{fig:flux_quali}, although our method is not tailored for inpainting, it still achieves visually competitive results compared to AP, DG, and DiffVax~\cite{diffvax}.
Specifically, in the U-Net–based model (Figure~\ref{fig:sd15_inpaint_quali}), the inpainting model either interprets the object as missing and synthesizes a new object based solely on the prompt, or yields unrealistic outputs.
A similar visual trend is also observed in the DiT-based model (Figure~\ref{fig:flux_quali}), where our method either generates a new object consistent with the prompt or produces unrealistic outputs.
These results suggest that our approach holds promise for extension as a defense mechanism against malicious diffusion-based inpainting.



\subsection{Qualitative Results on Real Images}

\begin{figure}[t]
\centering    \includegraphics[width=\linewidth]{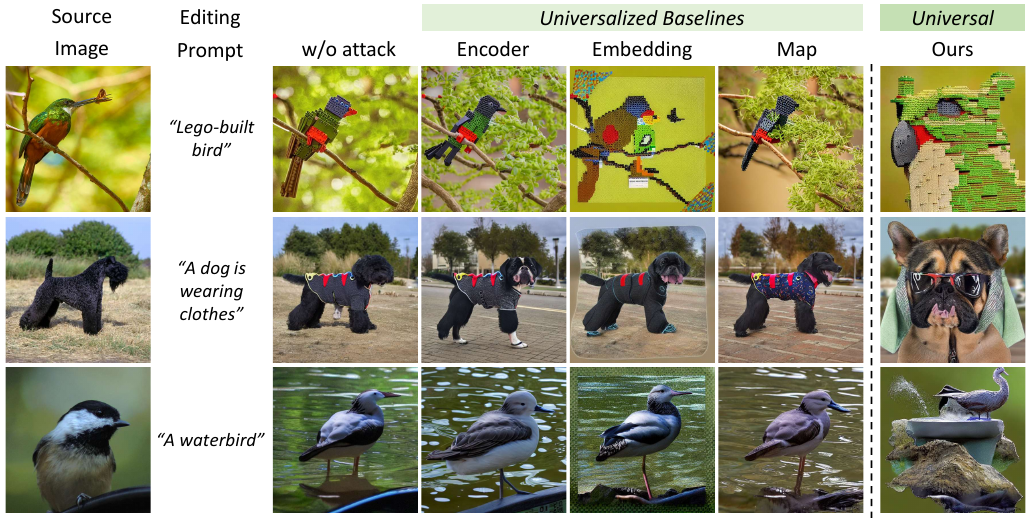}  
\vspace{-0.4cm}
    \caption{Qualitative comparison with universalized baselines on ImageNet-Edit dataset. 
    }
    \label{fig:imagenet-uap}
\vspace{0.3cm}
\centering    \includegraphics[width=\linewidth]{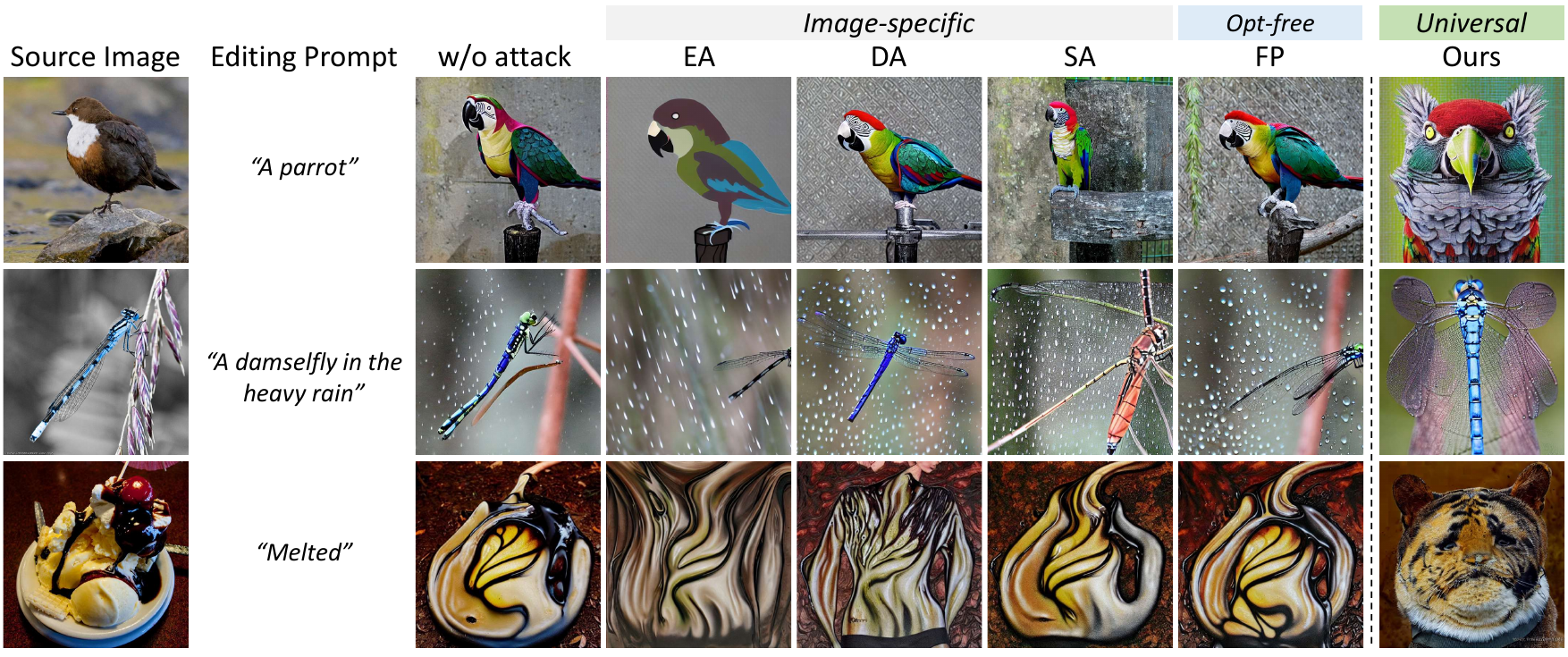}  
\vspace{-0.4cm}
    \caption{Qualitative comparison with image-specific and optimization-free immunization methods on ImageNet-Edit dataset. 
    }
    \vspace{-0.2cm}
    \label{fig:imagenet-specific}
\end{figure}

We compare our method with universalized baselines (\ie Encoder, Embbeding, Map) on real-world images in ImageNet-Edit dataset~\cite{blurguard}.
In Figure~\ref{fig:imagenet-uap}, our method consistently produces editing results that are more conditioned on the target semantics than on the original \textit{real-image} content, and further demonstrates superior immunization performance compared with universalized baselines.

We further provide a comparison with image-specific (EA, DA~\cite{raising}, SA~\cite{semantic}) and optimization-free (FP~\cite{nearly}) methods in Figure~\ref{fig:imagenet-specific}. Compared with image-specific and optimization-free methods, our universal approach yields visually more reliable defense results, and, together with Table~\ref{tab:imagenet-edit-main} and~\ref{imagenet-specific-bb}}, further validates the superiority of our method on real images in both quantitative and qualitative terms.
\begin{figure}[t!]
\centering    \includegraphics[width=\linewidth]{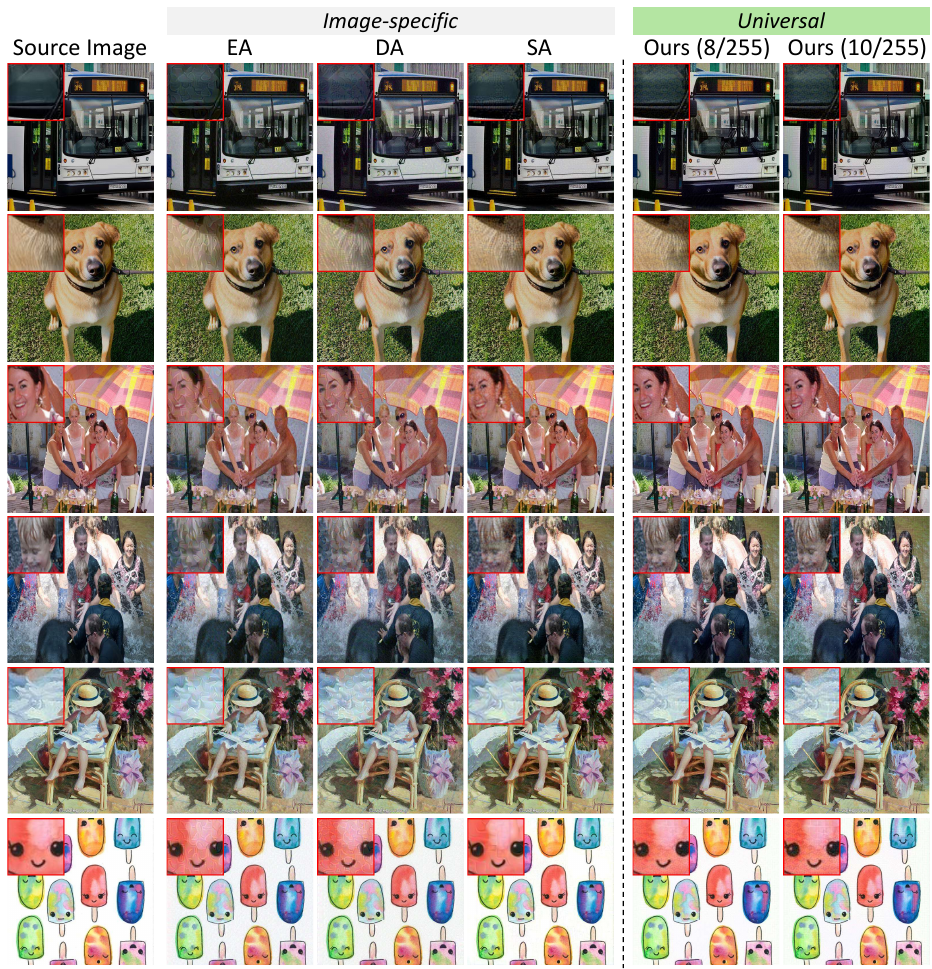}    
\vspace{-0.3cm}
    \caption{Comparison of immunized images produced by our method and previous image-specific approaches. Ours (8/255) and Ours (10/255) denote our method applied with $\epsilon = 8/255$ and $\epsilon = 10/255$respectively. The target of UAP is `\textit{tiger}.'
    }
    \label{fig:epsilon_ae}
\end{figure}

\subsection{Quality of Immunized Images}

We qualitatively compare the immunized images produced by prior image-specific methods that achieve strong immunization performance with those defended using our UAP.
As shown in Figure~\ref{fig:epsilon_ae}, the immunized images generated by our UAP exhibit artifacts that are comparably imperceptible to those produced by image-specific methods, even across diverse domains such as real-world and paintings
This observation is consistent with the high imperceptibility scores obtained in our human evaluation in Table~\ref{tab:user_1} and further highlights the practicality of our universal approach to real-world application.

\vspace{-0.2cm}
\subsection{Failure Cases}
Figure~\ref{fig:fail} illustrates several failure cases of our method.
In particular, when the protected image shares similar shapes or characteristics with the target semantics embedded by the UAP, our approach may fail to effectively block editing attempts, as shown in Figure~\ref{fig:fail_a}.
Nevertheless, this limitation can be mitigated by selectively choosing diverse targets, and our ablation study on diverse targets (see in Section~\ref{app:diverse_quan} and~\ref{app:diverse_target}) support this claim. 
Moreover, as shown in Figure~\ref{fig:fail_b}, our method occasionally does not prevent undesired inpainting. While it performs well in most image-to-image editing scenarios, it exhibits limitations in inpainting, particularly when the editing region is excessively large or extremely small relative to the overall image size. In contrast, inpainting-specific baselines such as DG~\cite{diffguard}, AP~\cite{advpaint}, and DiffVax~\cite{diffvax}, which are trained with access to human-provided masks or masks explicitly generated by GroundedSAM~\cite{groundedsam}, more effectively suppress such edits. Further discussion of these failure cases is provided in Sec.~\ref{limit}.




\begin{figure}[t]
\centering
\begin{subfigure}{\linewidth}
    \centering
    \includegraphics[width=\linewidth]{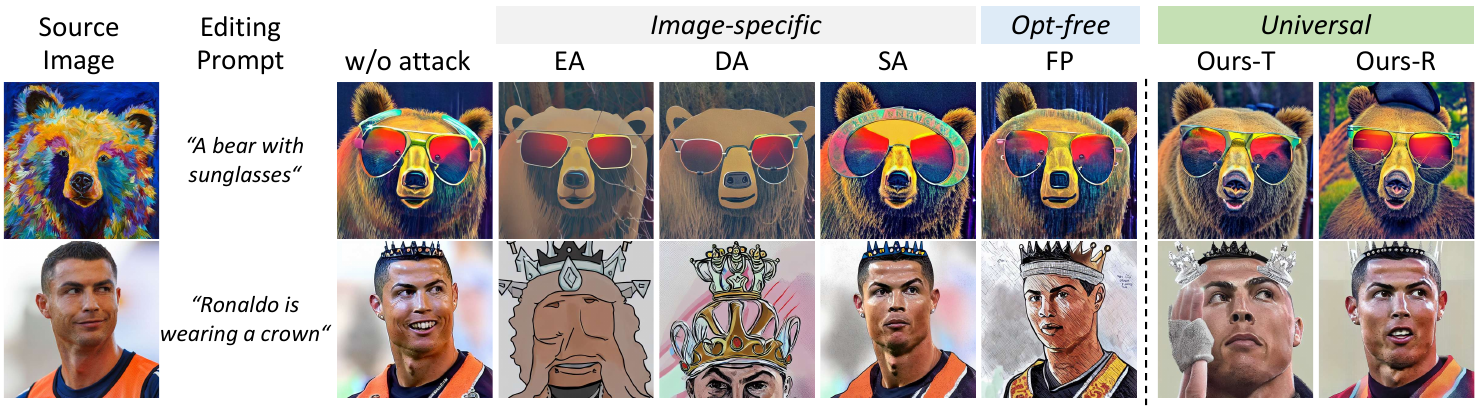}
    \vspace{-0.4cm}
    \caption{Failure cases on image editing immunization.}
    \label{fig:fail_a}
\end{subfigure}
\vspace{0.2cm}
\begin{subfigure}{\linewidth}
    \centering
    \includegraphics[width=\linewidth]{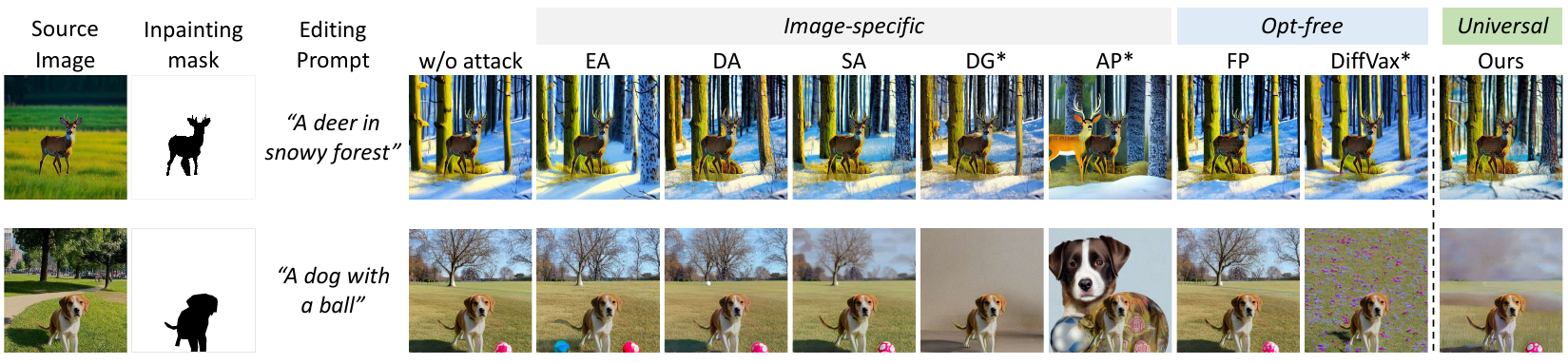}
    \vspace{-0.4cm}
    \caption{Failure cases on image inpainting immunization.}
    \label{fig:fail_b}
\end{subfigure}
\vspace{-0.5cm}
\caption{Visualization of failure cases. 
(a) \textit{Ours-T} and \textit{Ours-R} denote UAP is optimized with \textit{tiger} and \textit{ronaldo} as the targets, respectively.
(b) The white regions in the mask indicate areas where the generative model synthesizes new content. * refers inpainting-specified immunization methods.}
\vspace{-0.2cm}
\label{fig:fail}
\end{figure}


\section{Limitations and Discussions}
\label{limit}

{Despite the strong qualitative and quantitative results, our method encounters limitations in certain cases. One such limitation lies in its universality: unlike existing immunization methods that produce image-specific perturbations, our UAP is a single, input-agnostic perturbation shared across all images.
Consequently, when the structure or shape of an input image closely resembles that of the target images used to train the UAP (see in Figure~\ref{fig:tar_uap_vis} and~\ref{fig:fail_a}), semantic injection may fail, resulting in unsuccessful prevention of malicious editing. 
However, this limitation can be readily mitigated by switching the target semantics by using a different target prompt (\eg,\textit{Ours-R} for `\textit{Bear}' and \textit{Ours-T} for `\textit{Ronaldo}'), or, even under the same target prompt, by employing multiple target images during UAP construction (in Figure~\ref{fig:tigers}).}

Another limitation arises in inpainting scenarios.
As shown in Figure~\ref{fig:fail_b}, our method can be less robust to diverse inpainting masks (\ie, varying mask sizes and shapes), leading to occasional failures to fully prevent undesired edits when a substantial portion of the perturbation is removed or the conditioning signal is weakened.
This limitation is partly due to the fact that our approach is originally designed for image editing rather than inpainting, whereas inpainting-specific methods like AP~\cite{advpaint}, DG~\cite{diffguard}, and DiffVax~\cite{diffvax} explicitly incorporate human-provided masks or masks explicitly generated by GroundedSAM~\cite{groundedsam} during the training of their perturbations~\cite{diffguard, advpaint} and immunizer~\cite{diffvax}, resulting in stronger performance for such tasks.
However, this phenomenon is not unique to our method, as similar failures are observed in other approaches like EA, DA~\cite{raising}, SA~\cite{semantic}, and FP~\cite{nearly}, which also do not leverage masks during training.


\section{Broader Impacts}
\label{app:societal}
Our work introduces a universal adversarial perturbation (UAP) that effectively immunizes images against a wide range of diffusion-based manipulations using a single perturbation.
This lightweight, data-agnostic, and generalizable defense can help protect users from malicious image editing or content forgery, particularly in cases where unauthorized modifications may cause misinformation, reputational damage, or social harm.



However, we also recognize the potential for misuse.
Since our approach manipulates the behavior of generative models through adversarial signals, similar techniques could be exploited to bypass safety filters, induce unintended outputs, or interfere with intended ones.
In particular, targeted UAPs could be reverse-engineered to maliciously exploit vulnerabilities in diffusion models.
\section{Future Work}
In this work, we presented the first universal image immunization framework that enabled practical protection against diffusion-based image manipulation by eliminating test-time image adaptation while achieving strong immunization performance with near-zero additional runtime and without GPU memory overhead, thereby further realizing the practical objective pursued by prior optimization-free methods~\cite{nearly, diffvax}.

While our method is not explicitly tailored to withstand purification, it still outperforms universalized baselines after purification and remains competitive with or even outperforms representative image-specific and optimization-free methods~\cite{raising, semantic, nearly}.
Building on this promising result, an important direction for future work on universal immunization frameworks is to further improve robustness against purification-based defenses, as explored in prior image-specific immunization methods~\cite{blurguard, gridrobust, dct}, which have attracted growing attention in image protection.
\end{document}